%% file: main.tex
\documentclass[acmtog]{acmart}

\makeatletter

\copyrightyear{2024}
\acmYear{2024}
\setcopyright{rightsretained}
\acmConference[SA Conference Papers '24]{SIGGRAPH Asia 2024 Conference Papers}{December 3--6, 2024}{Tokyo, Japan}
\acmBooktitle{SIGGRAPH Asia 2024 Conference Papers (SA Conference Papers '24), December 3--6, 2024, Tokyo, Japan}
\acmDOI{10.1145/3680528.3687570}
\acmISBN{979-8-4007-1131-2/24/12}

\input{macros}

\title{\ourmethod{}: Generating Low-Resolution Quantized Imagery via Score Distillation}

\author{Alexandre Binninger}
\affiliation{\institution{ETH Zurich}\city{Zurich}\country{Switzerland}}
\author{Olga Sorkine-Hornung}
\affiliation{\institution{ETH Zurich}\city{Zurich}\country{Switzerland}}

\begin{document}

\begin{abstract}
  \input{sections/0_abstract.tex}
\end{abstract}

\ccsdesc[500]{Computing methodologies~Image processing}
\ccsdesc[100]{Applied computing~Fine arts}
\ccsdesc[300]{Computing methodologies~Image representations}
\keywords{pixel art, image processing}

\begin{teaserfigure}
  \includegraphics[width=\linewidth]{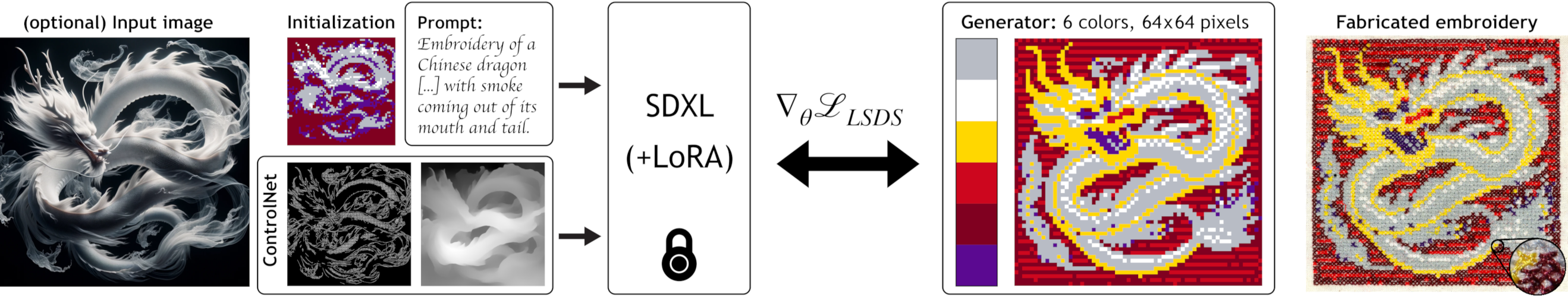}
  \Description{\ourmethod{} Teaser}
  \caption{\ourmethod{} specializes in creating pixel art, characterized by its intentionally low resolution and limited color palette. Our method enables varying degrees of control: the input is a text prompt, and optionally a reference (high-resolution) image for initialization or spatial control. \ourmethod{}'s output style can be adjusted using fine-tuned diffusion models. In this example, the full prompt reads ``Embroidery of a Chinese dragon flying through the air on a dark background with smoke coming out of its mouth and tail.''. The output pixel art can be used for crafted fabrications, such as the shown cross-stitch embroidery.}
  \label{fig:teaser}
  \Description{Teaser image.}
\end{teaserfigure}


\maketitle

\input{sections/1_introduction.tex}
\input{sections/2_related_work.tex}
\input{sections/3_background.tex}

\input{sections/4_method.tex}
\input{sections/5_results.tex}

\input{sections/6_conclusion.tex}
\input{sections/7_acknowledgments.tex}

\clearpage
\balance
\bibliographystyle{ACM-Reference-Format}
\bibliography{references}
\newpage
\clearpage
\input{sections/big_figure.tex}

\end{document}


\begin{abstract}
  \input{sections/supplementary/0_abstract_suppl}
\end{abstract}

\ccsdesc[500]{Computing methodologies~Image processing}
\ccsdesc[100]{Applied computing~Fine arts}
\ccsdesc[300]{Computing methodologies~Image representations}
\keywords{pixel art, image processing}

\begin{teaserfigure}
    \centering
	\setlength{\tabcolsep}{1pt}
 \footnotesize
    \begin{tabular}{cccccc}
    \includegraphics[width=0.16\linewidth]{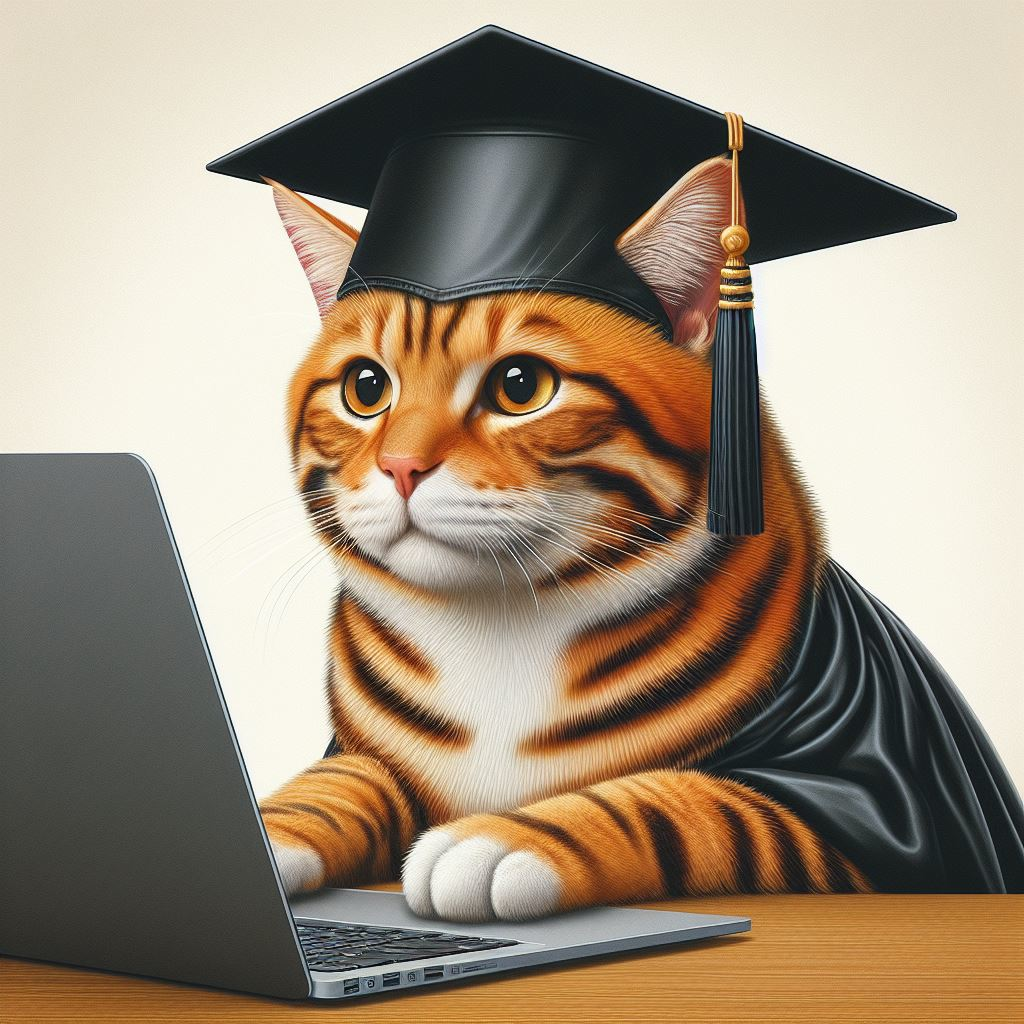} & \includegraphics[width=0.16\linewidth]{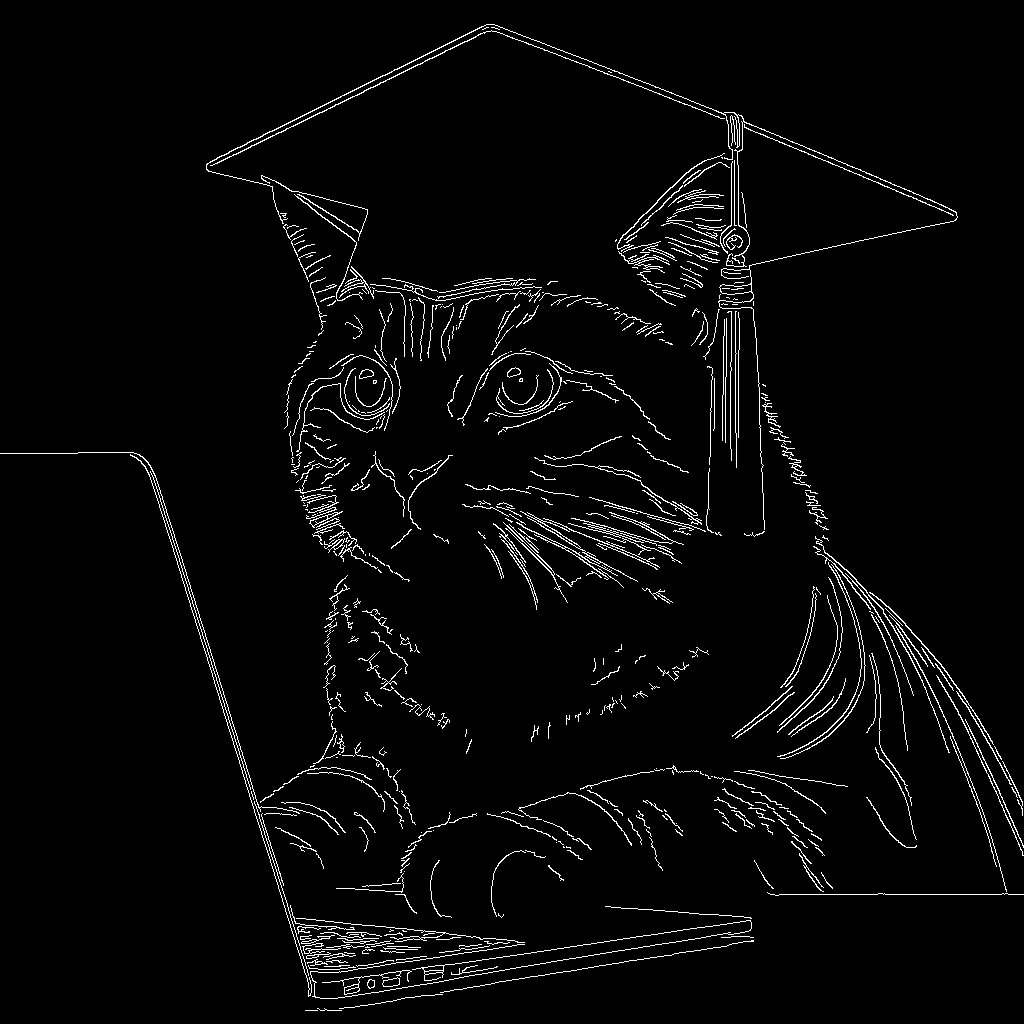} & \includegraphics[width=0.16\linewidth]{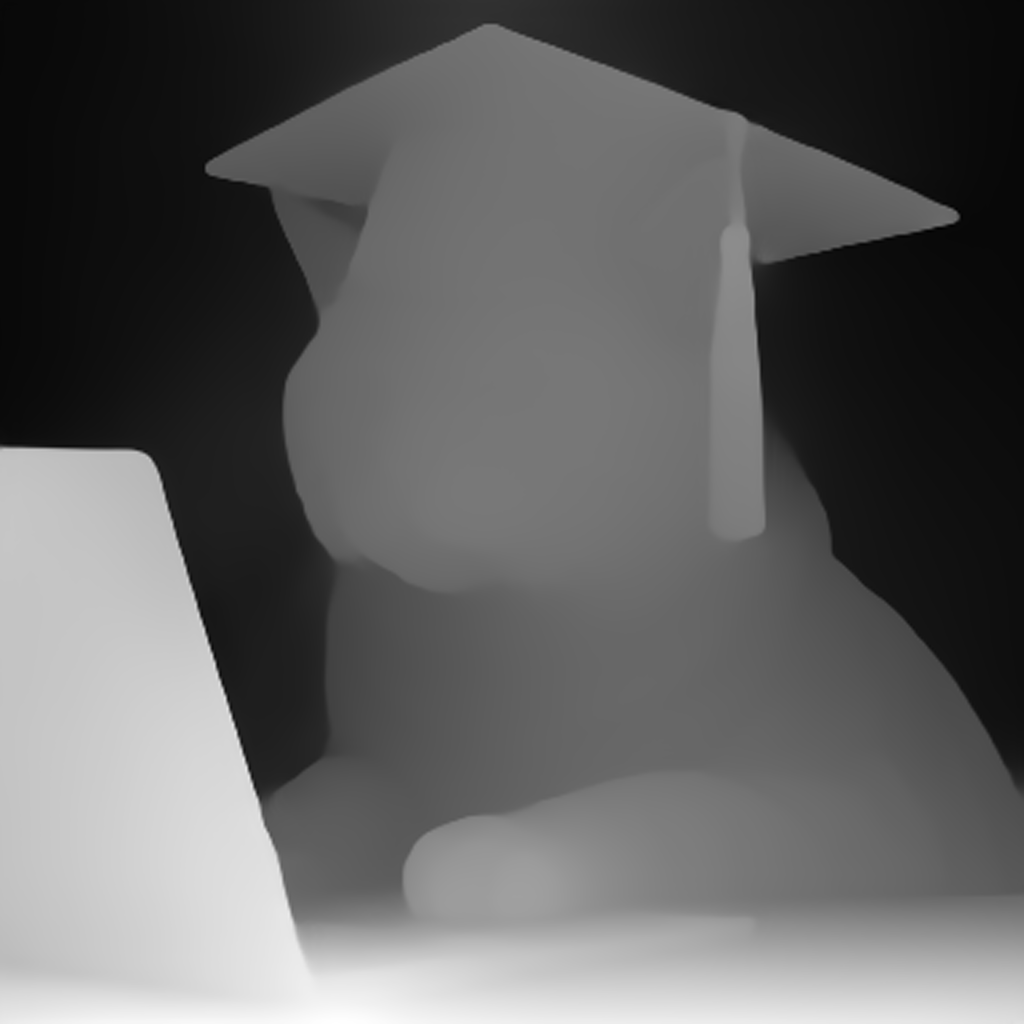} & \includegraphics[width=0.16\linewidth]{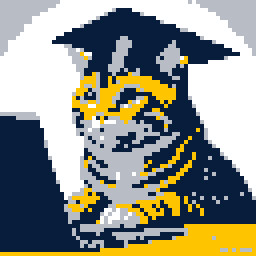} & \includegraphics[width=0.16\linewidth]{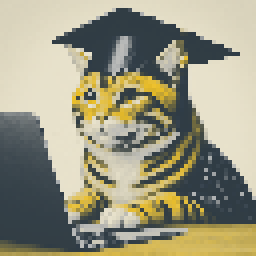} & 
    \includegraphics[width=0.16\linewidth]{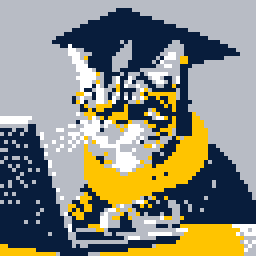}
        \\
         input & Canny edge & depth map prediction & initialization (argmax) & initialization (softmax) & result
    \end{tabular}
    \vspace{-0.25cm}
    \caption{Initial setup for our parameter comparisons. The sequence includes the original input image, its corresponding Canny edge detection \cite{cannyedge}, and depth map prediction via DPT \cite{DPTF2021Ranftl}. We also show the initial state of the generator, using the softmax- and argmax-generation modes, and the result using our default parameters. The used prompt is ``A cat wearing a graduation hat using a computer.''}
    \label{fig:inputAbl}
    \Description{Teaser.}
\end{teaserfigure}

\maketitle
\input{sections/supplementary/1_implementation}
\input{sections/supplementary/2_ablation}
\input{sections/supplementary/3_evaluation}
\input{sections/supplementary/4_additional_results}
\clearpage
\input{sections/supplementary/controlnetWeightsFigure}
\input{sections/supplementary/bigfigure_quantitative}
\input{sections/supplementary/userstudy_image}

\clearpage
\balance
\bibliographystyle{ACM-Reference-Format}
\bibliography{references}

%% file: macros.tex
\usepackage{bbm}
\usepackage{dsfont}

\newcommand{\ourmethod}{SD-$\pi$XL}

\newcommand{\figref}[1]{Fig.\ \ref{#1}}
\newcommand{\secref}[1]{Sec.\ \ref{#1}}

\newcommand{\tableref}[1]{Table \ref{#1}}

\usepackage{multirow}
\usepackage{pifont}
\usepackage{nicefrac}
\usepackage{colortbl}
\usepackage{xcolor}
\newcommand{\cmark}{\ding{51}}%
\newcommand{\xmark}{\ding{55}}%

%% file: sections/0_abstract.tex
Low-resolution quantized imagery, such as pixel art, is seeing a revival in modern applications ranging from video game graphics to digital design and fabrication, where creativity is often bound by a limited palette of elemental units. Despite their growing popularity, the automated generation of quantized images from raw inputs remains a significant challenge, often necessitating intensive manual input. We introduce \ourmethod{}, an approach for producing quantized images that employs score distillation sampling in conjunction with a differentiable image generator.
Our method enables users to input a prompt and optionally an image for spatial conditioning, set any desired output size $H \times W$, and choose a palette of $n$ colors or elements. Each color corresponds to a distinct class for our generator, which operates on an $H \times W \times n$ tensor. We adopt a softmax approach, computing a convex sum of elements, thus rendering the process differentiable and amenable to backpropagation. We show that employing Gumbel-softmax reparameterization allows for crisp pixel art effects.
Unique to our method is the ability to transform input images into low-resolution, quantized versions while retaining their key semantic features. Our experiments validate \ourmethod{}'s performance in creating visually pleasing and faithful representations, consistently outperforming the current state-of-the-art. Furthermore, we showcase \ourmethod{}'s practical utility in fabrication through its applications in interlocking brick mosaic, beading and embroidery design.

%% file: sections/1_introduction.tex
\section{Introduction}

Pixel art is a common form of low-resolution, quantized images, characterized by its minimalist aesthetic and distinctive use of color. Each pixel is clearly visible, and even a single pixel modification can have a significant perceptual impact. This art style has gained widespread popularity in various applications, such as video games and contemporary artistic design. Its charm lies not only in its visual appeal but also in its historical significance, as it evokes the early days of video games, when hardware limitations necessitated the use of simple, low-dimensional representations with a restricted amount of colors. Pixel art continues to be employed in numerous indie games and artistic creations, capitalizing on its unique visual style and lower memory footprint.

\begin{figure}[t]
	\centering
	\footnotesize
	\setlength{\tabcolsep}{1pt}
	\begin{tabular}{cccc}%
    \includegraphics[height=0.245\linewidth]{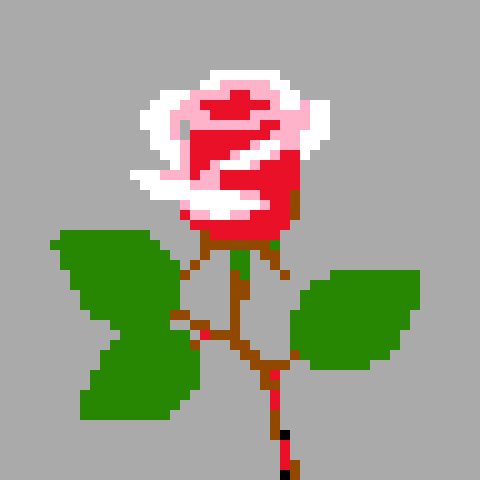}
     &
    \includegraphics[trim=0 0 0 0, clip, height=0.245\linewidth]{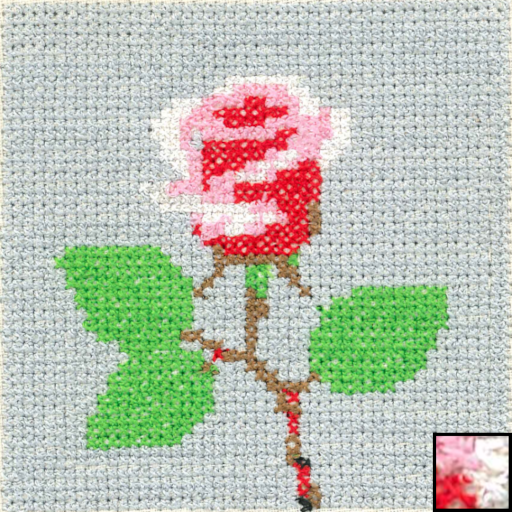} &
    \includegraphics[trim=0 0 0 0, clip, height=0.245\linewidth]{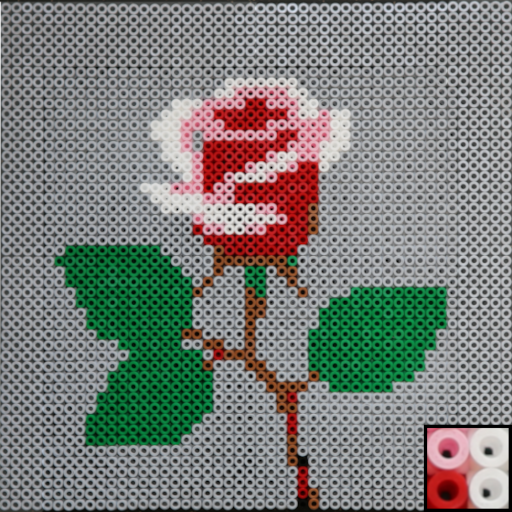} &
    \includegraphics[trim=0 0 0 0, clip, height=0.245\linewidth]{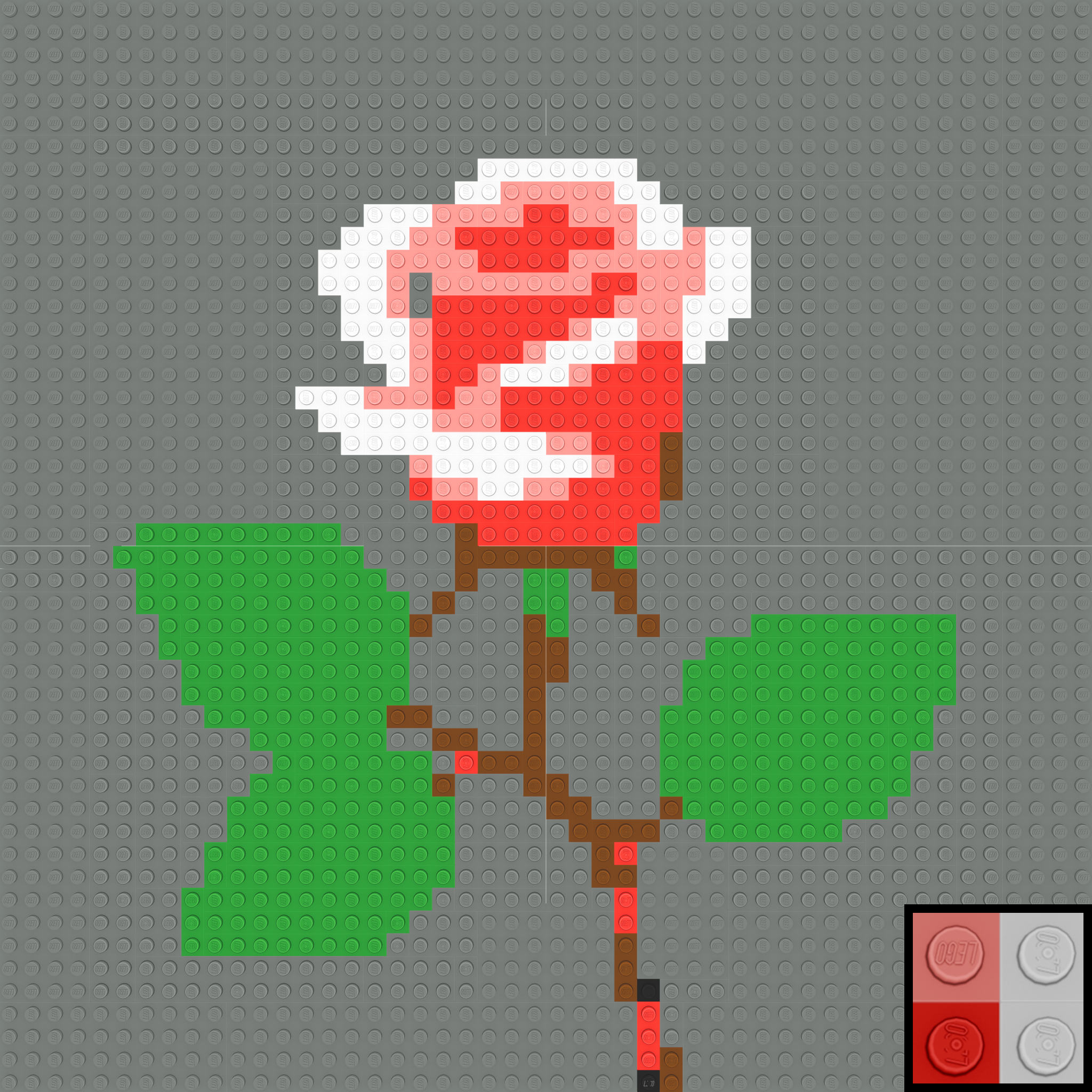} \\
    \ourmethod{} & Embroidery & Fuse beads & Interlocking bricks
	\end{tabular}
 \vspace{-0.2cm}
	\caption{\ourmethod{} generates low-resolution quantized images that are suitable for many fabrication applications, such as cross-stitch embroidery, fuse beads, or interlocking brick designs. The result image size is $48 \times 48$ pixels, generated without an initialization image, and only conditioned on the prompt ``A rose flower. The branch and leaves are visible.''}
	\label{fig:fabrication}
\vspace{-0.5cm}
\Description{Fabrication of a pixel art rose: embroidery, fuse beads and interlocking bricks.}
\end{figure}

As illustrated in \figref{fig:fabrication}, quantized images can reflect essential fabrication constraints or rationalization e.g.\, for embroidery \cite{Igarashi2022Handicraft} or interlocking brick games \cite{Zhou2023Lego}, where the production is constrained by a finite (usually small) amount of thread or brick colors. Creating pixel art from input images is a complex task, often requiring laborious manual effort. The challenges are compounded by the scarcity of suitable large, open datasets. Some common data augmentation techniques, such as rotation, color jitter, or blurring, may produce undesirable artifacts for pixel art style, worsening the dataset limitations.
Due to its fabrication opportunities, a pixel art generation method should respect the following properties:

\begin{enumerate}
    \item \emph{Hard constraints}: strict adherence to predefined constraints, such as input color palettes.
    \item \emph{Resolution independence}: ability to produce crisp images of various resolutions without anti-aliasing.
    \item \emph{Flexible generation and conditioning}: ability to base generation on an input prompt or image, with adjustable semantic and geometric conditioning.
    \item \emph{Style independence}: adaptability to different styles, such as realistic input to embroidery output as shown in \figref{fig:teaser}.
\end{enumerate}

\begin{figure}[t]
	\centering
	\small
	\setlength{\tabcolsep}{1pt}
	\begin{tabular}{cccc}%
    \includegraphics[width=0.245\linewidth]{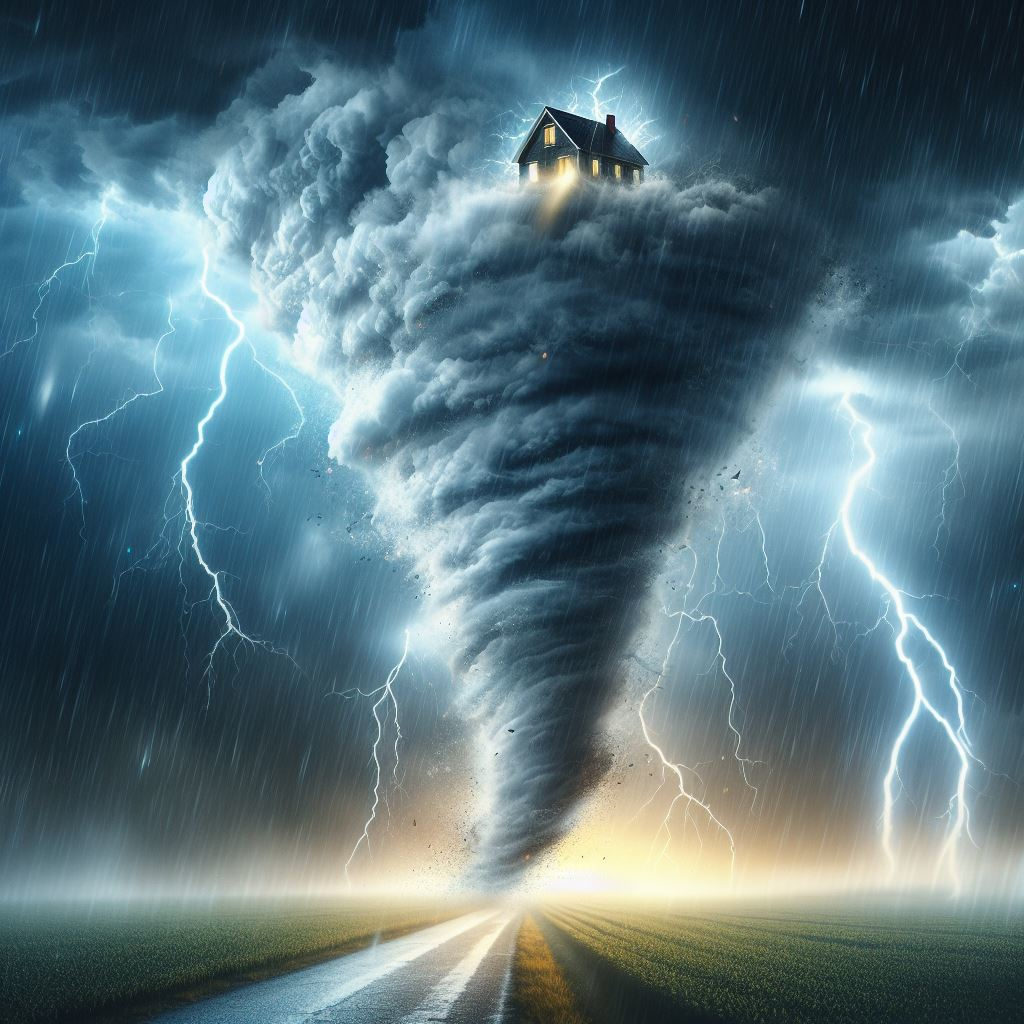}\ &
    \includegraphics[width=0.245\linewidth]{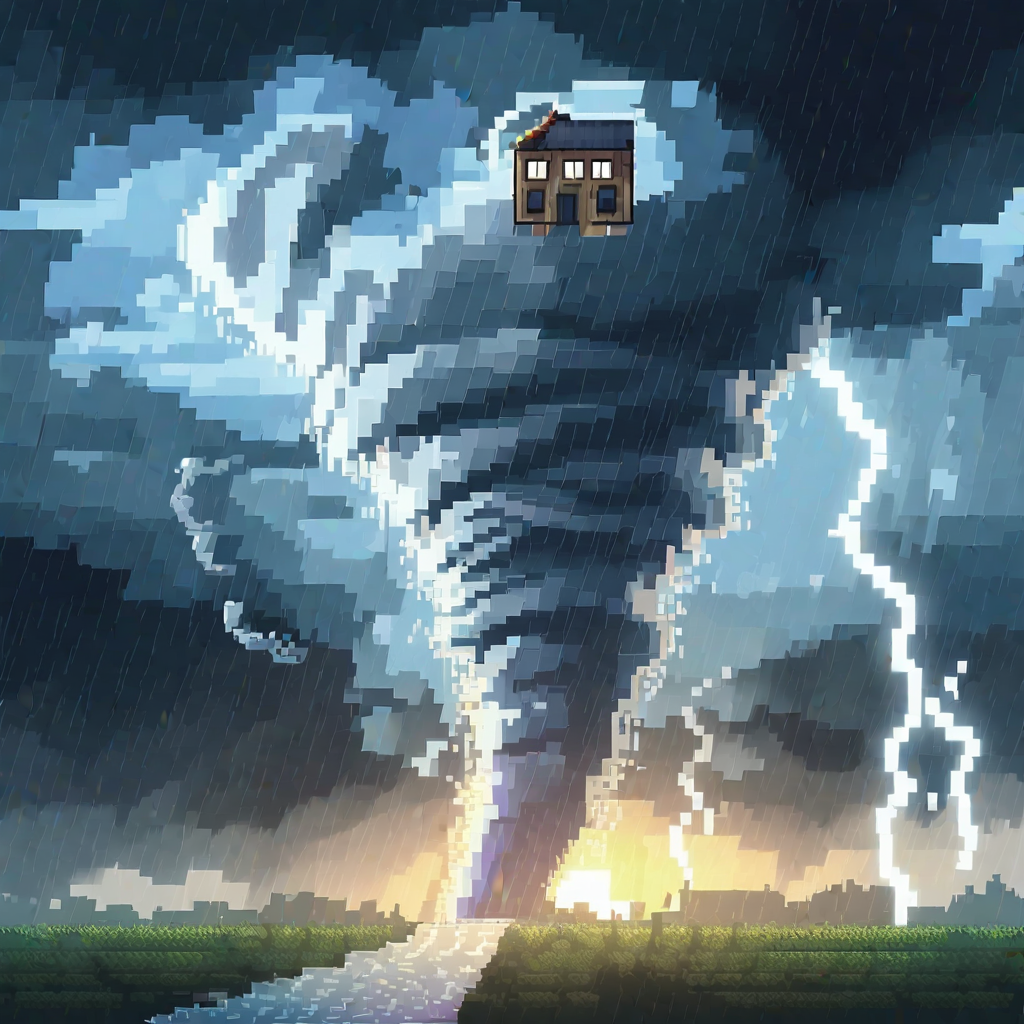}\ &
    \includegraphics[trim=0 0 0 0, clip, width=0.245\linewidth]{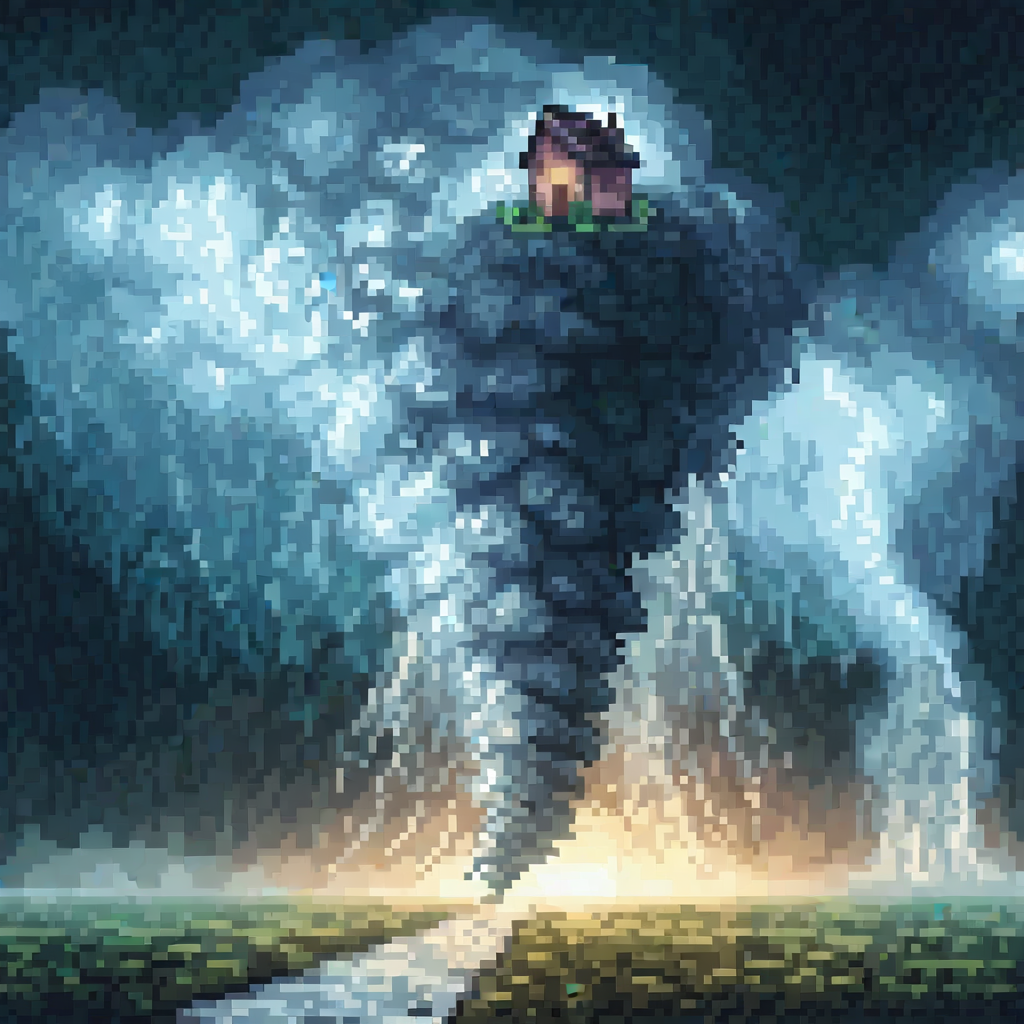}\  &
    \includegraphics[trim=0 0 0 0, clip, width=0.245\linewidth]{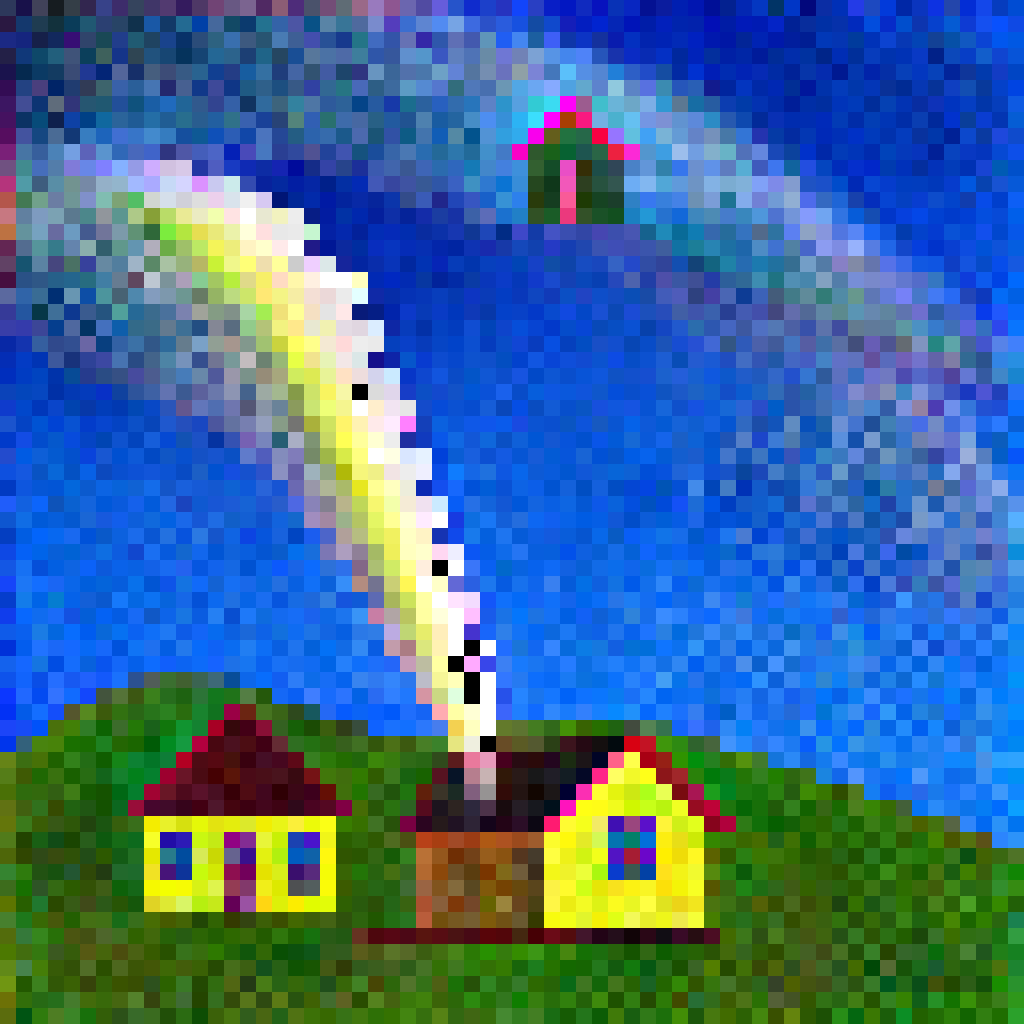}\ \\
    (1) & (2) & (3) & (4) \\
    \includegraphics[width=0.245\linewidth]{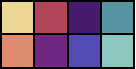}
     &
    \includegraphics[trim=0 0 0 0, clip, width=0.245\linewidth]{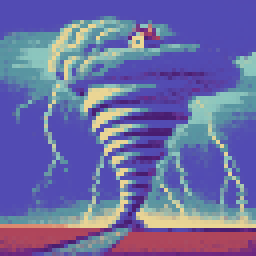} &
    \includegraphics[trim=0 0 0 0, clip, width=0.245\linewidth]{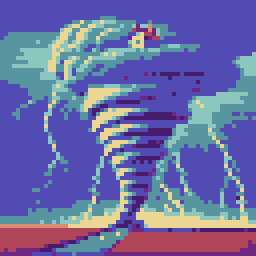} &
    \includegraphics[trim=0 0 0 0, clip, width=0.245\linewidth]{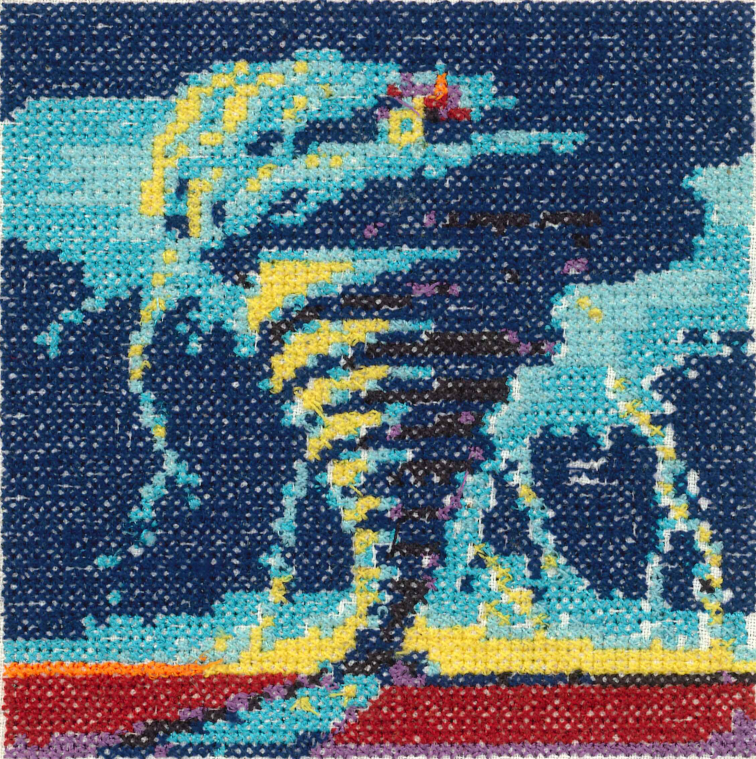} \\
    (5) & (6) & (7) & (8)
	\end{tabular}
 \vspace{-0.3cm}
	\caption{Diffusion models allow for the generation of high-resolution images (1). While using a diffusion-based image translation \cite{saharia2022palette, podell2023sdxl} with prompt-guided style is ineffective (2), fine-tuning the model for pixelized effects \cite{nerijspixelartxl2023} (3) is not generalizable across styles and requires retraining for different resolutions. VectorFusion \cite{jain2022vectorfusion} solves the resolution issue, but does not follow closely the input image (4). Our method supports outputs in any size and applies constraints to a finite palette (5), which can be enforced through either soft (6) or hard constraints (7). Color quantization further emphasizes the pixel art effect and is crucial for some fabrication applications, such as embroidery (8). 
 }
	\label{fig:introdiffusionComparison}
\vspace{-0.3cm}
 \Description{Comparison between methods for pixelization.}
\end{figure}

As detailed in \tableref{tab:methodComparison}, current methods do not fully satisfy the established criteria. Existing classic and neural pixelization techniques fall short in semantic conditioning, which is crucial for pixel art to effectively communicate at low resolutions,  
and no method strictly adheres to specific color palettes.
\figref{fig:introdiffusionComparison} shows limitations of current diffusion methods, as they cannot enforce strict color palette and resolution constraints, whether through prompt engineering, low-rank adaptation (LoRA) fine-tuning \cite{hu2021lora}, or existing score distillation approaches \cite{poole2022dreamfusion, jain2022vectorfusion}.

Our paper introduces \ourmethod{}, a method that leverages pretrained diffusion models to generate low-resolution, quantized images within specific constraints. 
\ourmethod{} offers a versatile approach: users can input a collection of visual elements (color palettes for pixel art or sets of images for mosaics), a prompt, and optionally, an image. To create an output image of size $H \times W$ using a palette of $n$ elements, we parameterize an image generator with a tensor of dimensions $H \times W\times n$. This tensor encodes the significance of each element at every pixel position. 
We use Gumbel-softmax reparameterization (\secref{sec:gumbel}) to sample elements from the palette, leading to a stochastic optimization process that efficiently produces crisp pixel art while still allowing for backpropagation.
We then employ diffusion networks with score distillation sampling for optimizing the parameters of the generator based on the input prompt, offering semantic understanding to the pixelization process.
We also integrate spatial fidelity to the input image through conditioning on depth maps and edge detection via ControlNet \cite{zhang2023addingControlNet}. 
Because our approach optimizes within a predefined constraint set, adherence to the input palette is guaranteed. 
Our main contributions are:
\begin{enumerate}
\item A differentiable image generator that strictly adheres to given constraints and works at any resolution.
\item Evidence showing that stochastic optimization via Gumbel-softmax reparameterization produces sharp, crisp pixel art.
\item Versatile generation capabilities from text or images, including semantic and spatial conditioning.
\item Elimination of dataset dependency via an optimization-based method that works with any input style.
\item State-of-the-art results in quantized image generation.
\end{enumerate}
Through our experiments, we demonstrate \ourmethod{}'s effectiveness in creating visually pleasing and accurate pixel art, surpassing existing methods. We also discuss the limitations of our approach and its potential for future work. Our supplementary material further includes ablation studies and details of our comparative evaluations. The source code is made available at \url{https://github.com/AlexandreBinninger/SD-piXL}.

%% file: sections/2_related_work.tex
\section{Related work} \label{sec:related_work}

In this work, we focus on generating imagery with a highly restricted number of pixels and colors. This task, requiring both semantic understanding and abstraction, is closely related to established research in color quantization and image downsampling. We review key studies in these areas to provide context and background for our approach, and refer to the survey by \citet{kumarComprehensive2019} for a comprehensive overview of the large topic of image abstraction.

\paragraph{Image quantization.} Since \ourmethod{} utilizes image palettes with a finite discrete set of elements and operates at low spatial resolution, we discuss relevant research in the fields of  color quantization and image downscaling.
Content-adaptive image downscaling \cite{Kopf2013ContentAdaptDownscaling} optimizes the shapes and locations of downsampling kernels to align with local image features, resulting in crisper output without ringing artifacts and effectively creating pixel art from vector graphics inputs.
Perceptually based image downscaling \cite{2015OtzireliPercepually} introduces an optimization method for image downscaling that retains perceptually important features.
Color manipulation is a well-studied field, often relying on layer decomposition \cite{aksoy2017unmixing} and manipulation via geometric tools like convex hull \cite{tan2016decomposing} or non-linear triads \cite{shugrina2020colortriads}.
Colour quantization often relies on the use of a color palette. Dynamic closest color warping \cite{kim2021dynamicsortpalette} assesses color palette similarity by sorting and aligning colors to share a common color tendency.
Floyd-Steinberg dithering \cite{Floyd:1976:AASDithering} is an error-diffusion method that minimizes color quantization artifacts. It distributes each pixel's quantization error to adjacent pixels, creating smoother images with a defined color palette.
\citet{ozturkhancerkaraboga2014} present a brief review of color quantization and propose a method based on the artificial bee colony algorithm.
Several quantization algorithms propose to abstract the input image with a non-grid clustering, to produce for instance mosaic effects \cite{2005MosaicFaustino}, low-polygon art \cite{2018Pic2Geom}, or posterization \cite{chao2021posterchild}. Superpixels are groups of connected pixels that share similar characteristics, such as color or texture, forming a coherent region within an image \cite{RenMalik:Superpixels:2003}. They can be used to segment the target image prior to clustering the color space for color quantization \cite{2022FrackiewiczMariusz}.

\begin{table}[b!]
    \centering
    \footnotesize
\setlength{\tabcolsep}{2pt}

    \caption{Comparison of pixelization techniques. Unlike other methods, ours allows users to enforce hard constraints on resolution and palette without additional post-processing. Classical methods provide flexibility across various resolutions or scales, whereas neural methods are typically limited to a finite set of resolutions or downscaling factors. Non-diffusion deep learning methods, albeit trainable or fine-tunable for different styles, often heavily rely on their training datasets due to a lack of semantic conditioning. VectorFusion \cite{jain2022vectorfusion} also relies on score distillation (SD) \cite{poole2022dreamfusion} to optimize the parameters of a differentiable image generator, but does not constrain the image generation to an input palette.}
    \begin{tabular}{cl@{\hspace{0pt}}cccc}
        \toprule
        & & Hard & Resolution & Semantic & Style \\
         & & constraints & independence & conditioning & flexibility \\ \midrule
        \multirow{3}{*}{\rotatebox{90}{\scriptsize classic}} & PIA \cite{Gerstner:2012:PIA} & \cmark & \cmark & \xmark &  \cmark \\
        & APIP \cite{2021YunyiPortraitPixelization} & \xmark & \cmark & \xmark & \xmark \\
        & AOP \cite{Peng2023AOP} & \xmark & \cmark & \xmark & \xmark \\
        \multirow{2}{*}{\rotatebox{90}{\scriptsize neural}} & DUP \cite{2018ChuDeepUnsupervizedPixelization} & \xmark & finite & \xmark & dataset \\
        & MYOS \cite{Wu2022MakeYourOwnSprite} & \xmark & finite & \xmark & dataset \\
        \multirow{2}{*}{\rotatebox{90}{\scriptsize SD}} & VectorFusion \cite{jain2022vectorfusion} & \xmark & \cmark & \cmark &  \cmark \\
        & \textbf{\ourmethod{}} & \cmark & \cmark & \cmark & \cmark \\
        \bottomrule
    \end{tabular}
    \label{tab:methodComparison}
\end{table}

\vspace{-2mm}
\paragraph{Classic pixelization methods.} Pixelated image abstraction \cite{Gerstner:2012:PIA, GERSTNER2013WithConstraints} also relies on superpixels with a modified version of simple linear iterative clustering (SLIC) \cite{2012AchantaSLIC} to generate pixel art-style images by simultaneously solving for feature mapping and a reduced color palette. While faithful to the input image, it lacks a semantics-aware mechanism.
Automatic portrait image pixelization \cite{2021YunyiPortraitPixelization} also relies on SLIC to introduce a pixelization algorithm for portrait images.
The art-oriented pixelation (AOP) method \cite{Peng2023AOP} converts cartoon images into pixel art through an iterative procedure involving gridding the image, extracting its content, and separately pixelating the contour and non-contour parts of the image. \citet{kuo2016} develop a method to animate pixel art by optimizing feature lines on each frame.
Vector graphics is also present in the context of pixel-art creation. \citet{Inglis2012} devise a pixelation algorithm for rasterizing vector line art while maintaining pixel art conventions. Conversely, \citet{Kopf2011} address the problem of \emph{depixelation} in generating vector representations from pixel art images by resolving pixel-scale feature ambiguities to produce smooth, connected features. This research has led to further works about pixel art depixelation via vectorization \cite{hoshyari2018vectorization,dominici2020vectorizationPolyfit,DepixelMatusovic2023}.

\vspace{-2mm}
\paragraph{Neural pixelization methods.} Neural techniques to generate pixelized images are not new. 
Current neural techniques for domain transfer often use unsupervised methods like CycleGAN \cite{CycleGAN2017}. These rely on generative adversarial networks (GANs) \cite{goodfellow2014generative} to transform images between different style domains.
Deep unsupervised pixelization \cite{2018ChuDeepUnsupervizedPixelization} generates pixel art without paired training data by using several networks dedicated to different tasks, namely transforming the input image into grid-structured images, generating pixel art with sharp edges, and recovering back the original image from the pixelized result for cyclic consistency.
\citet{CycleGanPixelArtKuang2021} present a pixel image generation algorithm based on CycleGAN, utilizing a nested U-Net generator structure for multi-scale feature fusion, and introducing a structure combination loss to ensure the integrity of linear structures like contours in pixel images. The \emph{Make Your Own Sprites} method  \cite{Wu2022MakeYourOwnSprite} produces cell-controllable pixel art by using a reference pixel art for regularizing the cell structure, and disentangling the pixelization process into cell-aware and aliasing-aware stages.
\citet{YUVPixelArt} also propose a GAN-based model for pixel art generation using the YUV color encoding system.

Generating sprites is an important aspect of pixel art creation, e.g.\ for game assets \cite{2021KarpGameAssetGAN}. \citet{2019AssetGenSerpa} use deep neural networks to generate pixel art sprites from line art sketches. Their work is based on Pix2Pix \cite{pix2pix2017}, a general method that translates an image to a different domain. 
Also based on Pix2Pix, GAN-based sprites generation \cite{coutinho2022generating} expedites the process of creating pixel art character sprite sheets by generating target side poses based on source poses. Subsequently, \citet{CoutinhoChaimowicz2022} propose two modifications, namely a color palette representation and a histogram loss, and discuss the difficulties of pixel-art sprite generation using GANs.
These neural methods take the stance of considering pixelization as a domain transfer problem, while we incorporate semantic conditioning for low-resolution, style-agnostic generation. This adaptability allows \ourmethod{} to be effective across various styles and applications.

VectorFusion \cite{jain2022vectorfusion}, and concurrently to our work, SVGDreamer \cite{xing2024svgdreamer}, leverage a diffusion model for semantics-aware optimization of the parameters of a differentiable vector rasterizer \cite{Li:2020:DVG} via score distillation sampling \cite{poole2022dreamfusion}. They can force the generation to a grid, producing low-resolution images, but the lack of color quantization makes their results saturated and noisy. Prior to score distillation, some methods used CLIP \cite{radford2021learning} for image abstraction, such as CLIPDraw \cite{frans2021clipdraw} or CLIPasso \cite{vinker2022clipasso}.

\vspace{-2mm}
\paragraph{Fabrication with quantized images.} Low-resolution and color-quantized images have various fabrication applications. Embroidery is limited by the number of thread colors. While image conversion methods exist for directionality-aware embroidery patterns \cite{LiuEmbroidery2023}, low-resolution pixel art is particularly adapted for cross-stitching. Though cross-stitching can be automatically performed by modern sewing machines, e.g.\ \cite{Pfaff20}, techniques
to correct human mistakes on-the-fly for pixel art fabrication have been developed \cite{Igarashi2022Handicraft}. Fuse beads is a popular form of pixel art fabrication, and is de facto limited by the available bead colors. Interlocking bricks such as LEGO\textsuperscript{\textregistered} are another suitable fabrication possibility. While advancements have been made in the realm of 3D LEGO\textsuperscript{\textregistered} design methodologies \cite{Xu2019Lego}, efforts are actively made to explore the design of 2D brick-based structures as well \cite{Zhou2023Lego}.

%% file: sections/3_background.tex
\begin{figure*}[!ht]
    \center
    \includegraphics[trim= 521 51 750 140, clip, width=0.9\linewidth]{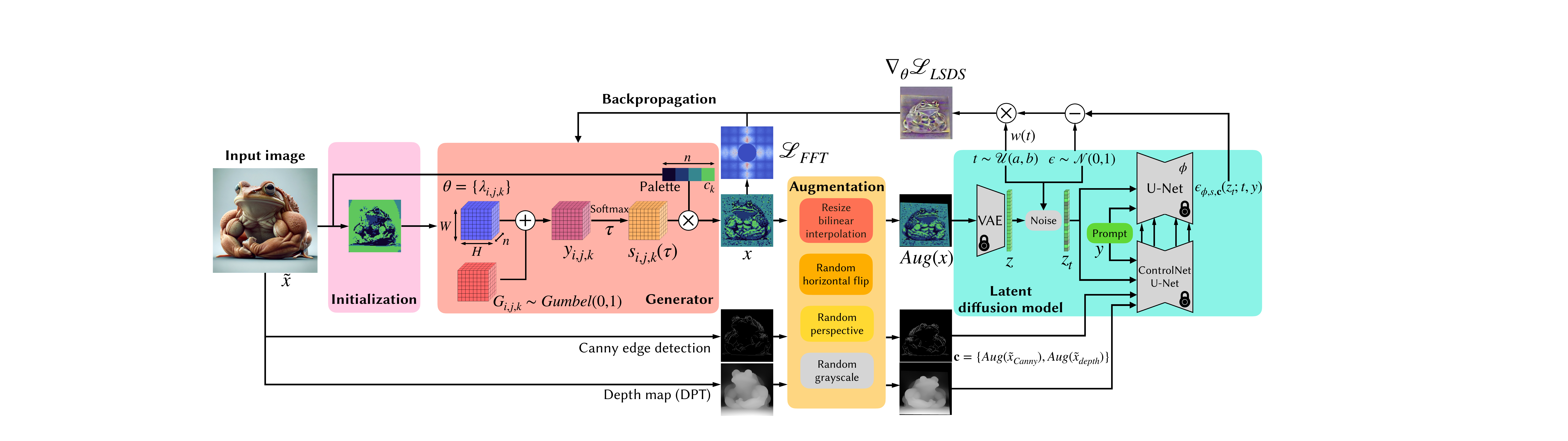}
    \caption{Visualization of the optimization process for generating a pixelized $H \times W$ image with a color palette of size $n$. If an input image is provided, the process starts with {\color{pink}initializing} the {\color{blue}logits $\lambda_{i, j, k}$} by downsampling the input image and matching each pixel to the nearest palette color. Otherwise, the logits are randomly initialized. Next, {\color{red}Gumbel-distributed random variables $G_{i, j, k}$} are {\color{purple}added} to the logits. Applying a {\color{orange}softmax function} and combining the palette colors weighted by {\color{orange}$s_{i, j, k}(\tau)$} yields an output image $x$. This $x$, the Canny edge map \cite{cannyedge} and an estimated depth map \cite{DPTF2021Ranftl} of the input image are then {\color{yellow}augmented} and used in a {\color{cyan}latent diffusion model} \cite{podell2023sdxl} to compute a semantic loss $\nabla_\theta \mathcal{L}_\mathit{LSDS}$, conditioned on an {\color{green} input prompt $y$}. Additionally, a smoothness loss $\mathcal{L}_\mathit{FFT}$ derived from $x$ is used to optimize the parameters $\theta$.}
    \label{fig:methodSummary}
    \Description{Description of our method.}
\end{figure*}

\section{Background}

\subsection{Diffusion}

We briefly review diffusion models, referring the reader to a comprehensive survey for more in-depth explanation \cite{po2023state}. Diffusion models are a family of generative models that map Gaussian noise into samples from a targeted image distribution $p_{\text{data}}$  \cite{sohldickstein2015deep,ho2020denoising}. They consist of two main stages. The first is the forward process: an initial sample $x_0 \sim p_{\text{data}}$ undergoes a progressive noising over $T$ steps, culminating in a Gaussian-distributed sample $x_T \sim \mathcal{N}(0, \sigma_T)$. To avoid exploding variance \cite{song2021scorebased}, the noisy sample is computed as $x_t = \alpha_t x_0 + \epsilon \sigma_t$, where $\epsilon \sim \mathcal{N}(0, 1)$, $t \in \{0, ..., T\}$ is the time step, $\alpha_t$ and $\sigma_t$ parameterize the diffusion \cite{kingma2023variational}. Following the initial phase is the backward process: it begins with a noisy sample $x_t$ and successively estimates the noise to progressively generate cleaner samples $x_{t-1}$. This iterative denoising continues until it reconstructs the final image $x_0$, which closely resembles the original data distribution $p_{\text{data}}$. Typically, this denoising function is implemented using a U-Net architecture \cite{ronneberger2015unet}, denoted as $\epsilon_\phi(x_t; t)$. This function specifically aims to deduce the noise $\epsilon$ that was initially mixed with the original data $x_0$ to create the noisy version $x_t$. 

\vspace{-2mm}
\paragraph{Conditioning in diffusion models} The denoising process can be conditioned by a parameter $y$, for instance with text for prompt-based image generation.
To generate samples aligned with a specific condition $y$, diffusion models utilize classifier-free guidance (CFG) \cite{ho2022classifierfree}. CFG modifies the conditioned prediction $\epsilon_\phi(x_t; y, t)$ away from the unconditioned prediction $\epsilon_\phi(x_t; \emptyset, t)$, with scaling $s \in \mathds{R}$ modulating the intensity of the conditioning: \begin{equation*}
\epsilon_{s, \phi}(x_t; y, t) = \epsilon_\phi(x_t; y, t) \ + \ s \,(\epsilon_\phi(x_t; y, t) - \epsilon_\phi(x_t; \emptyset, t)).
\end{equation*}

\subsection{Score distillation}

Score distillation employs pretrained diffusion models to compute semantics-aware gradients for updating the parameters of a differentiable renderer or generator \cite{poole2022dreamfusion}. Denote $g$ a differentiable image generator with parameters $\theta$, and $x = g(\theta)$ a generated image. For a given time step $t$, a noised version of $x$ is defined as $x_t = \alpha_t x + \epsilon \sigma_t$, with $\epsilon \sim \mathcal{N}(0, 1)$. The gradient of the score distillation sampling (SDS) loss is described by the equation \begin{equation} \label{eq:sds}
    \nabla_{\theta} \mathcal{L}_\mathit{SDS} = \mathbb{E}_{t, \epsilon}\left[w(t) \left( \epsilon_{s, \phi}(x_t; y, t) - \epsilon \right) \frac{\partial x}{\partial \theta}\right],
\end{equation}
where $w(t) = \sigma_t^2$ serves as a scaling factor. This gradient is subsequently used to refine the parameters of the generator $g(\theta)$.
Although initially developed for 3D generation, the application of score distillation extends beyond 3D. Given that an image generator is differentiable, score distillation can be used for semantics-based optimization, such as prompt-based image editing \cite{hertz2023deltaDDS}. Its utility is also evident in various other forms of image representation, such as vector graphics \cite{jain2022vectorfusion}, font design \cite{iluz2023wordasimage}, or tiling \cite{aigerman2023generative}.

\subsection{Gumbel reparameterization}
\label{sec:gumbel}

The Gumbel reparameterization technique utilizes the Gumbel distribution \cite{GumbelEmilJulius1954Stoe} for sampling from a categorical distribution using its logits. Its impact is analyzed in \secref{sec:gumbelanalysis}, and this section explains its operation. Consider a set of $n$ scalars \( (\lambda_0, ..., \lambda_{n-1})\) which represent the logits of a categorical probability distribution \( \mathit{Cat}(\pi_0, \ldots, \pi_{n-1}) \), where the probability of selecting the \( k \)-th category is determined by the softmax operation \( \pi_k = {e^{\lambda_k}}/{\sum_{l=0}^{n-1}{e^{\lambda_l}}} \). Let \(\{G_k\}_{0\leq k < n}\) be a series of $n$ independent random variables, each sampled from a Gumbel distribution \( \mathit{Gumbel}(0, 1) \), and let $y_k = \lambda_k + G_k$ for $0\leq k < n$. The random variable \( Y \coloneq \mathop{\mathrm{argmax}}_{0\leq k < n}\{ y_k\} \)
is then distributed according to \( \mathit{Cat}(\pi_0, \ldots, \pi_{n-1}) \).
The \emph{Gumbel-Softmax} reparameterization technique offers a way to perform stochastic sampling from categorical distributions while remaining amenable to backpropagation \cite{jang2017categoricalGumbel, maddison2017concreteGumbel}. This method utilizes a softmax function that is parameterized by a temperature scalar \( \tau \). Given \( n \) categories \(\{c_k\}_{0 \leq k < n}\) and the objective of sampling from the categorical distribution \( \mathit{Cat}(\pi_0, \ldots, \pi_{n-1}) \), the softmax function for each category is defined as 
\( s_{k}(\tau) = {e^{\frac{y_k}{\tau}}}/{\sum_{l=0}^{n-1}{e^{\frac{y_l}{\tau}}}} \), 
where $y_k$ are the logits modified by Gumbel noise. The sampling process of a category is then realized by  \( c_\tau = \sum_{k=0}^{n-1}{s_{k}(\tau) c_k} \). The parameter \( \tau \) modulates how closely \( c_{\tau} \) approximates a categorical distribution. As \( \tau \) approaches zero, \( s_{k}(\tau) \) converges to an indicator function \( \mathrm{1}_{k =\mathop{\mathrm{argmax}}_{0 \leq l < n}\{ y_l\} } \), implying that for small \( \tau \), \( c_\tau \) closely resembles the categorical sampling \( \mathit{Cat}(\pi_0, \ldots, \pi_{n-1}) \) from the categories \(\{c_k\}_{0 \leq k < n}\). Conversely, as \( \tau \) increases towards infinity, \( s_{k}(\tau) \) approaches \( \frac{1}{n} \), meaning larger \( \tau \) values lead to \( c_{\tau} \) resembling a uniform average of the categories.

%% file: sections/4_method.tex
\section{Method}
\label{sec:method}

\ourmethod{} optimizes the parameters of a differentiable image generator by using SDXL \cite{podell2023sdxl}, a pre-trained latent diffusion model, denoted as $\epsilon_\phi$, to derive a semantics-aware loss. The method requires an input text prompt $y$ and can optionally take an input image $\tilde{x}$ to guide the diffusion process. The inclusion of a smoothness loss is also supported. Our method is illustrated in \figref{fig:methodSummary}.

\subsection{Stochastic quantized image generation}

\label{sec:quantizedImageGenerator}

In the proposed framework, the goal is to synthesize an image using only $n$ distinct colors from a finite set $\mathcal{C} = \{c_k\}_{0 \leq k < n}$. Although $\mathcal{C}$ is typically a color palette—equivalent to a collection of $n$ single-pixel images—it can also represent any set of elements that are uniform in size and can be rendered as image pixels, as shown in the mosaics in the supplementary material.
To generate an image $x$ of dimensions $(H, W)$ using colors from the palette $\mathcal{C}$, we employ a generator $g$, parameterized by $\theta = {\lambda_{i,j,k}} \in \mathds{R}^{H \times W \times n}$. The logits $\lambda_{i, j, k}$ give the probability that the pixel at position $(i, j)$ in $x$ will take the value $c_k$, computed as $$\pi_{i, j, k} = \frac{e^{\lambda_{i, j, k}}}{\sum_{l=0}^{n-1}{e^{\lambda_{i, j, l}}}}.$$
By definition, our generator is invariant to translation of $\theta$.
We take advantage of the Gumbel-softmax reparameterization (\secref{sec:gumbel}) and sample $HWn$ independent random variables $G_{i, j, k} \sim Gumbel(0, 1)$, and define $y_{i, j, k} \coloneq \lambda_{i, j, k} + G_{i, j, k}$. After performing a softmax 
$$s_{i, j, k}(\tau) = \frac{e^{\frac{1}{\tau}{y_{i,j,k}}}}{\sum_{l=0}^{n-1}{e^{\frac{1}{\tau}{y_{i,j,l}}}}},$$ 
the color of each pixel in $x$ is computed as $x_{i, j}(\tau) = \sum_{k=0}^{n-1}{s_{i, j, k}(\tau)\, c_k}$. Lower $\tau$ values enhance the resemblance of the sampling process to a categorical distribution, but excessively small $\tau$ leads to backpropagation instability. In practice, we find $\tau = 1$ to achieve reasonable results. Further insights and discussions on this choice are presented in the supplementary material.

\subsection{Input image conditioning}
\label{sec:inputImage}
\begin{figure}[t]
    \center
    \includegraphics[width=0.95\linewidth]{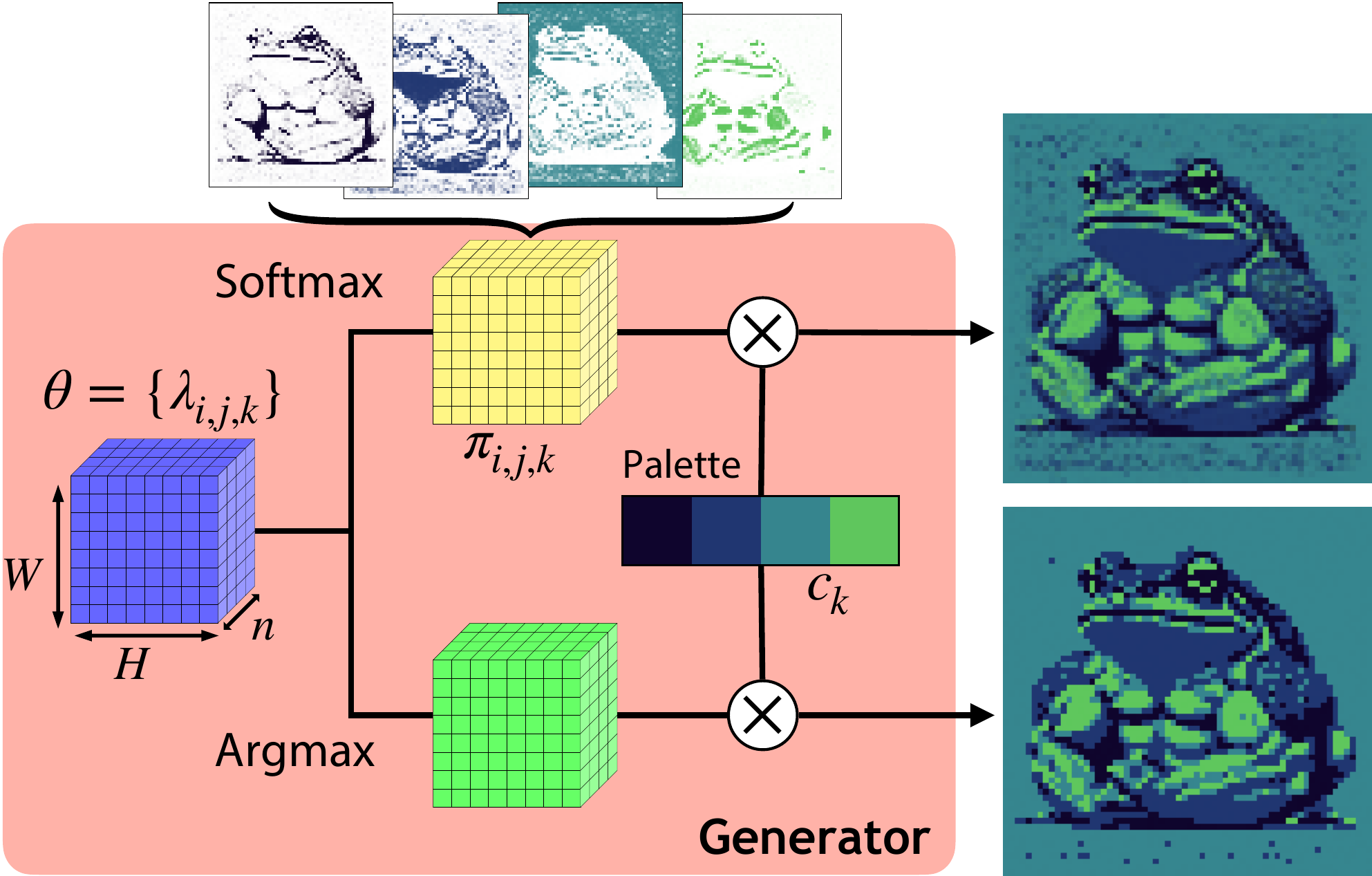}
    \caption{Our image generator can strictly adhere to the input palette using an argmax function ({bottom frog}). Using softmax yields an image whose pixel colors lie in the convex hull of the input palette, leading to less crisp, pixelized outputs ({top frog}).}
    \label{fig:generator}
    \Description{Difference between argmax and softmax generation.}
\end{figure}
\label{sec:initialization}
\ourmethod{} operates with a semantic loss, yet the optimization process can be enhanced by an input image for both initialization and spatial conditioning. Since we can use rejection sampling \cite{jain2022vectorfusion} to generate images from prompt $y$ and then select the best according to their CLIP score \cite{radford2021learning}, the content of this section also applies to text-only pixel art generation.
We {initialize} the generator with an image $\tilde{x}^d$, obtained by downsampling $\tilde{x}$ to size $(H, W)$ using bilinear interpolation. We set the initial values of $\theta$ to $\lambda_{i, j, k} = -\lVert \tilde{x}_{i, j}^d - c_{k}\rVert$.
If a color palette is not provided, we employ a K-means algorithm to partition the color space into $n$ clusters, using their centroids for the color palette $\mathcal{C}$.

ControlNet \cite{ChenControl3D2023} is a network architecture used to \textit{spatially condition} the diffusion process. In our approach, we employ ControlNet networks pretrained to condition the diffusion on edges and depth information. By applying Canny edge detection \cite{cannyedge} and the dense prediction transformer (DPT) \cite{DPTF2021Ranftl}, we condition the diffusion process on the structural and spatial characteristics of the input image $\tilde{x}$, providing the generation with spatial context.
We denote the denoiser conditioned by ControlNet on $\mathbf{c} = \{ \tilde{x}_{\mathit{Canny}}, \tilde{x}_{\mathit{depth}} \}$ as $\epsilon_{\phi, \mathbf{c}}(x_t; y, t)$. 
The impact of ControlNet conditioning is examined in \secref{sec:controlnet}.
\subsection{Image augmentation}

As illustrated in \figref{fig:methodSummary}, during our optimization, the current generated image $x$ and its associated conditioning images $\tilde{x}_{\mathit{Canny}}$ and $\tilde{x}_{\mathit{depth}}$ are fed to the diffusion model.
Prior to that, we apply data augmentation: The images are first resized to the target output dimensions of the diffusion model, and subsequent augmentations include random grayscale conversion, perspective alteration, and horizontal flipping. As the conditioning images spatially guide the denoising process, it is crucial that both the generated and conditioning images undergo identical augmentations.
To effectively utilize open-source latent diffusion models such as Stable Diffusion XL \cite{podell2023sdxl}, the augmented image is encoded, represented as $z = Enc(Aug(x))$. Subsequently, we denote $z_t = \alpha_t z + \sigma_t \epsilon$ the noise-altered version of $z$ at time step $t$.

\subsection{Loss function}

Adapting the score distillation sampling loss (Eq.\ \eqref{eq:sds}) for latent diffusion models, the {l}atent {s}core {d}istillation {s}ampling (LSDS) loss can be written as \cite{jain2022vectorfusion}:
\begin{equation}
\label{eq:lsds}
\nabla_{\theta} \mathcal{L}_\mathit{LSDS} = \mathbb{E}_{t, \epsilon, G}\left[w(t) \left( \epsilon_{s, \phi}(z_t; y, t) - \epsilon \right)  \frac{\partial z}{\partial x} \frac{\partial x}{\partial \theta}\right].
\end{equation}
In our case, the expected value also takes into account the Gumbel random variables $G = \{G_{i, j, k}\}$.
By decomposing $\epsilon_{s, \phi, \mathbf{c}}$, we find $$\epsilon_{s, \phi, \mathbf{c}}(z_t; y, t) - \epsilon = \underbrace{(\epsilon_{\phi, \mathbf{c}}(z_t; y, t) - \epsilon)}_{\text{variance-reduction}} + s \underbrace{(\epsilon_{\phi, \mathbf{c}}(z_t; y, t) - \epsilon_{\phi, \mathbf{c}}(z_t; \emptyset, t))}_{\text{semantic}}.$$
This brings a decomposition of the LSDS loss into two terms:
\begin{equation}
\label{eq:clsds}
\nabla_{\theta} \mathcal{L}_\mathit{LSDS} =\nabla_{\theta}  \mathcal{L}_\mathit{Noise} + s  \nabla_{\theta}  \mathcal{L}_\mathit{Sem},
\end{equation}
where
\begin{equation}
\begin{aligned}
\label{eq:lossTerms}
&\nabla_{\theta} \mathcal{L}_\mathit{Noise} = \mathbb{E}_{t, \epsilon, G}\left[w(t)  \left( \epsilon_{\phi, \mathbf{c}}(z_t; y, t) - \epsilon \right)  \frac{\partial z}{\partial x} \frac{\partial x}{\partial \theta}\right], \\
&\nabla_{\theta} \mathcal{L}_\mathit{Sem} = \mathbb{E}_{t, \epsilon, G}\left[w(t) \left( \epsilon_{\phi, \mathbf{c}}(z_t; y, t) - \epsilon_{\phi, \mathbf{c}}(z_t; \emptyset, t) \right)  \frac{\partial z}{\partial x} \frac{\partial x}{\partial \theta}\right].
\end{aligned}
\end{equation}
The noise-reduction loss component refines the parameters to yield a denoised image output, a desirable feature in contrast to its typically obstructive role in 3D generation. The semantic loss ensures that the generated result is in harmony with the provided prompt. A justification for this decomposition of the loss terms is elaborated in the supplementary material.

\ourmethod{}, being optimization-centric, allows for the integration of conventional loss functions. We introduce an additional fast Fourier transform (FFT) \cite{fastfouriertransform1967} loss to enhance smoothness. This involves calculating the FFT of the grayscale of $x$, centering it, masking out low frequencies with $M \in \mathds{R}^{H \times W}$, and averaging the absolute values:
\begin{equation}
\label{eq:fft}
    \mathcal{L}_\mathit{FFT} = \frac{\lVert \mathit{Shift}(\mathit{FFT}(x)) \odot M\rVert_1}{\lVert M\rVert_1}.
\end{equation}
Finally, the gradient of our loss can be written as:
\begin{align}
    \label{eq:totalLoss}
    \nabla_{\theta} \mathcal{L} = \nabla_{\theta}  \mathcal{L}_\mathit{Noise} + s  \nabla_{\theta}  \mathcal{L}_\mathit{Sem} + w_\mathit{FFT} \nabla_{\theta}\mathcal{L}_\mathit{FFT}.
\end{align}
In practice, we find $s = 40$ and $w_\mathit{FFT} = 20$ to yield effective results.

%% file: sections/5_results.tex
\section{Results}
This section outlines the final image generation process after optimization and justifies the adoption of the Gumbel-softmax reparameterization. We succinctly present the influence of ControlNet and the results of our comparative analysis, and refer the reader to the supplementary material for further details. We end with a discussion  of our method's limitations and future directions. 

\subsection{Final image generation}
\label{sec:imageGenFinal}

After optimization, our generator offers two image generation methods, shown in \figref{fig:generator}. The first option is \emph{argmax-generated} images, which respect hard constraints and strictly adhere to a color palette, 
$$x_{i, j} = c_{\tilde{k}_{i, j}}, \ \text{ where } \tilde{k}_{i, j} = \mathop{\mathrm{{argmax}}_{0 \leq k < n} \lambda_{i, j, k}}.$$ 
The second option is using $\pi_{i, j, k}$ as coefficients of a convex sum over the palette $\mathcal{C}$ to obtain \emph{softmax-generated} images, calculated as 
$$\textstyle x_{i, j} = \sum_{k=0}^{n-1}{\pi_{i, j, k} \, c_k}.$$ 
Their color space is merely constrained to the convex hull of the palette $\mathcal{C}$, softening the pixel art effect. We showcase in \figref{fig:bigfigureResultsRandomInit} the two generation methods. Note that softmax-generated images do not require Gumbel reparameterization during optimization, as explained in the following section.
\begin{figure}[t!]
\renewcommand{\arraystretch}{3.5}
	\centering
	\setlength{\tabcolsep}{1pt}
 \scriptsize
    \begin{tabular}{cccccc}
    & & argmax & softmax & normalized & average\\
\addlinespace[-18pt]
    & &generation & generation & entropy & normalized entropy  \\
\addlinespace[-5pt]
    \multirow{2}{*}{\rotatebox{90}{\scriptsize optimization without}} & \multirow{2}{*}{\rotatebox{90}{\scriptsize Gumbel-softmax}}&\multirow{2}{*}{\includegraphics[trim= 0 0 0 0, clip, width=.215\linewidth]{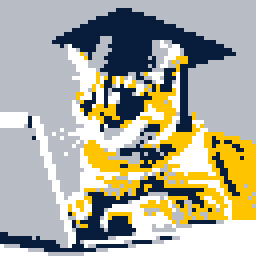}} & 
     \multirow{2}{*}{\includegraphics[trim= 0 0 0 0, clip, width=.215\linewidth]{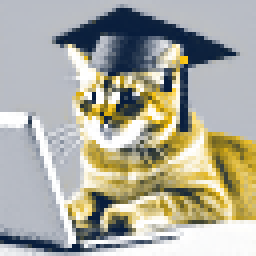}} & 
     \multirow{2}{*}{\includegraphics[trim= 0 0 0 0, clip, width=.215\linewidth]{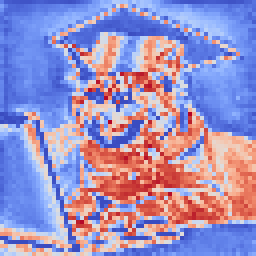}} & 
   \multirow{2}{*}{\includegraphics[trim= 0 5 0 5, clip, width=.28\linewidth]{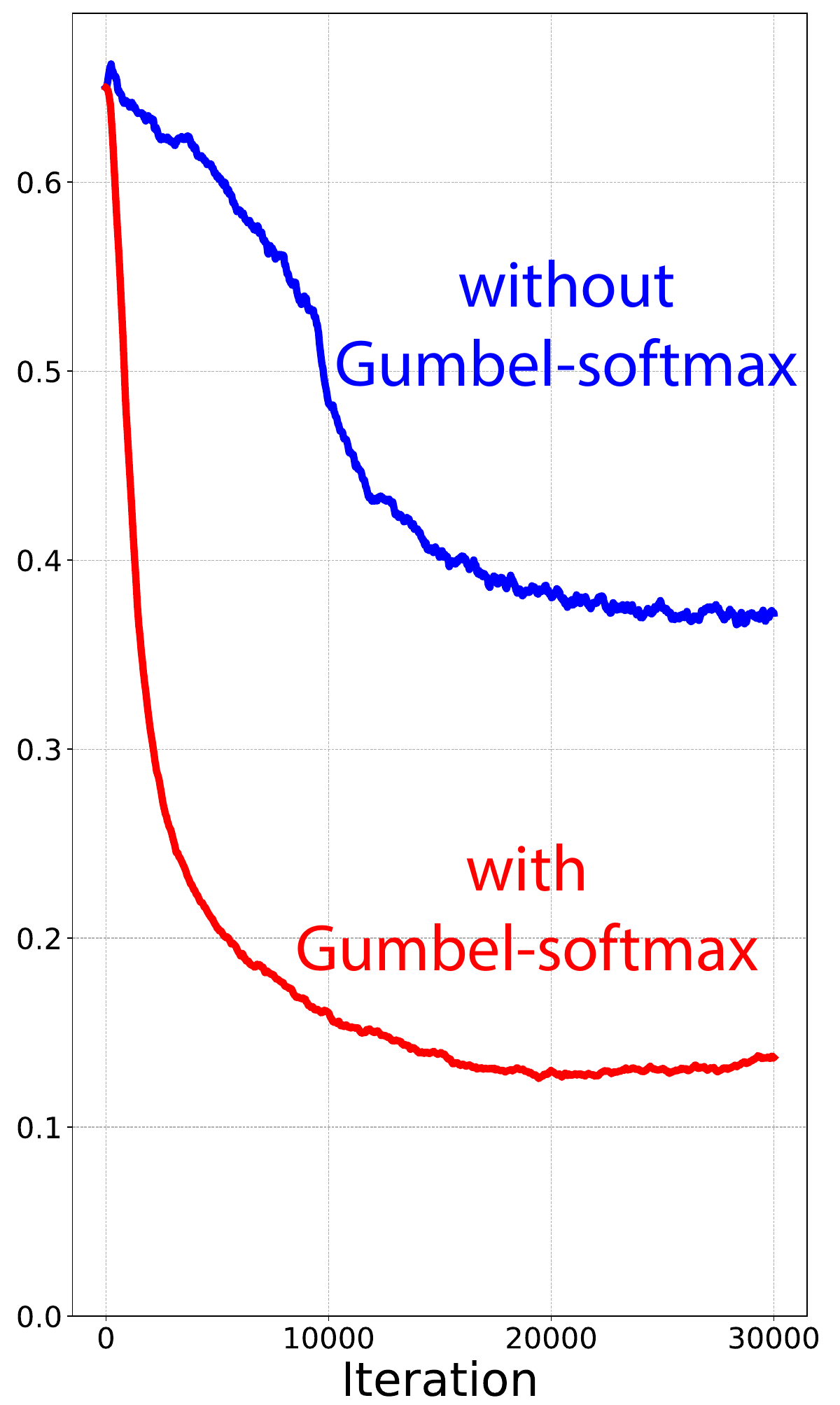}}\\
    & & & \\
\addlinespace[8pt]
    {\rotatebox{90}{\scriptsize \phantom{m}optimization with}} & {\rotatebox{90}{\scriptsize \phantom{m..}Gumbel-softmax}} &\includegraphics[trim= 0 0 0 0, clip, width=.215\linewidth]{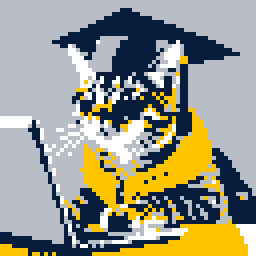} & 
    \includegraphics[trim= 0 0 0 0, clip, width=.215\linewidth]{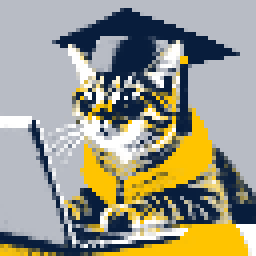} & 
    \includegraphics[trim= 0 0 0 0, clip, width=.215\linewidth]{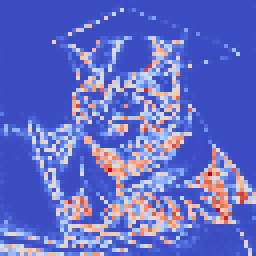} 
    & \\
\end{tabular}
    \caption{We show \ourmethod{}'s results {with} the Gumbel-softmax reparameterization (first row) and {without} (second row) during the optimization. The argmax-generation, the softmax-generation, the entropy per pixel and the average normalized entropy over time are displayed. Images are $64 \times 64$ pixels. The average normalized entropy is shown for 30,000 steps to ensure that the obtained results are not due to an early stop.}
    \label{fig:differentTrainingsGumbel}
    \Description{Influence of Gumbel reparameterization on the optmization.}
\end{figure}

\begin{figure*}[t!]
	\centering
	\scriptsize
	\setlength{\tabcolsep}{2pt}
	\begin{tabular}{cccccccccc}
 input & nearest-neighbor & PIA & DUP & MYOS & VectorFusion & \textbf{\ourmethod{}} & initialization & \textbf{\ourmethod{}} \\
  & interpolation & \cite{Gerstner:2012:PIA} & \cite{2018ChuDeepUnsupervizedPixelization} & \cite{Wu2022MakeYourOwnSprite} & \cite{jain2022vectorfusion} & \emph{K-means} & \multicolumn{2}{c}{\emph{palette}}\\
  
  \includegraphics[width=0.095\linewidth]{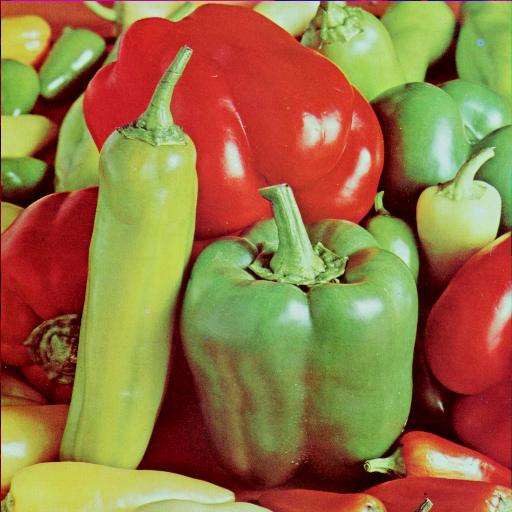}&
  \includegraphics[width=0.095\linewidth]{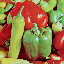} & 
  \includegraphics[width=0.095\linewidth]{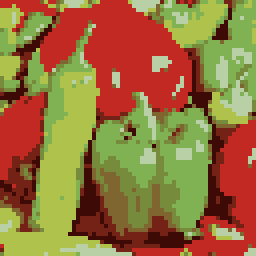} &   
  \includegraphics[width=0.095\linewidth]{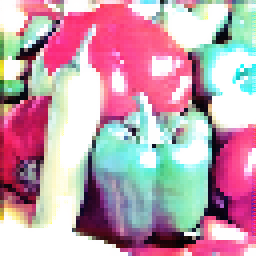} &
  \includegraphics[width=0.095\linewidth]{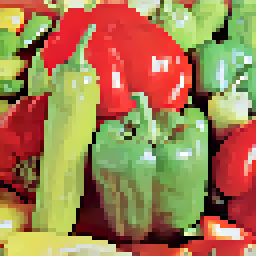} & 
  \includegraphics[width=0.095\linewidth]{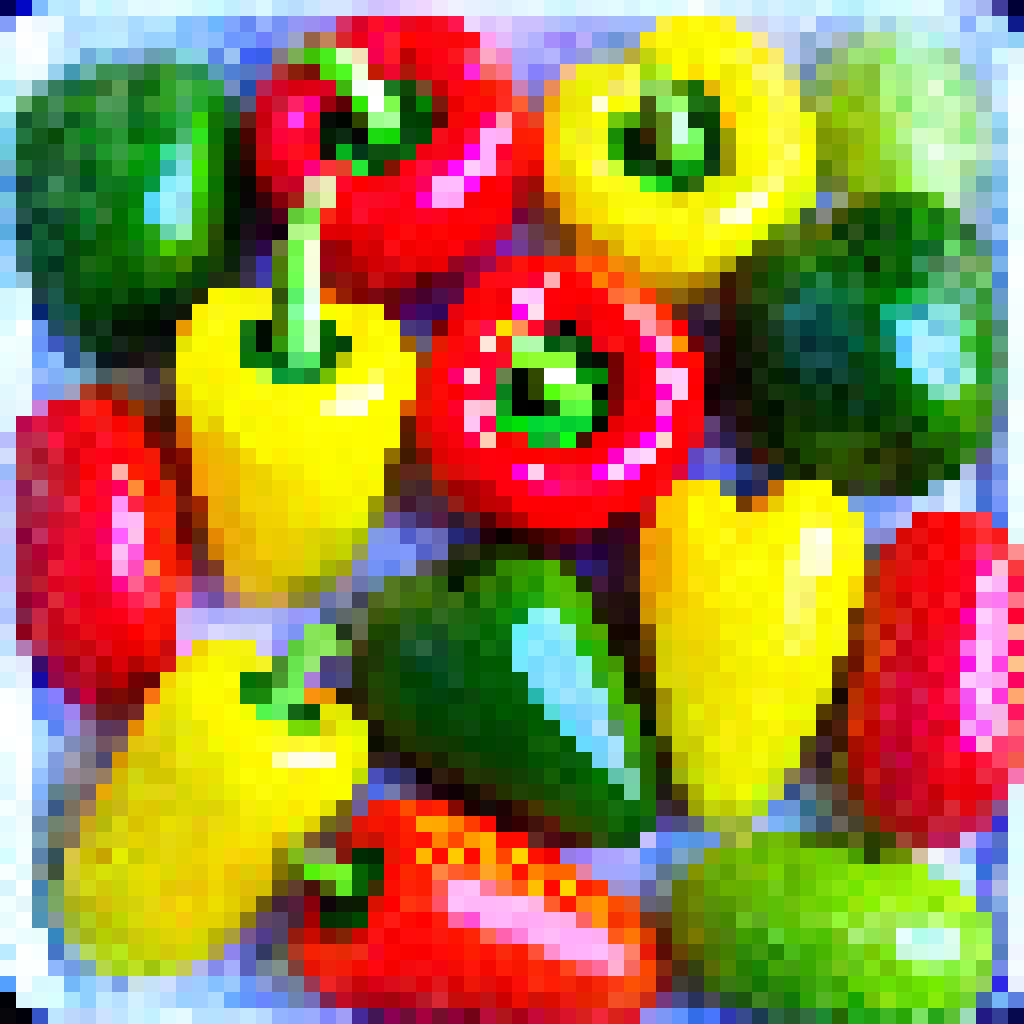}
  &
  \includegraphics[width=0.095\linewidth]{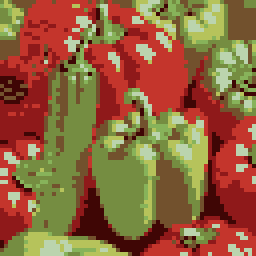} 
  &
  \includegraphics[width=0.095\linewidth]{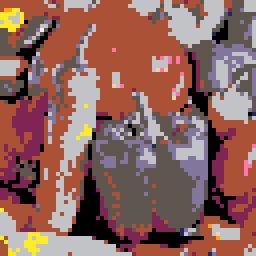} & 
  \includegraphics[width=0.095\linewidth]{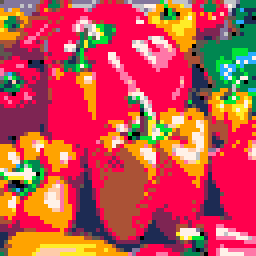}
  \\
  
  $512 \times 512$ & & & & & \multicolumn{4}{c}{An assortment of fresh bell peppers} \\
   \end{tabular}
 \caption{Visual comparison of pixelization methods with a downscale factor of 8. The input image is displayed with its size indicated below. Both VectorFusion and \ourmethod{} are initialized with the input image as their initial state, and conditioned on the prompt indicated below their results. 
We show the initialization with the palette to demonstrate how our method differs from classic palette matching. While PIA and the K-means variant of \ourmethod{} operate within a 8-color limit, nearest-neighbor interpolation, DUP, MYOS, and VectorFusion have no such constraints and are not quantized.}
 \label{fig:comparison}
 \Description{Visual comparison between several pixelization methods.}
\end{figure*}

\begin{figure}[t!]
\centering
\small
\setlength{\tabcolsep}{1pt}
\begin{tabular}{cccccc}
& & \multicolumn{4}{c}{Input conditioning} \\
& & \multicolumn{2}{c}{Depth Map} & \multicolumn{2}{c}{Canny Edge} \\
& & \multicolumn{2}{c}{\includegraphics[width=0.4\linewidth]{figures/supplementary/input/depth.png}}  & \multicolumn{2}{c}{\includegraphics[width=0.4\linewidth]{figures/supplementary/input/canny.png}} \\
\multicolumn{2}{r}{\multirow{2}{*}{\footnotesize Depth}} & \multirow{2}{*}{0.0} & \multirow{2}{*}{0.1} & \multirow{2}{*}{0.2} & \multirow{2}{*}{0.5}\\
\addlinespace[-0.1cm]
\raisebox{-7pt}[0pt][0pt]{{\footnotesize Canny}} & & & & &\\
\raisebox{19pt}[0pt][0pt]{0.0} & & \includegraphics[width=0.20\linewidth]{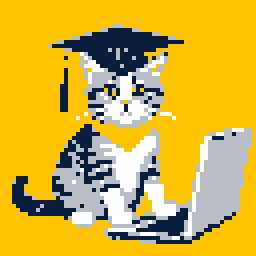} & \includegraphics[width=0.20\linewidth]{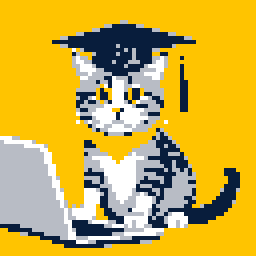} & \includegraphics[width=0.20\linewidth]{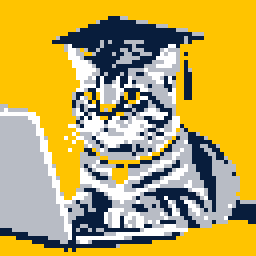} & \includegraphics[width=0.20\linewidth]{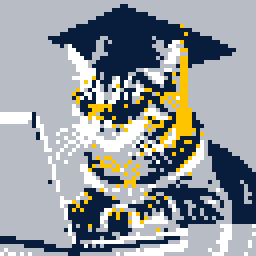} 
\\
\raisebox{19pt}[0pt][0pt]{0.1} & & \includegraphics[width=0.20\linewidth]{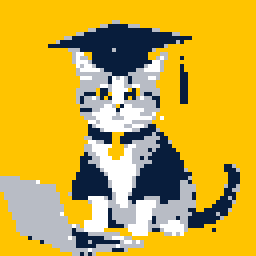} & \includegraphics[width=0.20\linewidth]{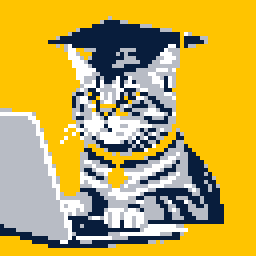} & \includegraphics[width=0.20\linewidth]{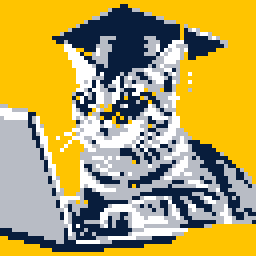} & \includegraphics[width=0.20\linewidth]{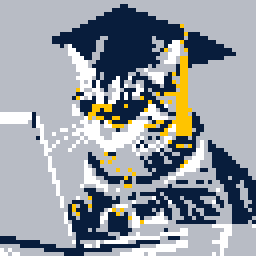} 
\\
\raisebox{19pt}[0pt][0pt]{0.2} & & \includegraphics[width=0.20\linewidth]{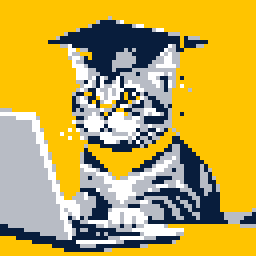} & \includegraphics[width=0.20\linewidth]{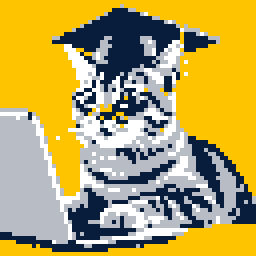} & \includegraphics[width=0.20\linewidth]{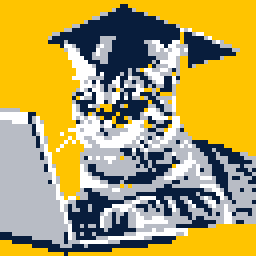} & \includegraphics[width=0.20\linewidth]{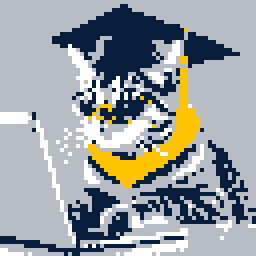} 
\\
\raisebox{19pt}[0pt][0pt]{0.5} & & \includegraphics[width=0.20\linewidth]{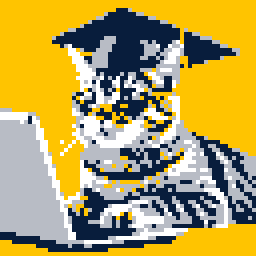} & \includegraphics[width=0.20\linewidth]{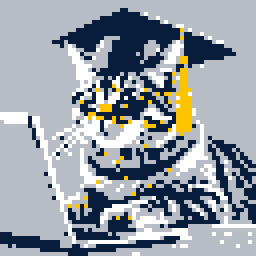} & \includegraphics[width=0.20\linewidth]{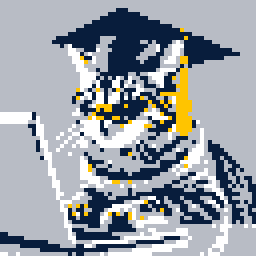} & \includegraphics[width=0.20\linewidth]{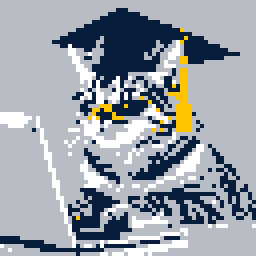} 
\\

\end{tabular}
\caption{We present the combined effects of ControlNet \cite{zhang2023addingControlNet} weights on both Canny edge and depth-conditioning networks \cite{vonPlaten2022diffusers}, examined concurrently. The image is randomly initialized to disambiguate the contribution of ControlNet from the influence of the initialization.}
\label{fig:ControlNetWeightsSimple}
\Description{Ablation on ControlNet to understand its influence on the generation.}
\end{figure}

\subsection{Stochastic vs.\ deterministic optimization}
\label{sec:gumbelanalysis}
We explain the rationale behind including Gumbel reparameterization during optimization for argmax-generated images.
In the stochastic optimization process, with Gumbel reparameterization, $\pi_{i, j, k}$ is interpreted as the likelihood of the element in position $(i, j)$ being $c_k$. Conversely, deterministic optimization (without Gumbel reparameterization) alters this perception, treating $\pi_{i, j, k}$ as coefficients in a convex combination of palette elements. This approach enables the generation of stylized low-resolution images through softmax-generation, as exemplified by the cat's fur texture in \figref{fig:differentTrainingsGumbel}, but it adversely impacts the readability of argmax-generated images.

To explain this phenomenon, we analyze the entropy per pixel. The entropy of a probability distribution quantifies its uncertainty \cite{Shannon48Entropy}, and is defined as $H(\pi_{i, j}) \coloneq -\sum_{k=0}^{n-1}{\pi_{i, j, k}\log(\pi_{i, j, k})}$. Given that a uniform distribution represents the peak of categorical distribution entropy, the maximum entropy is $\log n$. Thus, we use normalized entropy $\bar{H}(\pi_{i, j}) \coloneq \frac{1}{\log n}H(\pi_{i, j})$ to gauge pixel uncertainty independently of the palette size. Our findings reveal that the Gumbel reparameterization significantly reduces entropy, as displayed on \figref{fig:differentTrainingsGumbel}. Due to the pixel-wise independence in samples, employing the Gumbel-softmax reparameterization introduces noise in the results, which serves a beneficial purpose during the optimization phase, as the loss function is designed to counteract this noise. A probability distribution nearing uniformity, indicated by high entropy, leads to noisier images. Therefore, our optimization achieves denoising by encouraging the logits $\lambda_{i, j, k}$ to diverge significantly, effectively pushing the softmax towards a distinct class representation for each pixel. As a result, the optimized logits lead to a clearer, less noisy output by strongly favoring one class over the others in the softmax distribution, leading to crisper, pixelized visuals and lower entropy, demonstrated in Figs.\ \ref{fig:differentTrainingsGumbel} and \ref{fig:bigfigureResultsRandomInit}.

\begin{table}[b!]
\caption{Evaluation through a perceptual study, highlighting the performance of \ourmethod{} (Ours) in comparison to PIA and VectorFusion through semantic, fidelity and aesthetics questions. Each column aggregates the rankings across all questions in a specific category, representing the percentage of participants who placed each method at the respective rank (1, 2, 3, or 4) for that category.
}
\footnotesize
\centering
\setlength{\tabcolsep}{2pt}
\begin{tabular}{lcccccccccccccc}
\toprule
Method & \multicolumn{4}{c}{Semantic} & & \multicolumn{4}{c}{Fidelity} & & \multicolumn{4}{c}{Aesthetics} \\
Rank & \cellcolor{green!25}$1$ & $2$ & $3$ &$4$ & & \cellcolor{green!25}$1$ & $2$ & $3$ &$4$& & \cellcolor{green!25}$1$ & $2$ & $3$ &$4$ \\ \midrule
PIA  &  24.6  &  23.0  &  31.2  &  21.2  & &  \cellcolor{green!25}\textbf{49.3} &  37.9  &  11.1  &  1.8  & &  25.5  &  26.0  &  33.1  &  15.5  \\
VectorFusion  &  22.0  &  15.0  &  16.1  &  46.9  & &  0.8  &  0.8  &  8.0  &  90.4  & &  17.5  &  12.1  &  19.5  &  50.8  \\
Ours-\emph{K-means}  &  \cellcolor{green!25}\textbf{36.2}  &  37.0  &  22.6  &  4.2  & &  47.1  &  48.0  &  4.6  &  0.2  & &  26.5  &  38.8  &  22.3  &  12.4  \\
Ours-\emph{palette}  &  17.1  &  25.0  &  30.1  &  27.7  & &  2.7  &  13.3  &  76.3  &  7.6  & &  \cellcolor{green!25}\textbf{30.5}  &  23.1  &  25.1  &  21.3  \\
\bottomrule
\end{tabular}
\label{tab:userstudyResults}
\end{table}

\subsection{ControlNet influence}
\label{sec:controlnet}
As explained in \secref{sec:inputImage}, the Canny edge and depth maps of the input image can spatially condition the generation via ControlNet \cite{zhang2023addingControlNet}. The user can modulate the weights used for controlling the generation, and \figref{fig:ControlNetWeightsSimple} shows that incrementing ControlNet's weights increases the fidelity of the result to the input image layout. Additional comparisons are available in the supplementary material.

\subsection{Pixelization evaluation}
We extensively evaluate the use of our method for pixelization through a quantitative comparison and a perceptual study. We compare with Pixelated Image Abstraction (PIA) \cite{Gerstner:2012:PIA}, quantized Make Your Own Sprite (MYOS) \cite{Wu2022MakeYourOwnSprite} and VectorFusion \cite{jain2022vectorfusion}. Our method is presented in two forms: the ``palette'' variant utilizes a predefined palette, and the ``K-means'' variant computes a palette from the input image using K-means clustering. We provide one visual comparison in \figref{fig:comparison}, and several additional examples are provided in the supplementary material alongside additional details and result metrics of our quantitative evaluation.
For our quantitative evaluation, we generate 150 images and pixelize them. We analyze pixelization methods across three metrics: semantic similarity, fidelity and aesthetics. The metrics show distinct strengths: VectorFusion achieves the best semantic accuracy, while MYOS and PIA lead in fidelity. Our method excels in aesthetics due to its superior color harmony. Despite the limitations imposed by color quantization, our \ourmethod{} variants also deliver competitive results for both semantic accuracy and fidelity, effectively balancing these objectives and providing the most aesthetically pleasing results overall. These results are corroborated in \figref{fig:comparison}: the results from nearest-neighbor, PIA and MYOS are very close to the input, but at the expense of aesthetics or clarity. DUP tends to show saturated colors, and VF diverges significantly from the input image due to lack of spatial conditioning. Our method strikes a balance between fidelity and aesthetics, even on a color palette very different from the input's colors.
We also conducted a perceptual study, where 56 participants evaluated 45 images sampled randomly and rated each based on the given criteria. Results, displayed in \tableref{tab:userstudyResults} and \tableref{tab:userstudyIQR}, showed our \emph{K-means} variant excelling in semantic accuracy, while PIA led in fidelity. The \emph{palette} variant was favored for its aesthetic appeal, and VectorFusion generally received lower rankings across all categories, indicating some limitations in these aspects compared to other methods.
\subsection{Limitations and future work}

We acknowledge several limitations and future research areas for \ourmethod{}.
While our method does not require training a network from scratch, the overall optimization process can be quite slow, requiring 1.5 hours on an Nvidia RTX4090 for 6000 steps. Additionally, the model's reliance on prompts is a limitation. Further exploration into image-only semantic conditioning \cite{ye2023ip-adapter} could potentially eliminate the need for prompts and increase fidelity.
Another limitation of our method is the independent sampling for each pixel. Stochastic sampling conditioned on multiple pixels or joint probability distribution between neighboring pixels could improve the awareness of the method at a more global level, which could improve its overall quality and convergence speed.
Moreover, the prospect of achieving frame-to-frame consistency in pixelized animations offers a promising direction for future extensions of this work, especially as text-to-video diffusion models continue to advance \cite{vdmsurvey}.
On a more general level, \ourmethod{} is inherently constrained by the limitations of the underlying diffusion models, including ethical concerns \cite{birhane2021multimodal}. With further advancements in text-to-image models and diffusion techniques, we anticipate corresponding improvements in the capabilities of \ourmethod{}.

%% file: sections/6_conclusion.tex
\begin{table}[b!]
\vspace{-0.0cm}
\small
\captionof{table}{First quartile (Q1), median (Med.) and interquartile range (IQR) of the results of our perceptual study, according to semantic similarity, fidelity to input image and aesthetic appeal.}
\centering
\setlength{\tabcolsep}{3pt}
\begin{tabular}{l ccc ccc ccc}
\toprule
Method & \multicolumn{3}{c}{Semantics} & \multicolumn{3}{c}{Fidelity} & \multicolumn{3}{c}{Aesthetics} \\
 & Q1 & Med.& IQR &  Q1 &Med.& IQR& Q1 &Med.& IQR \\ \midrule 
PIA & 2.0 & 3.0 & 1.0& 1.0 &2.0 & 1.0 & 1.0 & 2.0 & 2.0\\
VF & 2.0 & 3.0 & 2.0& 4.0 & 4.0 & 0.0 & 2.0 & 4.0 & 2.0\\
Ours-\emph{K-means}& 1.0 &  2.0 & 2.0& 1.0 & 2.0 & 1.0 & 1.0 & 2.0 & 2.0\\
Ours-\emph{palette}& 2.0 & 3.0 & 2.0& 3.0& 3.0 & 0.0 & 1.0 &2.0 & 2.0\\
\bottomrule
\end{tabular}
\label{tab:userstudyIQR}
\end{table}
\section{Conclusion}

This paper introduced \ourmethod{}, a method for generating low resolution, color-quantized images via semantic conditioning through diffusion-based networks. Central to our approach is the ability to strictly adhere to predefined constraints, such as input color palettes, which ensures the generation of crisp pixel art. \figref{fig:bigfigure1} shows that our method has flexible generation capabilities, working for any desired input resolution or color palette, incorporating both semantic and image-based conditioning, and is amenable to stylization via LoRA finetuning.
We demonstrate through comprehensive experiments and comparative studies the performance of \ourmethod{} in generating quantized images that are not only visually appealing but also accurate to the specified constraints. Our technical contribution consists in the use of the Gumbel-softmax reparameterization, justified both on the theoretical and empirical front for pixel art generation.
Moreover, \ourmethod{}'s state-of-the-art results in quantized image generation are evident in its ability to produce pixel art that meets modern-day fabrication and design requirements. Thanks to its strict adherence to a given palette, it can be directly utilized to create instructions for crafting with beads, interlocking bricks, or to embroider images using discrete styles such as cross-stitch.
We produced several such physical creations, shown in Figs.\ \ref{fig:teaser}, \ref{fig:fabrication}, \ref{fig:introdiffusionComparison}. We believe that \ourmethod{} offers a powerful tool for artists, game developers and designers, helping make pixel art creation more accessible and versatile.

%% file: sections/7_acknowledgments.tex
\begin{acks}
We thank the anonymous reviewers for their constructive feedback and Danielle Luterbacher for her help with setting up the embroidery machine. Ximing Xing's open-source version of VectorFusion was instrumental in the development and design of our source code. This work was supported in part by the European Research Council (ERC) under the European Union’s Horizon 2020 research and innovation program (grant agreement No. 101003104, ERC CoG MYCLOTH).
\end{acks}

%% file: sections/big_figure.tex
\begin{figure*}[t]
	\centering
	\small
	\setlength{\tabcolsep}{1pt}
	\begin{tabular}{cccccccc}
 
    Kmeans
    & \includegraphics[width=0.123\linewidth]{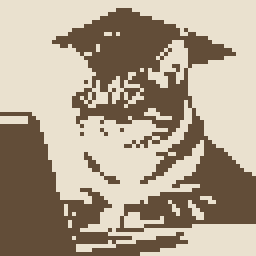}\
    & \includegraphics[width=0.123\linewidth]{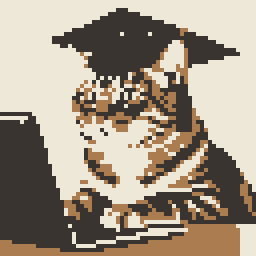}\
    & \includegraphics[width=0.123\linewidth]{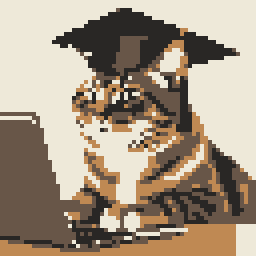}\
    & \includegraphics[width=0.123\linewidth]{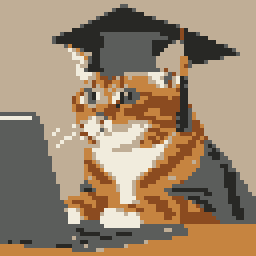}\
    & \includegraphics[width=0.123\linewidth]{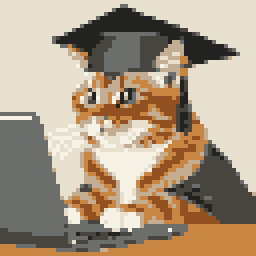}\
    & \includegraphics[width=0.123\linewidth]{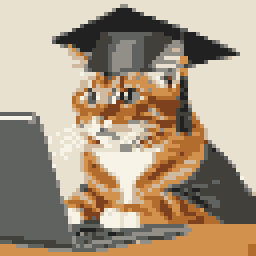}\
    & \includegraphics[width=0.123\linewidth]{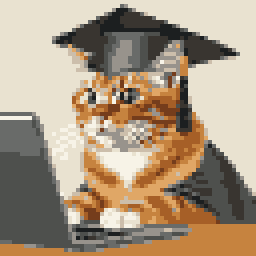} \\
    colors & 2 & 3 & 4 & 6 & 8 & 10 & 12\\
    
 
    & \includegraphics[width=0.123\linewidth]{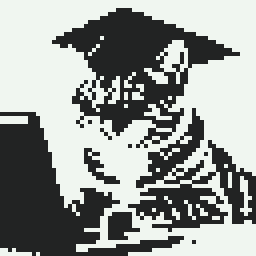}\
    & \includegraphics[width=0.123\linewidth]{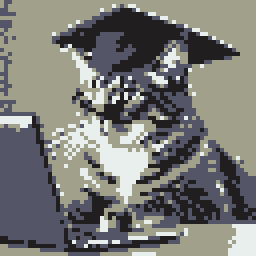}
    & \includegraphics[width=0.123\linewidth]{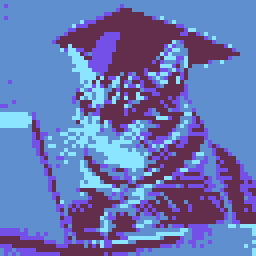}
    & \includegraphics[width=0.123\linewidth]{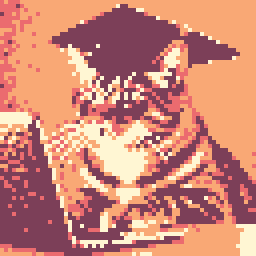}
    & \includegraphics[width=0.123\linewidth]{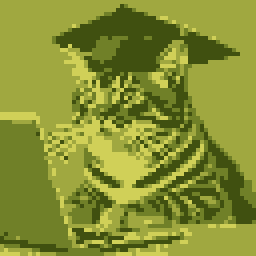}
    &\includegraphics[width=0.123\linewidth]{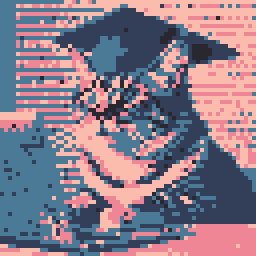} 
    & \includegraphics[width=0.123\linewidth]{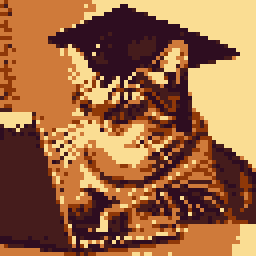} 
    \\
    palette & 
    \includegraphics[height=9pt]{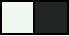}
    & 
    \includegraphics[height=9pt]{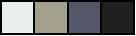}
    & 
    \includegraphics[height=9pt]{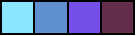}
    & 
    \includegraphics[height=9pt]{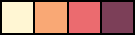}
    & 
    \includegraphics[height=9pt]{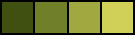}
    & 
    \includegraphics[height=9pt]{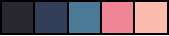}
    & 
    \includegraphics[height=9pt]{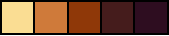}
     \\
    & \includegraphics[width=0.123\linewidth]{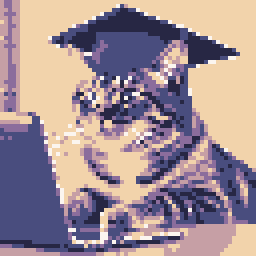}
    & \includegraphics[width=0.123\linewidth]{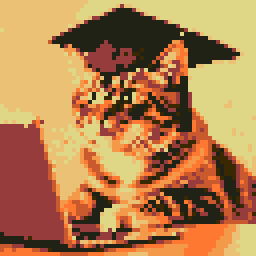}
    & \includegraphics[width=0.123\linewidth]{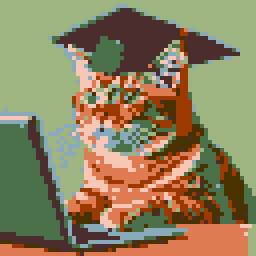}
    & \includegraphics[width=0.123\linewidth]{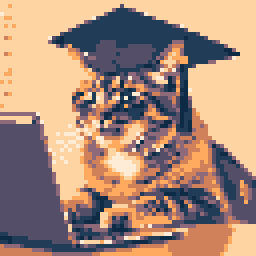}
    & \includegraphics[width=0.123\linewidth]{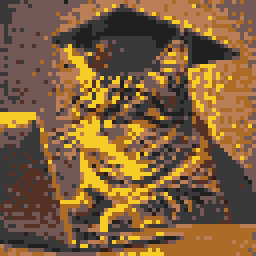}
    & \includegraphics[width=0.123\linewidth]{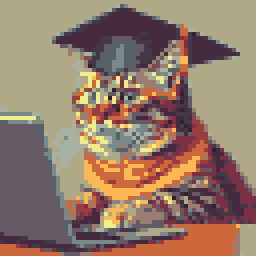} 
    & \includegraphics[width=0.123\linewidth]{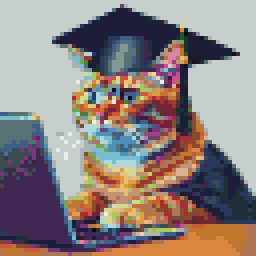} \\
    
    \multirow{2}{*}{palette} & 
    \multirow{2}{*}{\includegraphics[height=18pt]{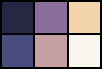}}
    & 
    \multirow{2}{*}{\includegraphics[height=18pt]{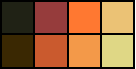}}
    & 
    \multirow{2}{*}{\includegraphics[height=18pt]{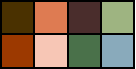}}
    & 
    \multirow{2}{*}{\includegraphics[height=18pt]{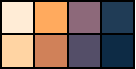}}
    & 
    \multirow{2}{*}{\includegraphics[height=18pt]{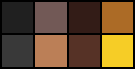}}
    & 
    \multirow{2}{*}{\includegraphics[height=18pt]{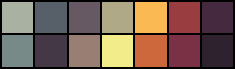}}
    & 
    \multirow{2}{*}{\includegraphics[height=18pt]{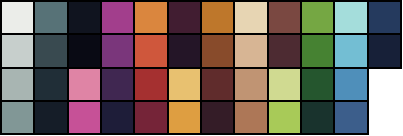}}
     \\
 
 & & & & & & & \\
    
  & \includegraphics[width=0.123\linewidth]{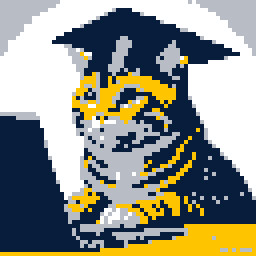}
    & \includegraphics[width=0.123\linewidth]{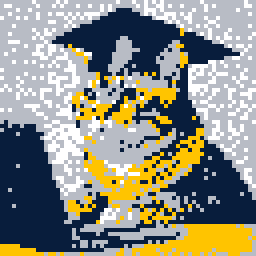}\
    & \includegraphics[width=0.123\linewidth]{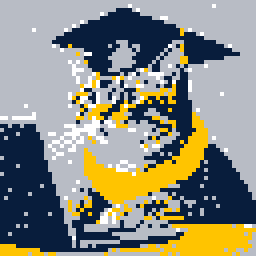} \
    & \includegraphics[width=0.123\linewidth]{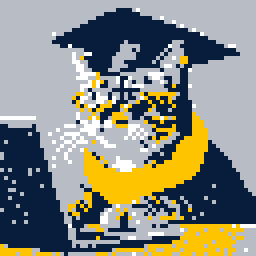} \
    & \includegraphics[width=0.123\linewidth]{figures/bigfigure/bigfig1/steps/6000.png} \
    & \includegraphics[width=0.123\linewidth]{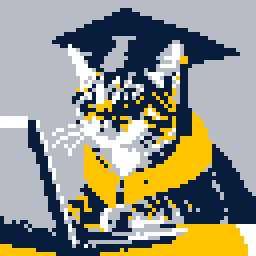} \
    & \includegraphics[width=0.123\linewidth]{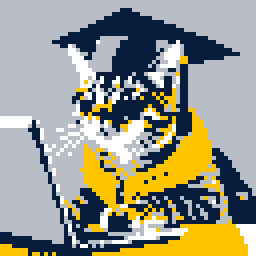} \\
    
    steps & initialization & $600$ & $1500$ & $3000$ & $6000$ & $15000$ & $30000$ \\
    
    & \includegraphics[width=0.123\linewidth]{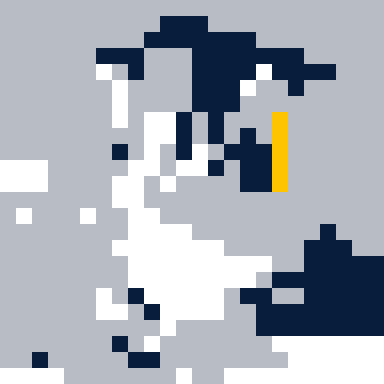} \
    & \includegraphics[width=0.123\linewidth]{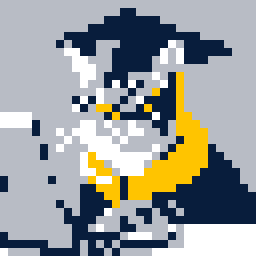} \
    & \includegraphics[width=0.123\linewidth]{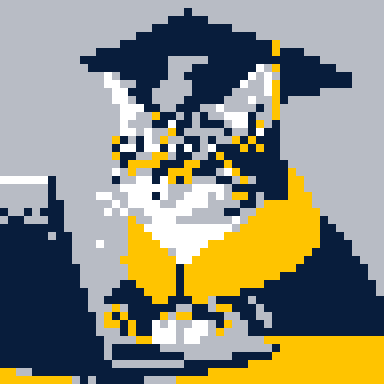} \
    & \includegraphics[width=0.123\linewidth]{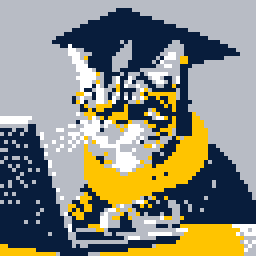} \
    & \includegraphics[width=0.123\linewidth]{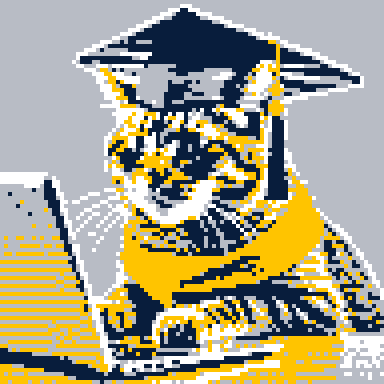} \
    & \includegraphics[width=0.123\linewidth]{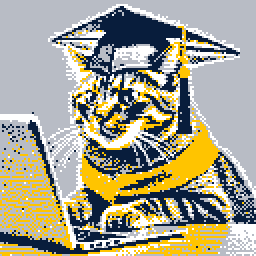} \
    & \includegraphics[width=0.123\linewidth]{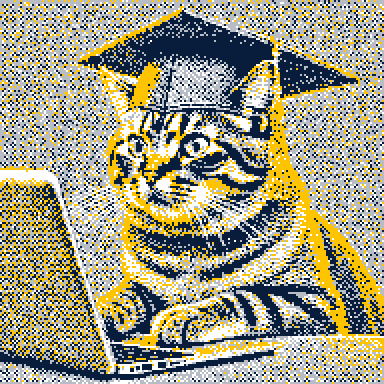} \\
    resolution & $24\times 24$ & $32\times 32$ & $48\times 48$ & $64\times 64$ & $96\times 96$ & $128\times 128$ & $192\times 192$ \\

    & \includegraphics[width=0.123\linewidth]{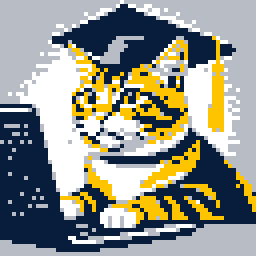} \
    & \includegraphics[width=0.123\linewidth]{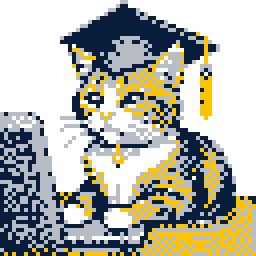}
    & \includegraphics[width=0.123\linewidth]{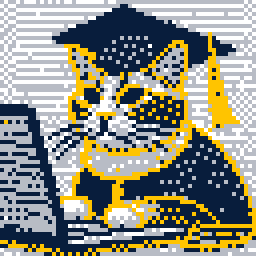}
    & \includegraphics[width=0.123\linewidth]{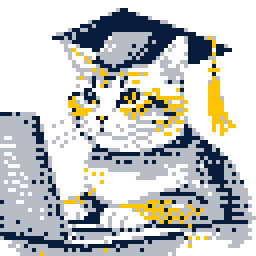}
    & \includegraphics[width=0.123\linewidth]{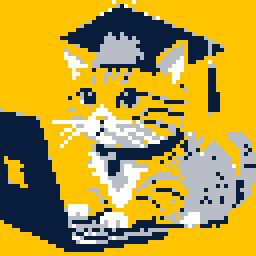}
    & \includegraphics[width=0.123\linewidth]{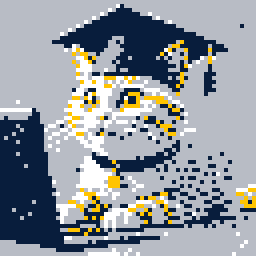}
    & \includegraphics[width=0.123\linewidth]{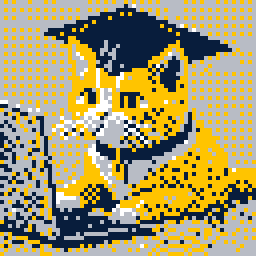} \\
    LoRA & none &  \href{https://huggingface.co/nerijs/pixel-art-xl}{Pixel art style} &
    \href{https://huggingface.co/ostris/embroidery_style_lora_sdxl}{Embroidery style} &  
    \href{https://huggingface.co/ostris/watercolor_style_lora_sdxl}{Water color style} & 
    \href{https://huggingface.co/TheLastBen/Papercut_SDXL}{Papercut style} &  
    \href{https://huggingface.co/goofyai/3d_render_style_xl}{3D render style} &  
    \href{https://huggingface.co/Fictiverse/Voxel_XL_Lora}{Voxel style} \\
	\end{tabular}
 \vspace{-0.3cm}
	\caption{The first row illustrates \ourmethod{} using K-means color clustering with varying numbers of colors. Rows 2 and 3 display the application of our method with different color palettes, and shows that our method works with any number of colors. The progression of \ourmethod{} through various time steps is depicted in row 4. In row 5, we showcase outputs at different resolutions. The final row showcases \ourmethod{} with diffusion models fine-tuned to distinct styles via low-rank adaption (LoRA) \cite{hu2021lora}, to demonstrate the generalizability of our approach. Each name is a clickable link that directs to the corresponding LoRA. For a clearer distinction in style variations, we opt not to use ControlNet for the images in the last row. The chosen prompt for this demonstration is ``A cat wearing a graduation hat using a computer'', with the input image and further conditioning details provided in supplementary material. 
 }
	\label{fig:bigfigure1}
 \Description{Our method with different palettes, resolution, LoRA, and at different steps during the optimization.}
\end{figure*}

\newpage




\begin{figure*}[t]
	\centering
	\small
	\setlength{\tabcolsep}{1pt}
 \begin{tabular}{*{10}c}
 \toprule
 \multicolumn{10}{c}{with Gumbel-softmax reparameterization - argmax generation}\\
 \raisebox{20pt}[0pt][0pt]{\rotatebox{0}{\scriptsize $32\times32$}} & \includegraphics[width=0.1\linewidth]{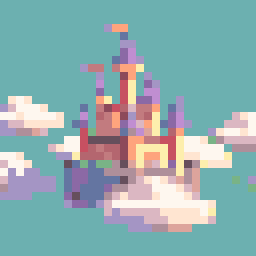} & \includegraphics[width=0.1\linewidth]{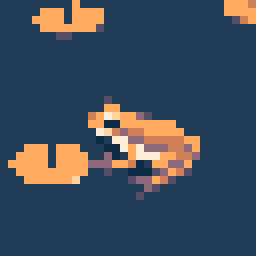} & \includegraphics[width=0.1\linewidth]{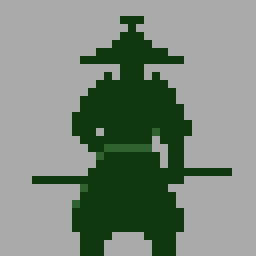} & \includegraphics[width=0.1\linewidth]{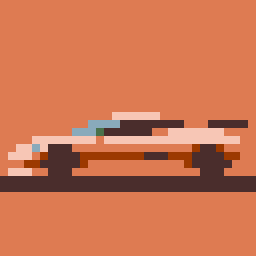} & \includegraphics[width=0.1\linewidth]{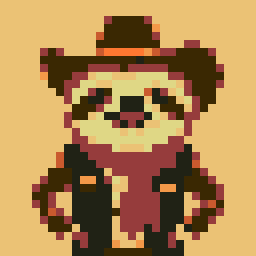} & \includegraphics[width=0.1\linewidth]{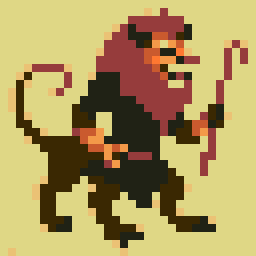} & \includegraphics[width=0.1\linewidth]{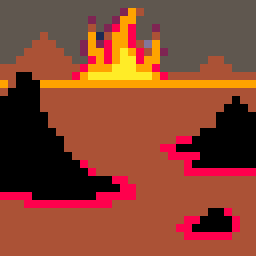} & \includegraphics[width=0.1\linewidth]{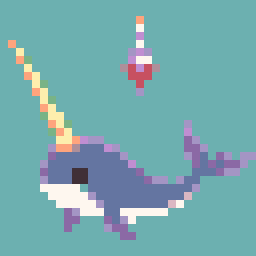} & \includegraphics[width=0.1\linewidth]{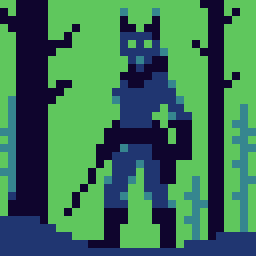} \\
 \raisebox{20pt}[0pt][0pt]{\rotatebox{0}{\scriptsize $48\times48$}} & \includegraphics[width=0.1\linewidth]{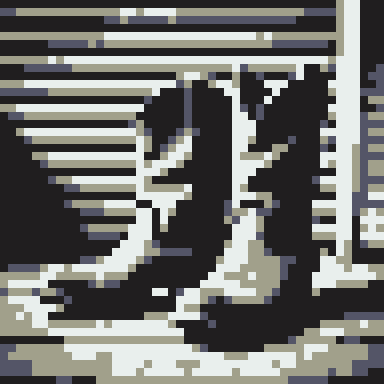} & \includegraphics[width=0.1\linewidth]{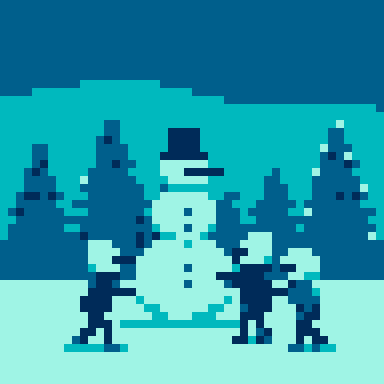} & \includegraphics[width=0.1\linewidth]{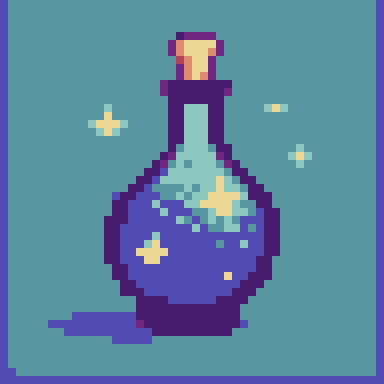} & \includegraphics[width=0.1\linewidth]{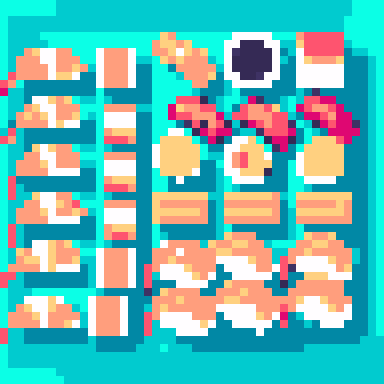} & \includegraphics[width=0.1\linewidth]{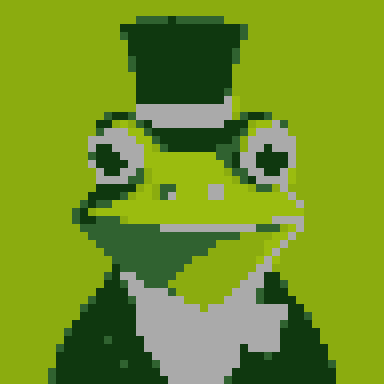} & \includegraphics[width=0.1\linewidth]{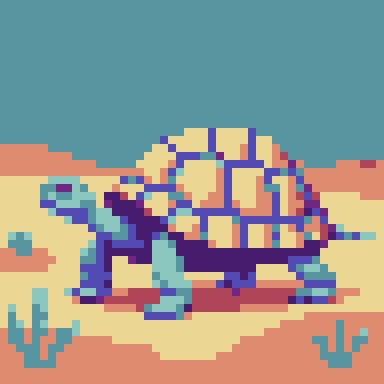} & \includegraphics[width=0.1\linewidth]{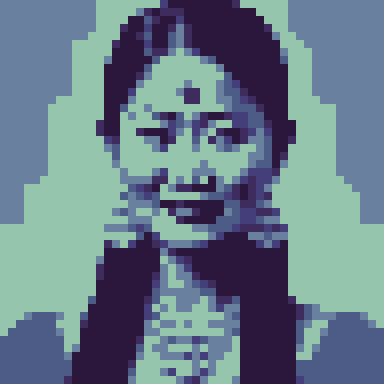} & \includegraphics[width=0.1\linewidth]{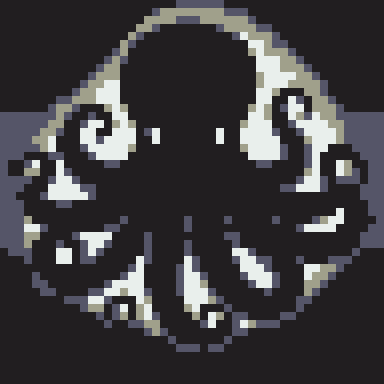} & \includegraphics[width=0.1\linewidth]{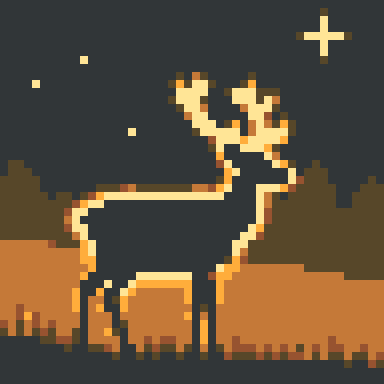} \\
 \raisebox{20pt}[0pt][0pt]{\rotatebox{0}{\scriptsize $64\times64$}} & \includegraphics[width=0.1\linewidth]{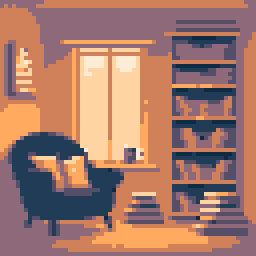} & \includegraphics[width=0.1\linewidth]{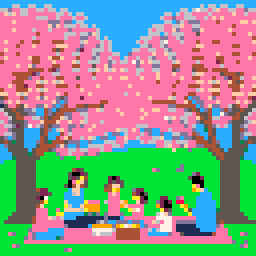} & \includegraphics[width=0.1\linewidth]{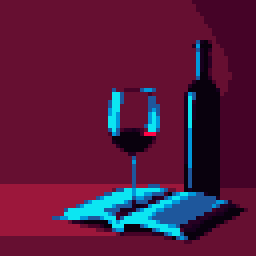} & \includegraphics[width=0.1\linewidth]{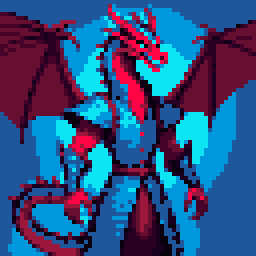} & \includegraphics[width=0.1\linewidth]{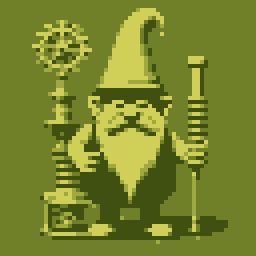} & \includegraphics[width=0.1\linewidth]{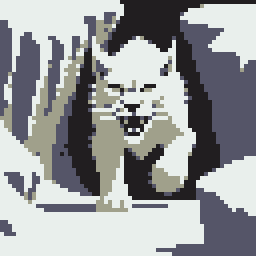} & \includegraphics[width=0.1\linewidth]{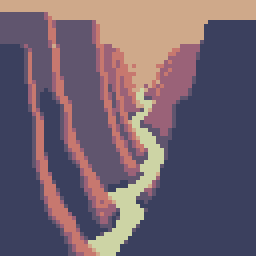} & \includegraphics[width=0.1\linewidth]{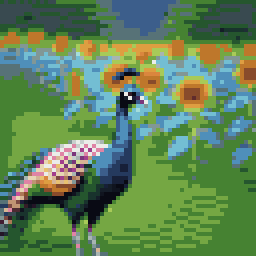} & \includegraphics[width=0.1\linewidth]{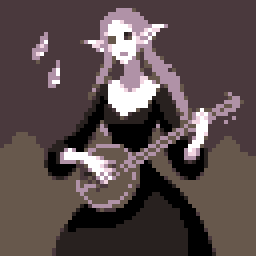} \\
 \raisebox{20pt}[0pt][0pt]{\rotatebox{0}{\scriptsize $96\times96$}} & \includegraphics[width=0.1\linewidth]{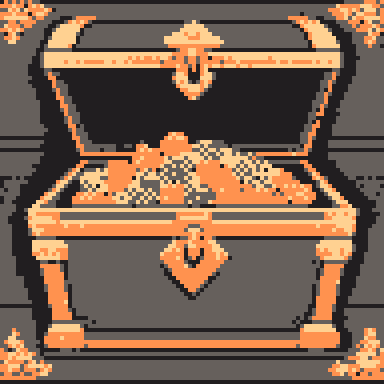} & \includegraphics[width=0.1\linewidth]{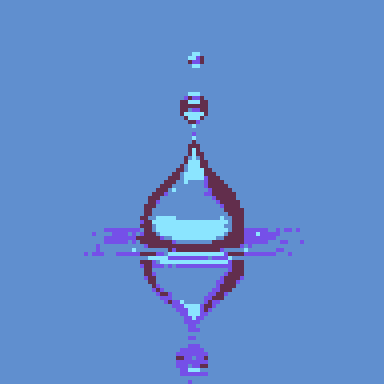} & \includegraphics[width=0.1\linewidth]{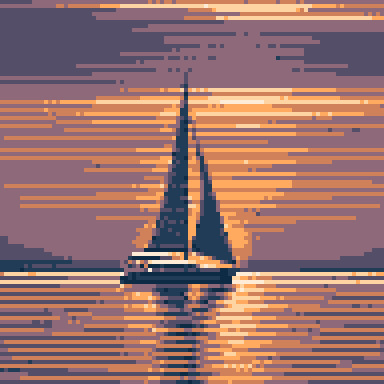} & \includegraphics[width=0.1\linewidth]{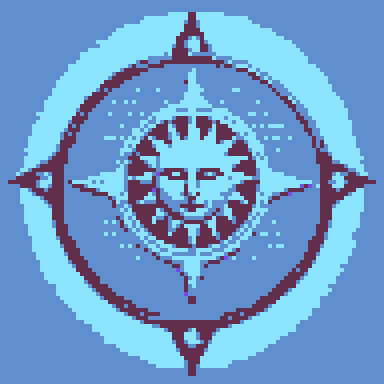} & \includegraphics[width=0.1\linewidth]{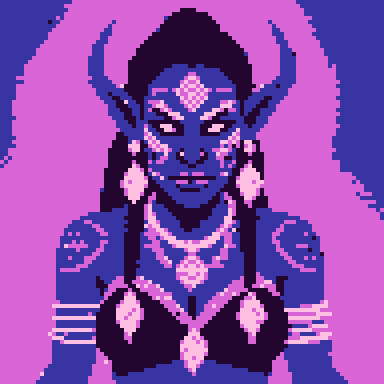} & \includegraphics[width=0.1\linewidth]{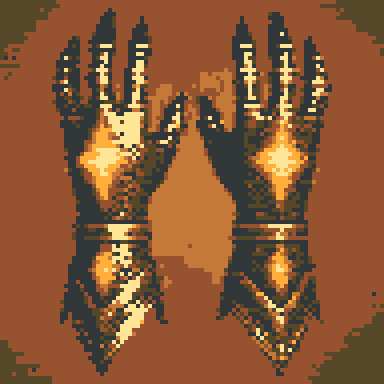} & \includegraphics[width=0.1\linewidth]{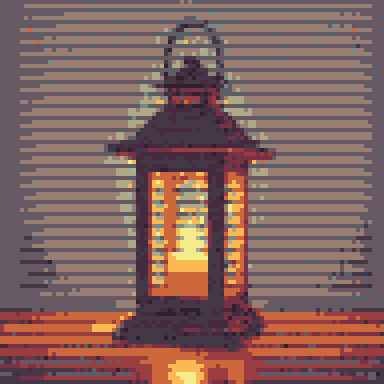} & \includegraphics[width=0.1\linewidth]{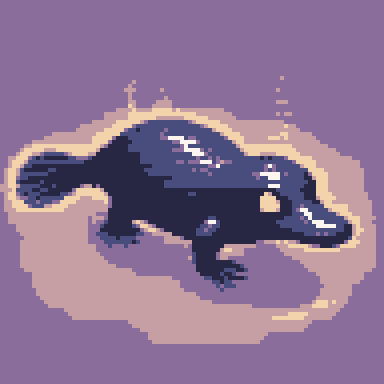} & \includegraphics[width=0.1\linewidth]{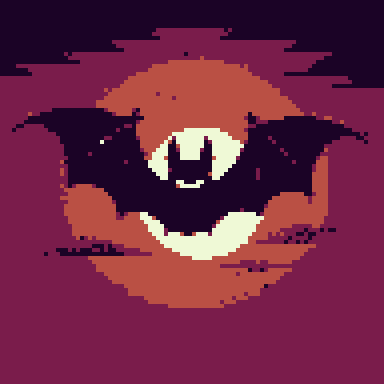} \\
 \midrule
  \multicolumn{10}{c}{without Gumbel-softmax reparameterization - softmax generation}\\
 \raisebox{20pt}[0pt][0pt]{\rotatebox{0}{\scriptsize $32\times32$}} & \includegraphics[width=0.1\linewidth]{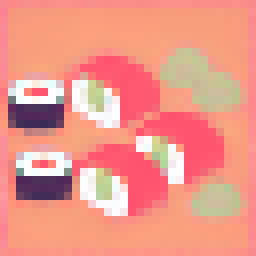} & \includegraphics[width=0.1\linewidth]{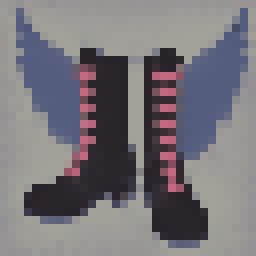} & \includegraphics[width=0.1\linewidth]{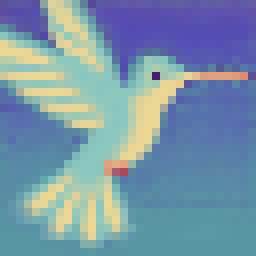} & \includegraphics[width=0.1\linewidth]{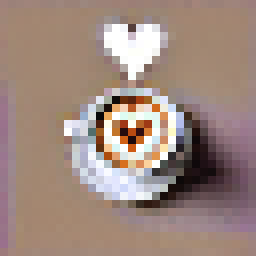} & \includegraphics[width=0.1\linewidth]{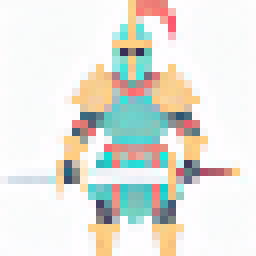} & \includegraphics[width=0.1\linewidth]{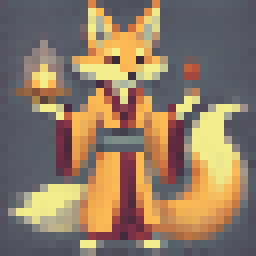} & \includegraphics[width=0.1\linewidth]{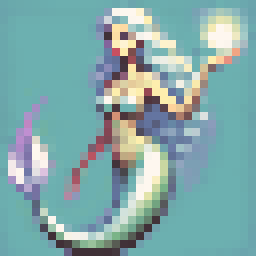} & \includegraphics[width=0.1\linewidth]{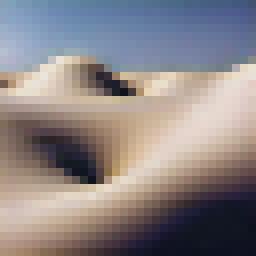} & \includegraphics[width=0.1\linewidth]{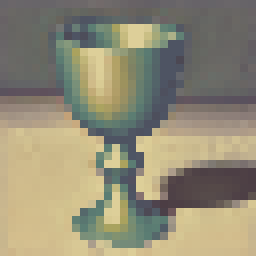} \\
 \raisebox{20pt}[0pt][0pt]{\rotatebox{0}{\scriptsize $48\times48$}} & \includegraphics[width=0.1\linewidth]{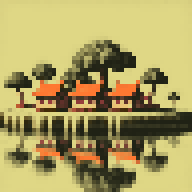} & \includegraphics[width=0.1\linewidth]{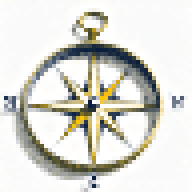} & \includegraphics[width=0.1\linewidth]{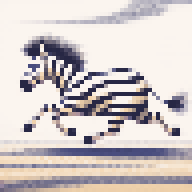} & \includegraphics[width=0.1\linewidth]{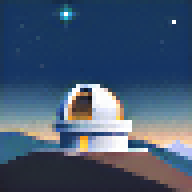} & \includegraphics[width=0.1\linewidth]{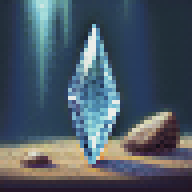} & \includegraphics[width=0.1\linewidth]{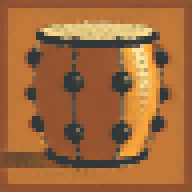} & \includegraphics[width=0.1\linewidth]{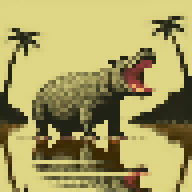} & \includegraphics[width=0.1\linewidth]{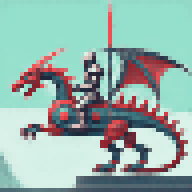} & \includegraphics[width=0.1\linewidth]{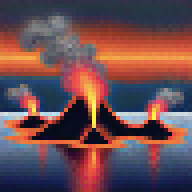} \\
 \raisebox{20pt}[0pt][0pt]{\rotatebox{0}{\scriptsize $64\times64$}} & \includegraphics[width=0.1\linewidth]{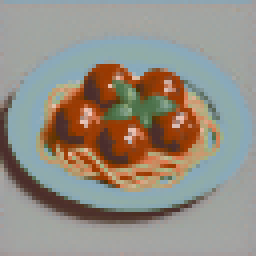} & \includegraphics[width=0.1\linewidth]{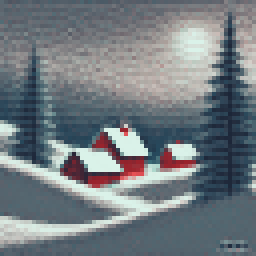} & \includegraphics[width=0.1\linewidth]{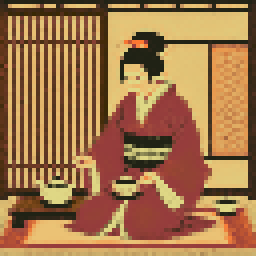} & \includegraphics[width=0.1\linewidth]{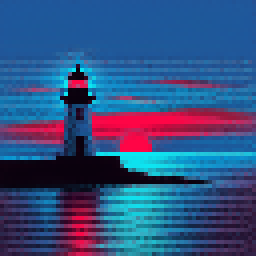} & \includegraphics[width=0.1\linewidth]{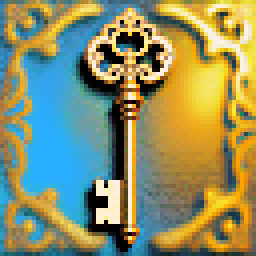} & \includegraphics[width=0.1\linewidth]{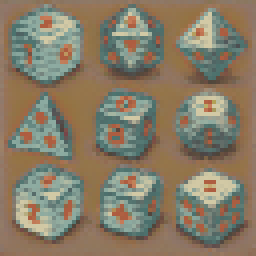} & \includegraphics[width=0.1\linewidth]{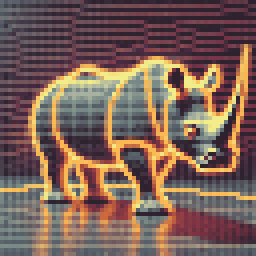} & \includegraphics[width=0.1\linewidth]{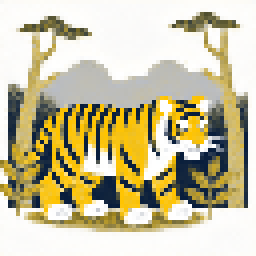} & \includegraphics[width=0.1\linewidth]{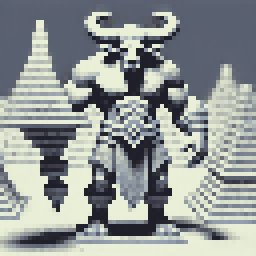} \\
 \raisebox{20pt}[0pt][0pt]{\rotatebox{0}{\scriptsize $96\times96$}} & \includegraphics[width=0.1\linewidth]{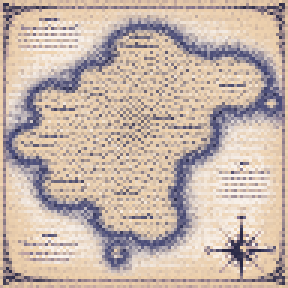} & \includegraphics[width=0.1\linewidth]{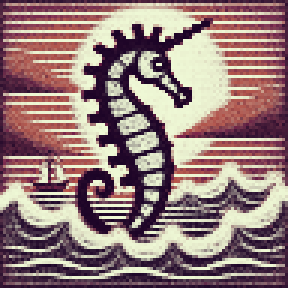} & \includegraphics[width=0.1\linewidth]{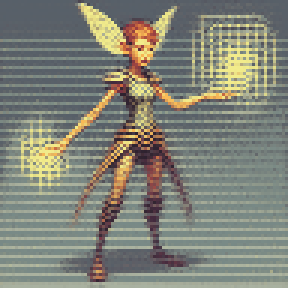} & \includegraphics[width=0.1\linewidth]{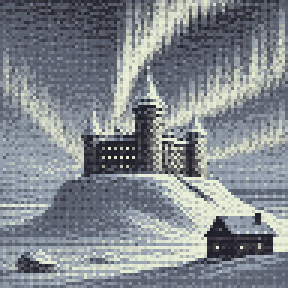} & \includegraphics[width=0.1\linewidth]{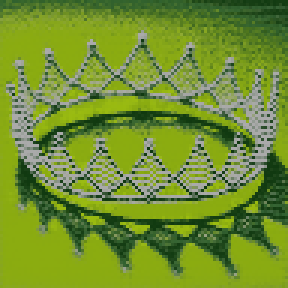} & \includegraphics[width=0.1\linewidth]{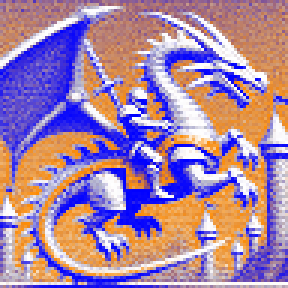} & \includegraphics[width=0.1\linewidth]{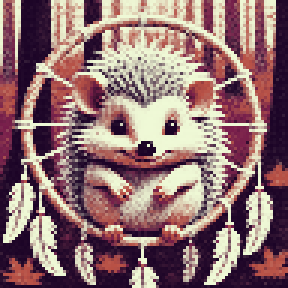} & \includegraphics[width=0.1\linewidth]{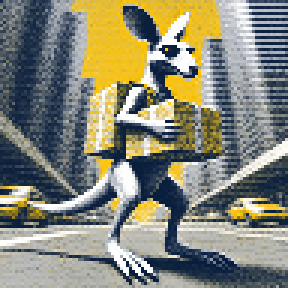} & \includegraphics[width=0.1\linewidth]{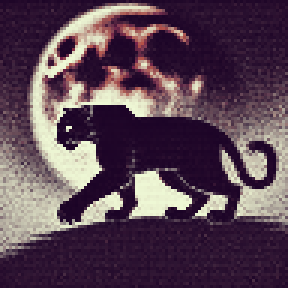} \\
 \bottomrule
 \end{tabular}
    \caption{Pixel art generation with \ourmethod{}, used without initial image or spatial conditioning. We present results on several resolutions, written on the leftmost column. The table is divided in two generation methods: the first part presents results with Gumbel-softmax reparameterization during optimization, generated with argmax. This generation method produces crisp pixel art that strictly adheres to the input palette. The second part does not use the Gumbel-softmax reparameterization, but uses a softmax generation to produce smooth, low-resolution images whose colors lie in the convex hull of the input palette.}	\label{fig:bigfigureResultsRandomInit}
    \Description{Examples of pixel art generation, both with softmax and argmax optimization.}
\end{figure*}

%% file: sections/supplementary/0_abstract_suppl.tex
Our supplementary material offers further insights into the implementation of our method, illustrates the impact of various parameters, and includes details about the evaluation of our approach.

%% file: sections/supplementary/1_implementation.tex
\section{Implementation Details}

This section outlines the specifics of our method's implementation, including the default parameters and technical configurations utilized. The source code is made available at \url{https://github.com/AlexandreBinninger/SD-piXL}.

\subsection{Technical details}

Our method employs backpropagation through the encoder of the latent diffusion model. To enhance efficiency and minimize memory usage, we utilize a distilled version of the stable diffusion VAE, namely taesdxl \cite{madebyollin2023taesdxl}.
We adopt mid versions of the Canny edge and depth ControlNets \cite{zhang2023addingControlNet}, specifically ``controlnet-canny-sdxl-1.0-mid'' and ``controlnet-depth-sdxl-1.0-mid'' \cite{vonPlaten2022diffusers}. This choice strikes a balance between computational resource demands and the effectiveness of spatial conditioning.
An aspect of our generator $g$ is its  invariance to translation by a constant across the last channel due to the softmax operation. Indeed, given a constant $a$, the relation $g(\theta) = g(\theta + a)$ holds because $ \frac{e^{\lambda_{i, j, k} + a}}{\sum_{l=0}^{n-1}{e^{\lambda_{i, j, l} + a}}} = \frac{e^{\lambda_{i, j, k}}}{\sum_{l=0}^{n-1}{e^{\lambda_{i, j, l}}}} = \pi_{i, j, k}$. Therefore, we center the weights around zero in every iteration, ensuring they remain within a reasonable range to avoid floating point precision issues.

\subsection{Parameters}

The execution of our method typically involves 6000 to 10000 epochs, translating to a runtime of approximately 1.5 to 2.5 hours on an Nvidia RTX 4090 GPU. The standard parameters are set as follows: the temperature parameter of the Gumbel-Softmax reparameterization $\tau$ is fixed at 1, the guidance scale $s$ at 40, and the FFT loss weight $w_{\mathit{FFT}}$ at 20. Unless specified otherwise, the default image size for the presented results is $64 \times 64$, and the norm used for initializing the parameter of the generator is the $L_1$ norm (see main paper, Sec. 4.2).
We use PyTorch \cite{paszke2019pytorch}, with the generator weights optimized using the AdamW optimizer \cite{loshchilov2019decoupled}. The optimization process follows a constant learning rate of 0.25, starting with a warm-up phase of 250 steps.
Although backpropagation through an argmax function can be realized by duplicating the gradient from the softmax operation \cite{esser2021taming}, our method primarily employs the softmax function. To approximate an argmax-like behavior, we can reduce the value of $\tau$, see details in a comparative analysis in \secref{sec:tauAbl}.

Image augmentations are applied randomly with the following probabilities: grayscale (0.2), horizontal flip (0.5), perspective distortion (0.5) with a distortion scale of 0.3.
The scales for both the Canny edge and depth map ControlNet conditioning are set uniformly at 0.35. We apply Gaussian blur with a radius of 1 pixel to the Canny edge detection for smoothing effects.
Finally, the uniform sampling of the parameter $t \sim \mathcal{U}(a, b)$ starts with $a = 20$ and $b = 980$, with $b$ linearly decreasing to $800$ at the midpoint of our method's execution, and staying constant afterwards. This time step annealing strategy is inspired from \cite{yu2023textto3dCSD}.

%% file: sections/supplementary/2_ablation.tex
\section{Parameter comparisons}

\begin{figure*}[t]
    \centering
	\setlength{\tabcolsep}{1pt}
 \small
    \begin{tabular}{cccccccccccc}
    \raisebox{20pt}[0pt][0pt]{$s  \nabla_{\theta}  \mathcal{L}_\mathit{Sem}$}  & \includegraphics[width=0.075\linewidth]{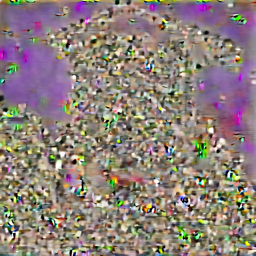} & \includegraphics[width=0.075\linewidth]{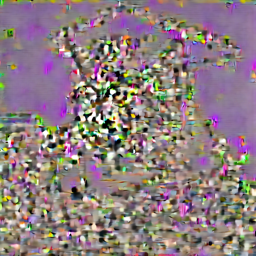}& \includegraphics[width=0.075\linewidth]{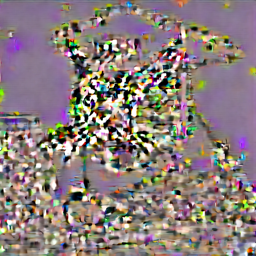}& \includegraphics[width=0.075\linewidth]{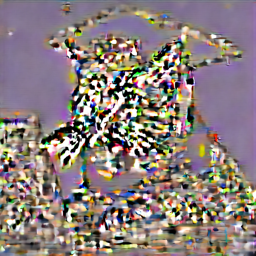}& \includegraphics[width=0.075\linewidth]{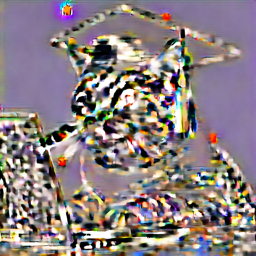}& \includegraphics[width=0.075\linewidth]{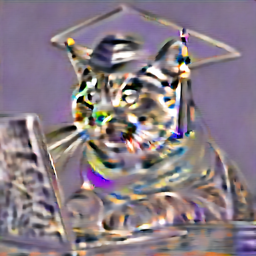}& \includegraphics[width=0.075\linewidth]{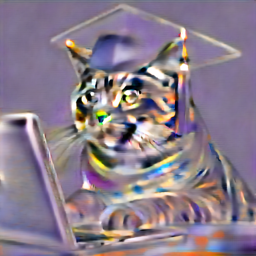}& \includegraphics[width=0.075\linewidth]{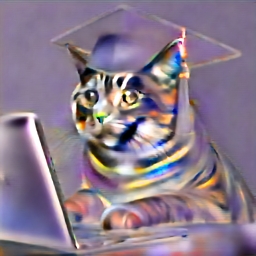}& \includegraphics[width=0.075\linewidth]{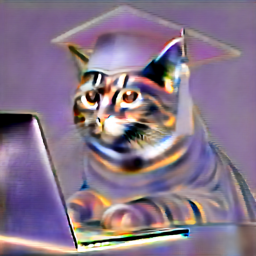}& \includegraphics[width=0.075\linewidth]{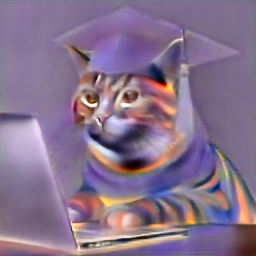}& \includegraphics[width=0.075\linewidth]{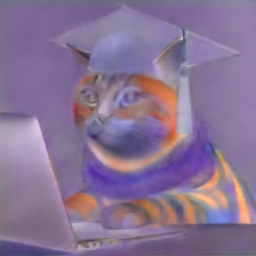}\\
    
    \raisebox{20pt}[0pt][0pt]{$\nabla_{\theta}  \mathcal{L}_\mathit{Noise}$}  & \includegraphics[width=0.075\linewidth]{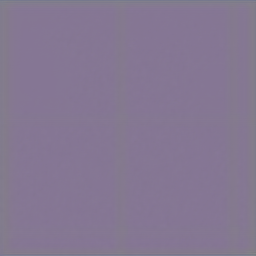} & \includegraphics[width=0.075\linewidth]{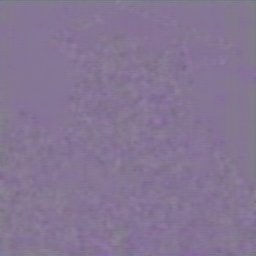}& \includegraphics[width=0.075\linewidth]{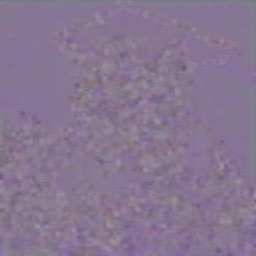}& \includegraphics[width=0.075\linewidth]{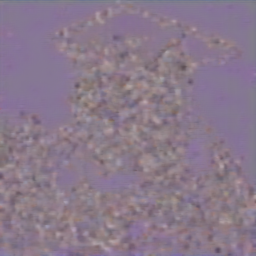}& \includegraphics[width=0.075\linewidth]{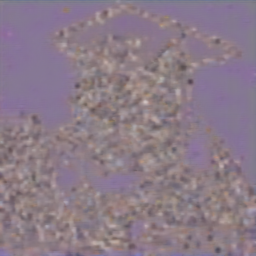}& \includegraphics[width=0.075\linewidth]{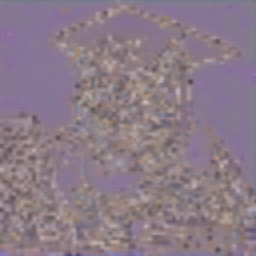}& \includegraphics[width=0.075\linewidth]{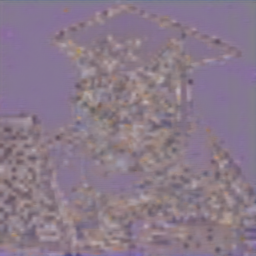}& \includegraphics[width=0.075\linewidth]{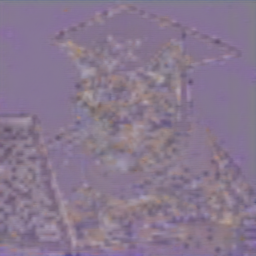}& \includegraphics[width=0.075\linewidth]{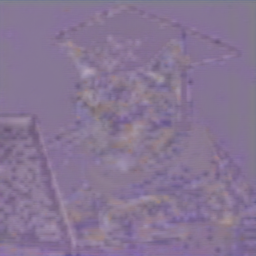}& \includegraphics[width=0.075\linewidth]{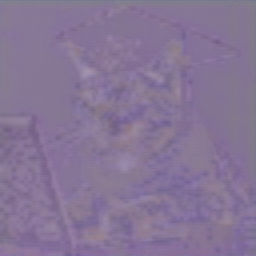}& \includegraphics[width=0.075\linewidth]{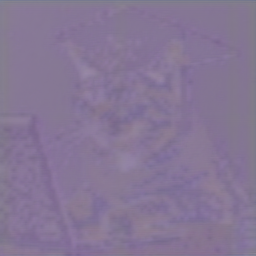}\\

\raisebox{21pt}[0pt][0pt]{\begin{minipage}{.09\linewidth}
\[
\begin{array}{l}
        \nabla_{\theta} \mathcal{L}_\mathit{Noise} \\
        + s  \nabla_{\theta}  \mathcal{L}_\mathit{Sem}
\end{array}
\]
\end{minipage}}
    & \includegraphics[width=0.075\linewidth]{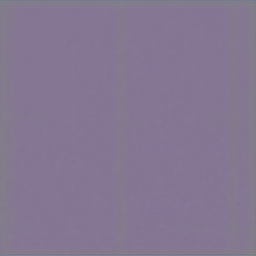} & \includegraphics[width=0.075\linewidth]{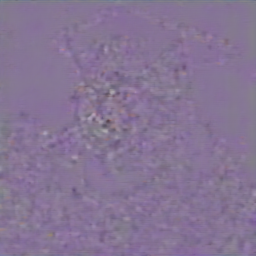}& \includegraphics[width=0.075\linewidth]{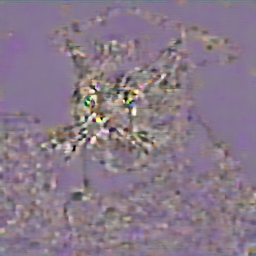}& \includegraphics[width=0.075\linewidth]{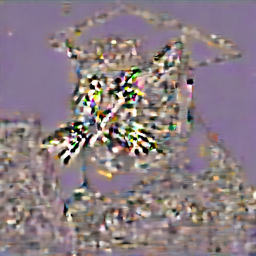}& \includegraphics[width=0.075\linewidth]{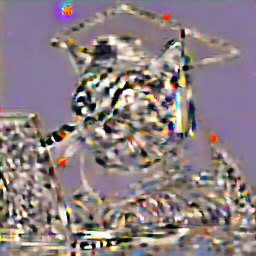}& \includegraphics[width=0.075\linewidth]{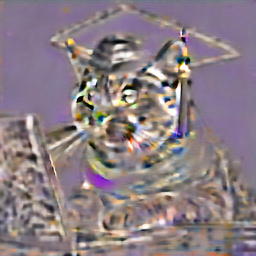}& \includegraphics[width=0.075\linewidth]{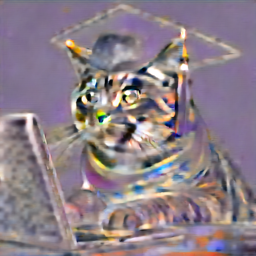}& \includegraphics[width=0.075\linewidth]{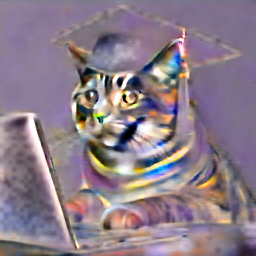}& \includegraphics[width=0.075\linewidth]{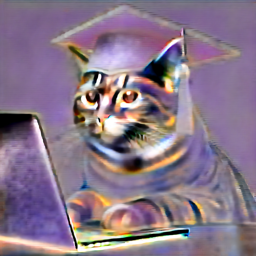}& \includegraphics[width=0.075\linewidth]{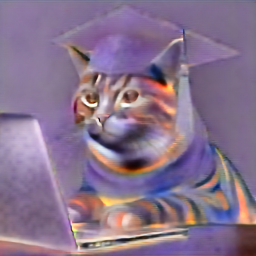}& \includegraphics[width=0.075\linewidth]{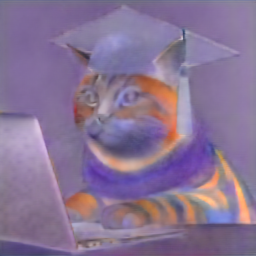}\\
    timestep $t$ & 10 & 100 & 200 & 300 & 400 & 500 & 600 & 700 & 800 & 900 & 999
    \end{tabular}
    \caption{Illustration of the latent score distillation sampling loss with fixed guidance scale $s=40$. The first row represents the semantic loss, while the second row is the noise loss. The full latent score distillation sampling loss $\mathcal{L}_{\mathit{LSDS}}$ is shown on the third row. Note that this illustration is computed by using the latent decoder of the latent diffusion model on the loss gradient, which is not a linear operation.}
    \label{fig:semLoss}
\end{figure*}

\begin{figure*}[t]
    \centering
	\setlength{\tabcolsep}{1pt}
    \begin{tabular}{cccccccccc}
    \includegraphics[width=0.0915\linewidth]{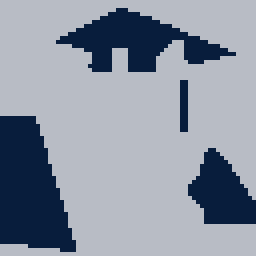}
    & \includegraphics[width=0.0915\linewidth]{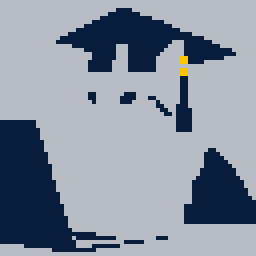}
    & \includegraphics[width=0.0915\linewidth]{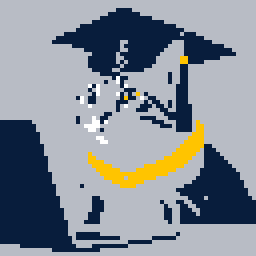}
    & \includegraphics[width=0.0915\linewidth]{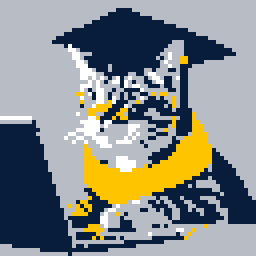}
    & \includegraphics[width=0.0915\linewidth]{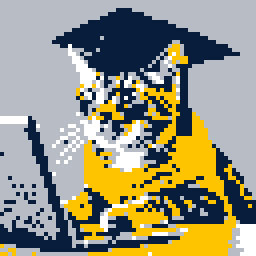}
    & \includegraphics[width=0.0915\linewidth]{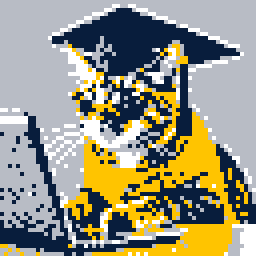}
    & \includegraphics[width=0.0915\linewidth]{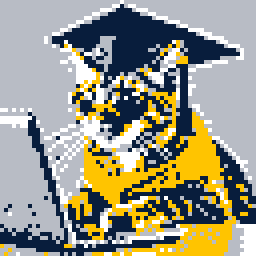}
    & \includegraphics[width=0.0915\linewidth]{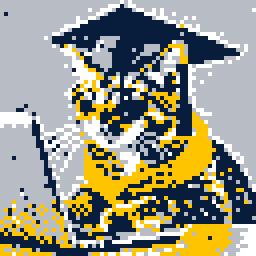}
    & \includegraphics[width=0.0915\linewidth]{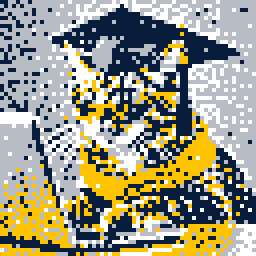}
    & \includegraphics[width=0.0915\linewidth]{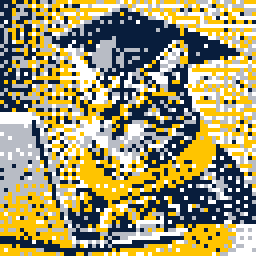}
    \\
    $s$: 0 & 7.5 & 15 & 25 & 40 & 55 & 75 & 100 & 200 & 500 
    \end{tabular}
    \caption{Illustration of the influence of the guidance scale $s$. For this experiment, the FFT loss was not used, i.e. $w_{\mathit{FFT}} = 0$.}
    \label{fig:noiseloss}
\end{figure*}

\begin{figure*}[t]
    \centering
	\setlength{\tabcolsep}{1pt}
    \begin{tabular}{cccccccc}
    \includegraphics[width=0.1125\linewidth]{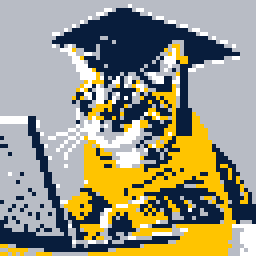}&  \includegraphics[width=0.1125\linewidth]{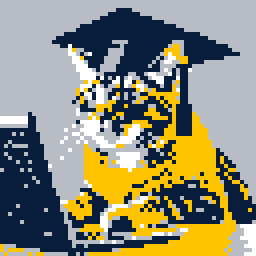}& \includegraphics[width=0.1125\linewidth]{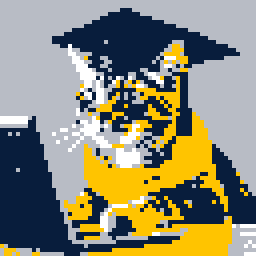}& \includegraphics[width=0.1125\linewidth]{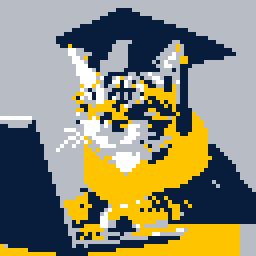}& \includegraphics[width=0.1125\linewidth]{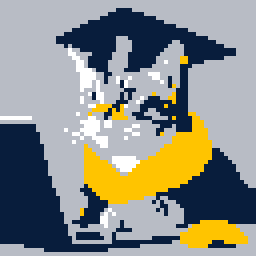}& \includegraphics[width=0.1125\linewidth]{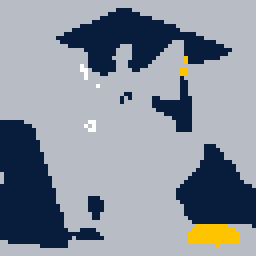}& \includegraphics[width=0.1125\linewidth]{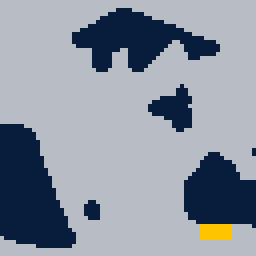}& \includegraphics[width=0.1125\linewidth]{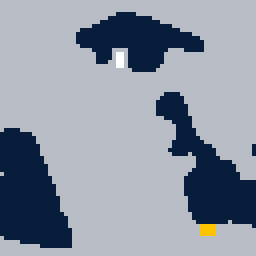} \\
    $w_{\mathit{FFT}}$: 0 & 25 & 50 & 100 & 200 & 500 & 1000 & 2500
    \end{tabular}
    \caption{Illustration of the influence of the Smoothness Loss $\mathcal{L}_{\mathit{FFT}}$ with increasing smoothness loss weight $w_{\mathit{FFT}}$ on the final image.}
    \label{fig:fftloss}
\end{figure*}

In this section, we conduct an indicative comparison to justify the choice of our default parameters. We also propose to analyze how each parameter can be combined to achieve different effects.

\subsection{Input}
\label{sec:InputApp}

For our experiments, we decided to use an image of a cat generated via a state-of-the-art diffusion model \cite{BetkerImprovingIG} as  input. \figref{fig:inputAbl} shows the input image, its Canny edge and depth map prediction. We also show the initialized state of the generator, both with the argmax- and softmax-generation, as explained in Sec.\ 5.1 in the main paper. We also showcase the result obtained with our default parameters.

\subsection{Loss functions}
\label{sec:LossFuncAppend}

We recall that our loss function can be written as a sum of three terms:
\begin{align}
    \nabla_{\theta} \mathcal{L} = \nabla_{\theta}  \mathcal{L}_\mathit{Noise} + s  \nabla_{\theta}  \mathcal{L}_\mathit{Sem} + w_\mathit{FFT} \nabla_{\theta}\mathcal{L}_\mathit{FFT}.
\end{align}
We analyze the noise and semantic loss together, as they come from the latent score distillation sampling term, and the smoothness loss separately.

\subsubsection{Noise loss and semantic loss}

To show the respective role of the noise loss $\nabla_{\theta}  \mathcal{L}_\mathit{Noise}$ and the semantic loss $\nabla_{\theta}  \mathcal{L}_\mathit{Sem}$, we perform an ablation by setting $w_{\mathit{FFT}} = 0$ and varying the guidance scale $s$ in \figref{fig:noiseloss}. We notice that when the semantic loss $s$ is too small, the resulting final image is smoothed out. In contrast, too high values of the guidance scale leads to noisier and saturated results. 
While the semantic loss is responsible for the semantics-awareness of our optimization, the noise loss acts as a variance reduction term \cite{poole2022dreamfusion}. The visualization in \figref{fig:semLoss} corroborates this argument, showing that the semantic loss is responsible for the semantic details, while the noise loss prevents noisy sampling, especially for the lower time steps. The visualization is performed by simply using the latent decoder of the latent diffusion model \cite{madebyollin2023taesdxl} on the gradient of the loss functions.

\begin{figure*}[t]
    \centering
	\setlength{\tabcolsep}{1pt}
 \small
    \begin{tabular}{ccccccccccc}
    \includegraphics[width=0.085\linewidth]{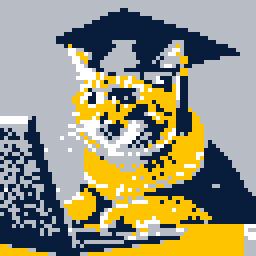}
    & \includegraphics[width=0.085\linewidth]{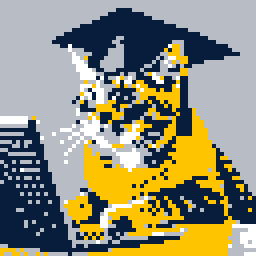}
    & \includegraphics[width=0.085\linewidth]{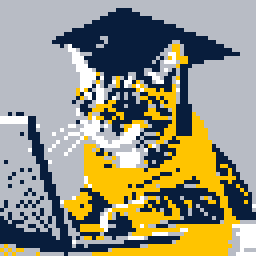}
    & \includegraphics[width=0.085\linewidth]{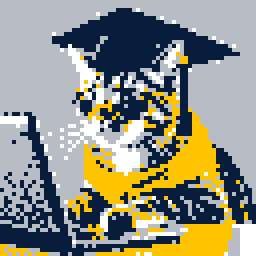}
    & \includegraphics[width=0.085\linewidth]{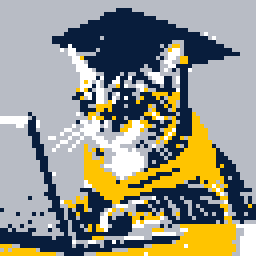}
    & \includegraphics[width=0.085\linewidth]{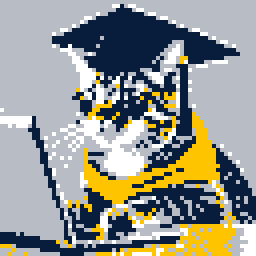}
    & \includegraphics[width=0.085\linewidth]{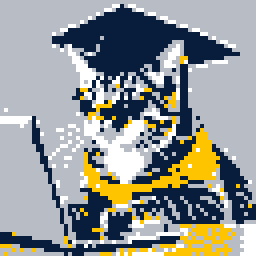}
    & \includegraphics[width=0.085\linewidth]{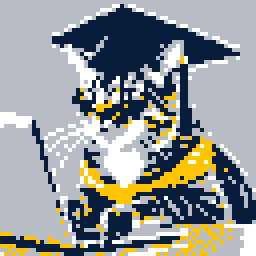}
    & \includegraphics[width=0.085\linewidth]{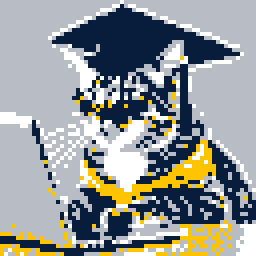}
    & \includegraphics[width=0.085\linewidth]{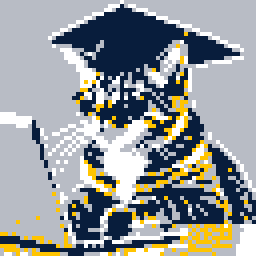}
    & \includegraphics[width=0.085\linewidth]{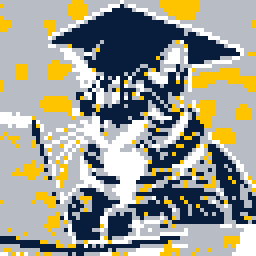}
    \\
    { $p_{\mathit{grayscale}}$}: 0.0 & 0.1 & 0.2 & 0.3 & 0.4 & 0.5 & 0.6 & 0.7 & 0.8 & 0.9 & 1..0
    \end{tabular}
    \caption{Depiction of how varying the probability of applying grayscale augmentation during the augmentation phase impacts our optimization process.}
    \label{fig:grayscale}
\end{figure*}

\begin{figure*}[t]
    \centering
	\setlength{\tabcolsep}{1pt}
 \small
    \begin{tabular}{cccccccccccc}
    \includegraphics[width=0.085\linewidth]{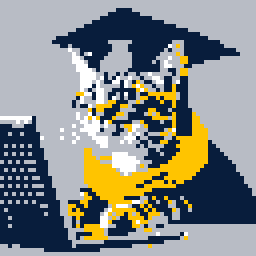}
    & \includegraphics[width=0.085\linewidth]{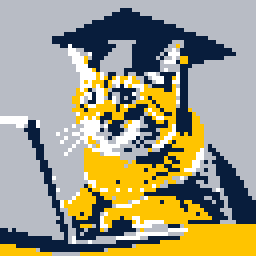}
    & \includegraphics[width=0.085\linewidth]{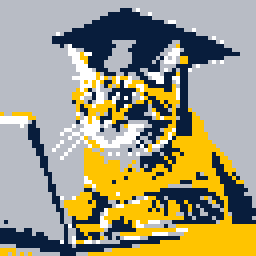}
    & \includegraphics[width=0.085\linewidth]{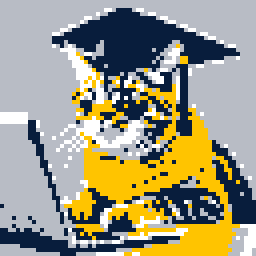}
    & \includegraphics[width=0.085\linewidth]{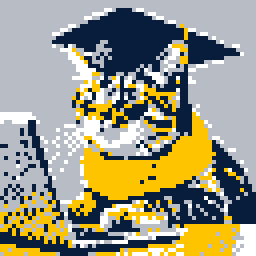}
    & \includegraphics[width=0.085\linewidth]{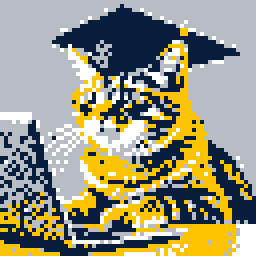}
    & \includegraphics[width=0.085\linewidth]{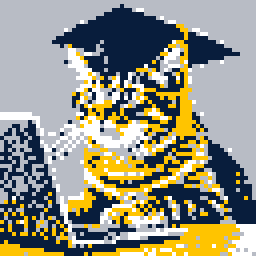}
    & \includegraphics[width=0.085\linewidth]{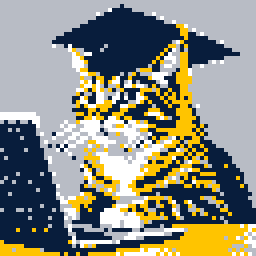}
    & \includegraphics[width=0.085\linewidth]{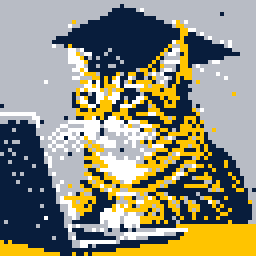}
    & \includegraphics[width=0.085\linewidth]{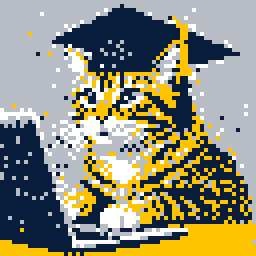}
    & \includegraphics[width=0.085\linewidth]{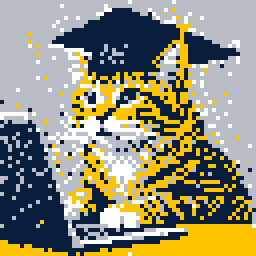}
    \\
    $d_s$:\ 0.0 & 0.1 & 0.2 & 0.3 & 0.4 & 0.5 & 0.6 & 0.7 & 0.8 & 0.9 & 1..0
    \end{tabular}
    \caption{Illustration of the influence of the perspective transformation distortion scale. To better show the influence of the random perspective, the probability of perspective transformation during the augmentation is set to $1.0$ for this experiment.}
    \label{fig:perspective}
\end{figure*}

\begin{figure*}[t]
    \centering
    \small
	\setlength{\tabcolsep}{1pt}
    \begin{tabular}{ccccccccccc}
   \includegraphics[width=0.085\linewidth]{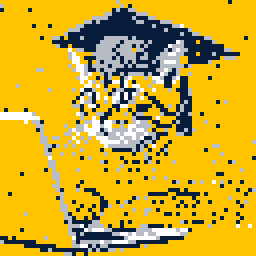}
   & \includegraphics[width=0.085\linewidth]{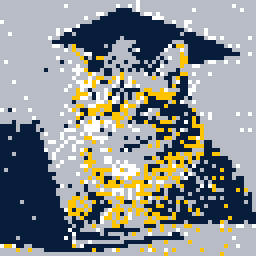}
   & \includegraphics[width=0.085\linewidth]{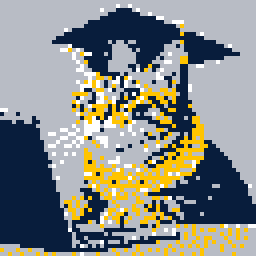}
& \includegraphics[width=0.085\linewidth]{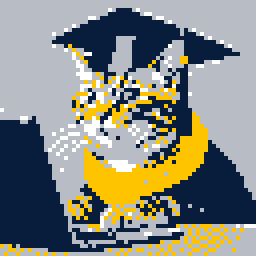}
& \includegraphics[width=0.085\linewidth]{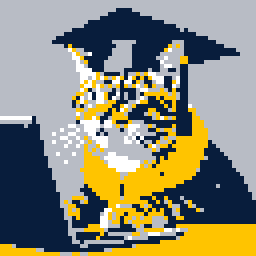}
& \includegraphics[width=0.085\linewidth]{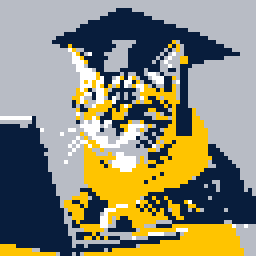}
& \includegraphics[width=0.085\linewidth]{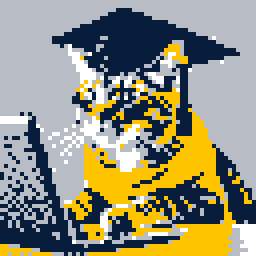}
& \includegraphics[width=0.085\linewidth]{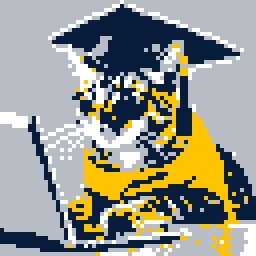}
& \includegraphics[width=0.085\linewidth]{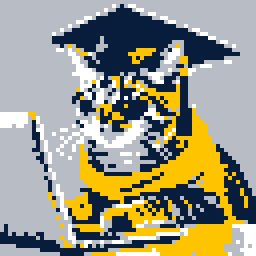}
& \includegraphics[width=0.085\linewidth]{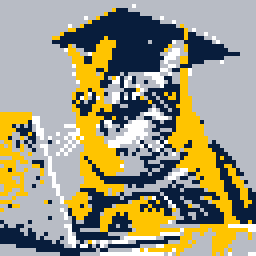}
& \includegraphics[width=0.085\linewidth]{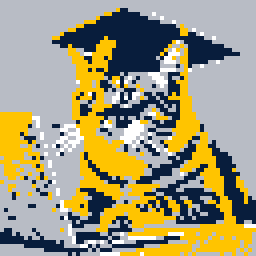}
\\
    argmax & $\tau$:\  0.05 & 0.1 & 0.25 & 0.5 & 0.75 & 1.0 & 1.5 & 2.0 & 3.0 & 4.0
    \end{tabular}
    \caption{We compare the effect of varying the magnitude of the temperature $\tau$ parameterizing the Gumbel-softmax operation. For the first column, we use an argmax operation during the optimization and copy the gradient from the softmax operation for backpropagation.}
    \label{fig:temperature}
\end{figure*}

\subsubsection{Smoothness loss}

The fast Fourier transforms serves the purpose of smoothing out the resulting image during the optimization process. While not central to our technique, it tends to improve the result and show that any classic loss can be used to redirect the generation of \ourmethod{} according to any objective. We demonstrate its influence in \figref{fig:fftloss}  by varying the weight $w_{\mathit{FFT}}$. We can see that the higher the weight, the fewer details appear in the generated image. While oversmoothed results are generally not desirable, the smoothness loss can act as a stylistic parameter for image abstraction purposes.

\subsection{Augmentation}

In our optimization, we randomly apply a grayscale filter and a perspective transformation. We show the results of applying these filters at different frequencies.

\subsubsection{Grayscale}

In \figref{fig:grayscale} we show a comparative demonstration of varying the probability (or the frequency) that the augmentation performs a grayscale filter before being fed to the denoiser. Interestingly, we can notice that applying a grayscale filter tends to tone down the color distribution of the output. When no grayscale filter is applied ($p_{\mathit{grayscale}} \approx 0$), the color tends to be saturated. Conversely, frequent use of the grayscale filter ($p_{\mathit{grayscale}} \approx 1$) restricts the optimization process to a monochromatic view of the input palette. In this scenario, the diffusion model only perceives color based on their contribution to the grayscale tones, leading to the prominence of yellow blotches in our example.

\subsubsection{Random perspective} 

The perspective transform simulates the effect of viewing the image from different angles. This transform is useful, as it introduces variation in the generated images, forcing the optimization to generalize its semantic update of features in images irrespective of their orientation or angle. The impact of this transformation is illustrated in \figref{fig:perspective}, where we observe that increased distortion scales lead to outputs more aligned with spatial conditioning, as the optimization is required to adjust its updates more broadly.
However, excessively high distortion scales introduce undesired noise into the final image.
In our experiments, we demonstrate the effects using a $1.0$ probability for applying the perspective transformation during optimization. However, in practical applications, we balance its impact by not applying random perspective transformations in every instance.

\begin{figure}[t]
    \centering
    \small
	\setlength{\tabcolsep}{1pt}
    \begin{tabular}{cc}
  \includegraphics[width=0.49\linewidth]{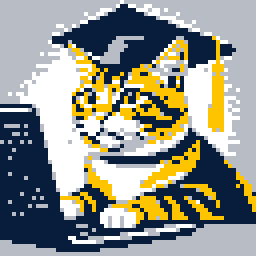} & \includegraphics[width=0.49\linewidth]{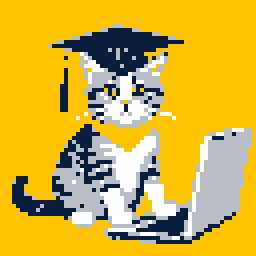}\\
    with input image initialization & random initialization\\
    \end{tabular}
    \caption{We compare the effect of initializing the network based on the input image with random initialization of the generator parameters. The results are computed without ControlNet conditioning, to better showcase the role of initialization in our method.}
    \label{fig:initialization}
    \Description{Initialization}
\end{figure}

\begin{figure*}[ht!]
	\centering
	\scriptsize
	\setlength{\tabcolsep}{2pt}
	\begin{tabular}{cccccccccc}
 \toprule
 input & nearest-neighbor & PIA & DUP & MYOS & VectorFusion & \textbf{\ourmethod{}} & initialization & \textbf{\ourmethod{}} \\
  & interpolation & \cite{Gerstner:2012:PIA} & \cite{2018ChuDeepUnsupervizedPixelization} & \cite{Wu2022MakeYourOwnSprite} & \cite{jain2022vectorfusion} & \emph{K-means} & \multicolumn{2}{c}{\emph{palette}}\\
  
  
  \includegraphics[width=0.095\linewidth]{figures/supplementary/comparisons/input/peppers.png}&
  \includegraphics[width=0.095\linewidth]{figures/supplementary/comparisons/nearest/peppers.png} & 
  \includegraphics[width=0.095\linewidth]{figures/supplementary/comparisons/PIA/peppers.png} &   
  \includegraphics[width=0.095\linewidth]{figures/supplementary/comparisons/DUP/peppers.png} &
  \includegraphics[width=0.095\linewidth]{figures/supplementary/comparisons/MYOS/peppers.png} & 
  \includegraphics[width=0.095\linewidth]{figures/supplementary/comparisons/VF/peppers.png}
  &
  \includegraphics[width=0.095\linewidth]{figures/supplementary/comparisons/kmeans/peppers.png} 
  &
  \includegraphics[width=0.095\linewidth]{figures/supplementary/comparisons/palette_init/peppers.png} & 
  \includegraphics[width=0.095\linewidth]{figures/supplementary/comparisons/palette_result/peppers.png}
  \\
  
  $512 \times 512$ & & & & & \multicolumn{4}{c}{An assortment of fresh bell peppers} \\
  & & & & & \multicolumn{4}{c}{with different shapes and sizes in a close-up view.} \\

  \includegraphics[width=0.095\linewidth]{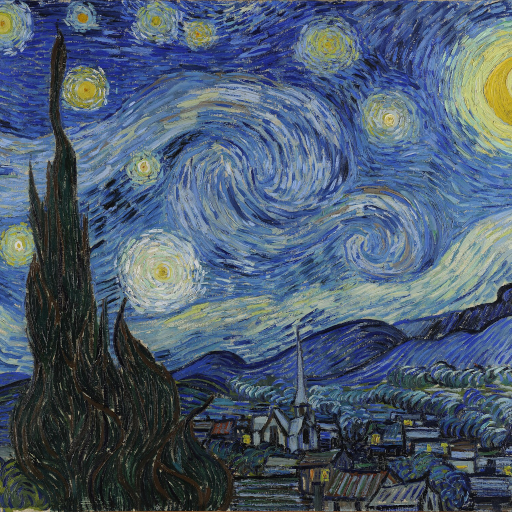}&
  \includegraphics[width=0.095\linewidth]{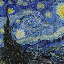} & 
  \includegraphics[width=0.095\linewidth]{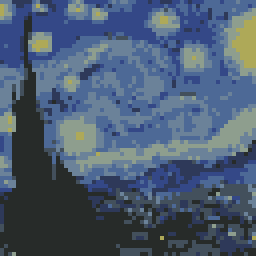} & 
  \includegraphics[width=0.095\linewidth]{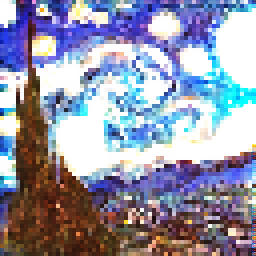}&
  
  \includegraphics[width=0.095\linewidth]{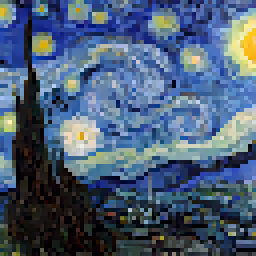} & 
  \includegraphics[width=0.095\linewidth]{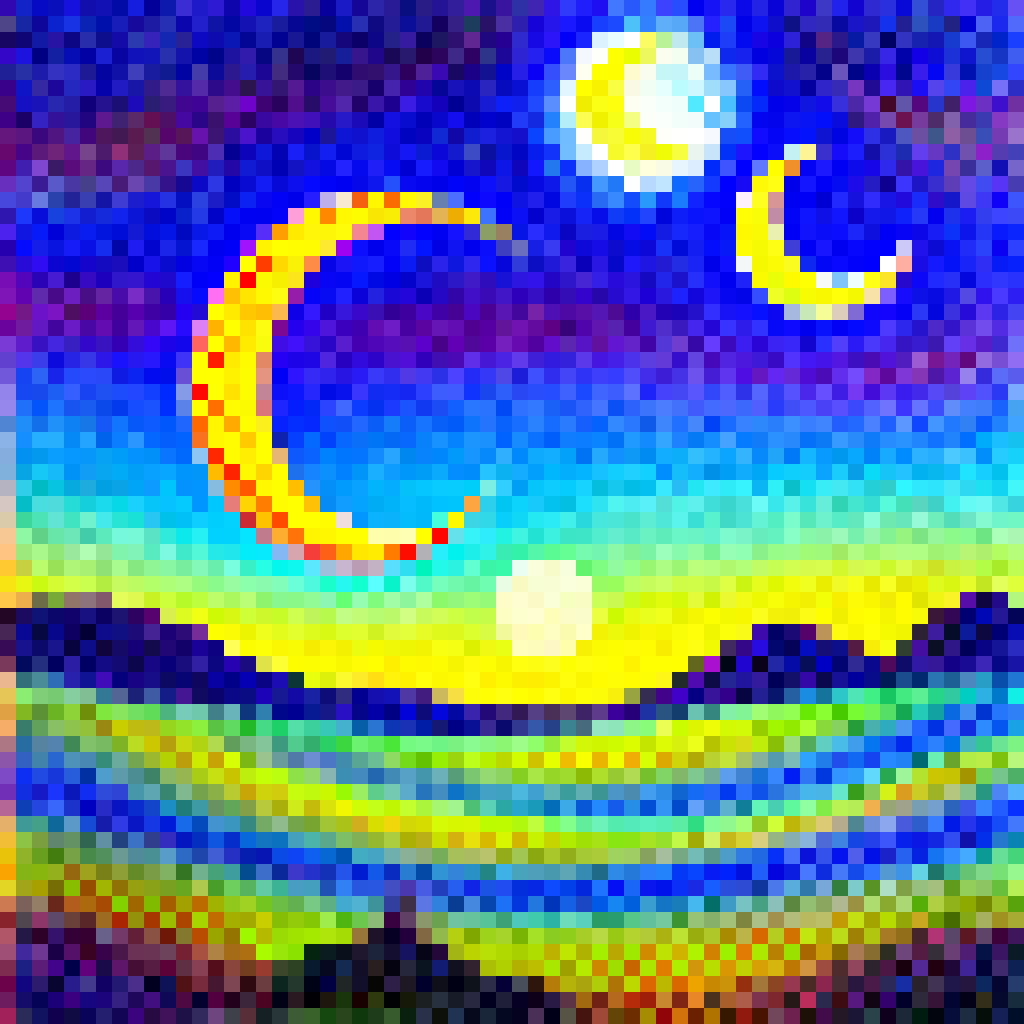}
  & 
  \includegraphics[width=0.095\linewidth]{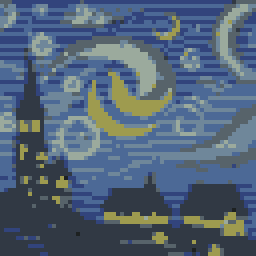} & 
  
  \includegraphics[width=0.095\linewidth]{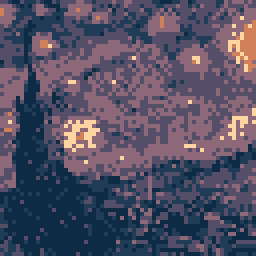} & 
  \includegraphics[width=0.095\linewidth]{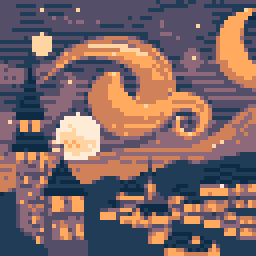} \\
  
  $512 \times 512$ & & & & & \multicolumn{4}{c}{Painting with swirling skies, a bright crescent moon, and} \\
  & & & & & \multicolumn{4}{c}{a peaceful village beneath a dynamic celestial sky.}
  \\

  \includegraphics[width=0.095\linewidth]{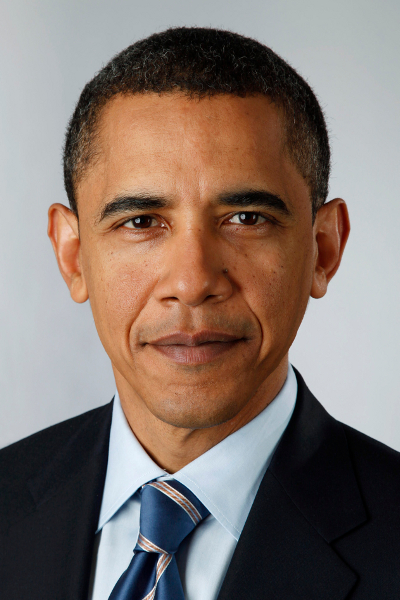}&
  \includegraphics[width=0.095\linewidth]{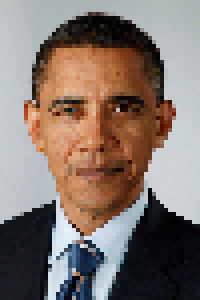} & 
  \includegraphics[width=0.095\linewidth]{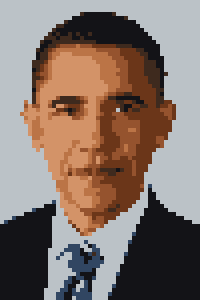}
  & 
  \includegraphics[width=0.095\linewidth]{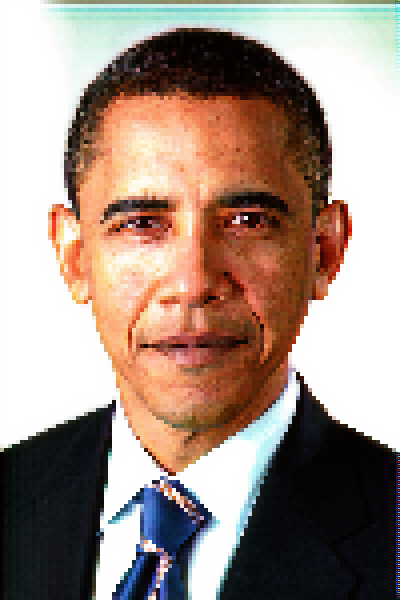}
  &
  \includegraphics[width=0.095\linewidth]{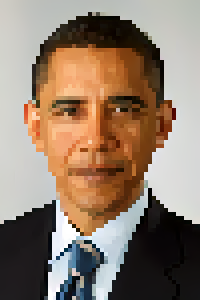} & 
  \includegraphics[width=0.095\linewidth]{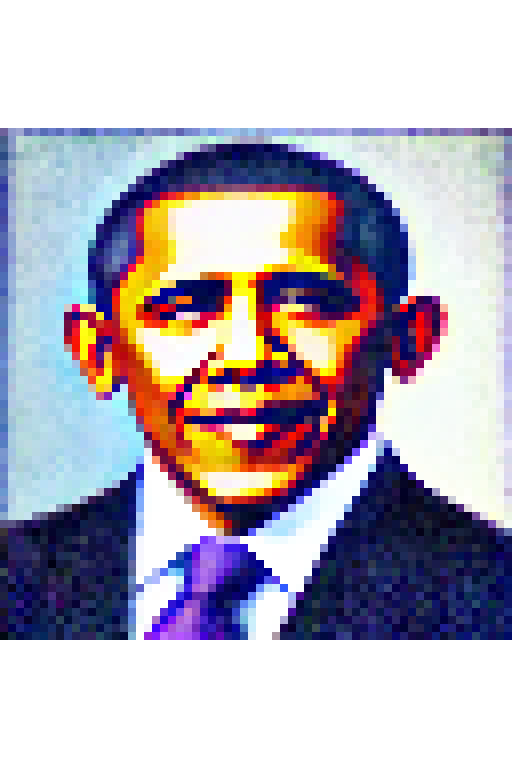}
  &
  \includegraphics[width=0.095\linewidth]{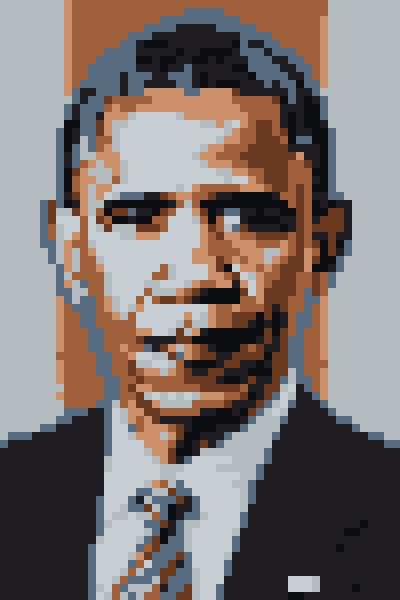} & 
  
  \includegraphics[width=0.095\linewidth]{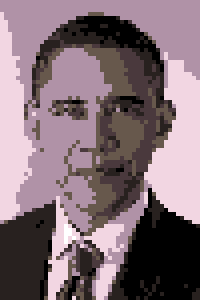} & 
  \includegraphics[width=0.095\linewidth]{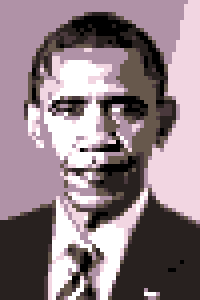} \\
  
 $400 \times 600$  & & & & & \multicolumn{4}{c}{A formal portrait of Obama, wearing a suit with a tie.} \\[2pt]
 \bottomrule
 \end{tabular}
 \caption{Visual comparison of various pixelization methods applied to images downscaled by a factor of 8. The input images are displayed on the left, with their sizes indicated below. Both VectorFusion and \ourmethod{} are initialized with the input image as their initial state, and conditioned on the prompt indicated below their respective results. Due to VectorFusion's limitation with non-square images, a $64 \times 64$ grid is employed for the examples in the last row. \ourmethod{} is demonstrated in two variations: one utilizing \emph{K-means} for color palette determination, and the other employing a challenging 5-color \emph{palette}. We also show the initialization with the palette, to demonstrate how our method differs from classic palette matching. While PIA and the K-means variant of \ourmethod{} operate within a 8-color limit, nearest-neighbor interpolation, DUP, MYOS, and VectorFusion have no such constraints on their color palettes and are not quantized.}
 \label{fig:visual_comparison}
\end{figure*}

\subsection{Temperature parameter $\tau$}
\label{sec:tauAbl}

The Gumbel-softmax is parameterized by a scalar $\tau$ that modulates its proximity to a categorical distribution \cite{jang2017categoricalGumbel, maddison2017concreteGumbel}. 
In \figref{fig:temperature}, we investigate the assertion that $\tau$ should remain within certain limits. As discussed in Sec.\ 4.1 of the main paper, lower values of $\tau$ enhance the resemblance of Gumbel-softmax to categorical sampling, but this leads to an increase in noise in the generated images. Even though our loss integrates a noise-preservation component, too noisy generated images result in imprecise parameter updates. 
Conversely, higher values of $\tau$ cause the Gumbel-softmax to approximate a uniform distribution, resulting in images with colors that are more uniform and smooth, with diminished semantic clarity. Although the optimization process is semantically driven and continues to adjust the generator parameters to align the image with the input prompt, excessively high values of $\tau$ can compromise the optimization's ability to capture color nuances. It is noteworthy that $\tau$ values in the range of $0.25$ to $2$ yield aesthetically pleasing results, although with distinct stylistic differences.

\subsection{Initialization}

Initialization can be important for generative techniques based on score distillation \cite{jain2022vectorfusion}. We show its effect in \figref{fig:initialization}. 
Without the use of ControlNet \cite{zhang2023addingControlNet}, the optimization process lacks access to spatial information from the input image, resulting in outcomes that can deviate from the original image.

\subsection{ControlNet weights}
\label{sec:ControlnetAbl}

In \figref{fig:ControlNetWeights}, we explore the impact of varying weight parameters within the ControlNet framework \cite{zhang2023addingControlNet}. This analysis reveals a natural progression: the absence of ControlNet results in outputs that are less faithful to the original input, which demonstrates the role of ControlNet in balancing strict adherence to the initial input spatial conditioning.
Interestingly, even without explicit ControlNet guidance, the initialization phase of our method inherently allows for a degree of control that influences the final outcome, as shown in \figref{fig:initialization}.

%% file: sections/supplementary/3_evaluation.tex
\begin{table}[hb!]
\caption{Evaluation through a perceptual study, highlighting the performance of our method (Ours) in comparison to PIA and VectorFusion through semantic, fidelity and aesthetics questions. Each column aggregates the rankings across all questions in a specific category, representing the percentage of participants who placed each method at the respective rank (1, 2, 3, or 4) for that category.
}
\footnotesize
\centering
\setlength{\tabcolsep}{2pt}
\begin{tabular}{lcccccccccccccc}
\toprule
Method & \multicolumn{4}{c}{Semantic} & & \multicolumn{4}{c}{Fidelity} & & \multicolumn{4}{c}{Aesthetics} \\
Rank & \cellcolor{green!25}$1$ & $2$ & $3$ &$4$ & & \cellcolor{green!25}$1$ & $2$ & $3$ &$4$& & \cellcolor{green!25}$1$ & $2$ & $3$ &$4$ \\ \midrule
PIA  &  24.6  &  23.0  &  31.2  &  21.2  & &  \cellcolor{green!25}\textbf{49.3} &  37.9  &  11.1  &  1.8  & &  25.5  &  26.0  &  33.1  &  15.5  \\
VectorFusion  &  22.0  &  15.0  &  16.1  &  46.9  & &  0.8  &  0.8  &  8.0  &  90.4  & &  17.5  &  12.1  &  19.5  &  50.8  \\
Ours-\emph{K-means}  &  \cellcolor{green!25}\textbf{36.2}  &  37.0  &  22.6  &  4.2  & &  47.1  &  48.0  &  4.6  &  0.2  & &  26.5  &  38.8  &  22.3  &  12.4  \\
Ours-\emph{palette}  &  17.1  &  25.0  &  30.1  &  27.7  & &  2.7  &  13.3  &  76.3  &  7.6  & &  \cellcolor{green!25}\textbf{30.5}  &  23.1  &  25.1  &  21.3  \\
\bottomrule
\end{tabular}
\label{tab:userstudyResults}
\end{table}

\begin{table}[hb!]
\small
\captionof{table}{First quartile (Q1), median (Med.) and interquartile range (IQR) of the results of our perceptual study, according to semantic similarity, fidelity to input image and aesthetic appeal.}
\centering
\setlength{\tabcolsep}{3pt}
\begin{tabular}{l ccc ccc ccc}
\toprule
Method & \multicolumn{3}{c}{Semantics} & \multicolumn{3}{c}{Fidelity} & \multicolumn{3}{c}{Aesthetics} \\
 & Q1 & Med.& IQR &  Q1 &Med.& IQR& Q1 &Med.& IQR \\ \midrule 
PIA & 2.0 & 3.0 & 1.0& 1.0 &2.0 & 1.0 & 1.0 & 2.0 & 2.0\\
VF & 2.0 & 3.0 & 2.0& 4.0 & 4.0 & 0.0 & 2.0 & 4.0 & 2.0\\
Ours-\emph{K-means}& 1.0 &  2.0 & 2.0& 1.0 & 2.0 & 1.0 & 1.0 & 2.0 & 2.0\\
Ours-\emph{palette}& 2.0 & 3.0 & 2.0& 3.0& 3.0 & 0.0 & 1.0 &2.0 & 2.0\\
\bottomrule
\end{tabular}
\label{tab:userstudyIQR}
\end{table}

\begin{table*}[hb!]
\caption{Quantiative evaluation of pixelization methods across various metrics. We compare pixelated image abstraction (PIA) \cite{Gerstner:2012:PIA}, two quantized variants of Make Your Own Sprites (MYOS) \cite{Wu2022MakeYourOwnSprite}, VectorFusion \cite{jain2022vectorfusion}, and three variants of \ourmethod{}: \emph{palette}, \emph{K-means}, and \emph{adaptive}. The evaluation metrics are grouped into three categories: semantic similarity with CLIPScore L/14 and Human Preference Score V2 (HPSV2); fidelity to the input image with the Peak Signal-to-Noise Ratio (PSNR) and the Structural Similarity Index Measure (SSIM); and aesthetic score with the LAION Aesthetic Predictor Scores \cite{schuhmann2023improved}. Scores are provided for images of sizes $32\times 32$, $48 \times 48$, and $64 \times 64$, along with their averages (Avg). MYOS does not provide non-integer scaling factor, and as input images have size $1024 \times 1024$, output images of size $48 \times 48$ are not provided. The higher the value, the better. Best scores across categories are in bold.}
\centering
\small
\setlength{\tabcolsep}{2pt}
\begin{tabular}{l@{\hspace{5pt}}c@{\hspace{5pt}}cccc@{\hspace{10pt}}cccc@{\hspace{10pt}}cccc@{\hspace{10pt}}cccc@{\hspace{10pt}}cccc}
\toprule
& & \multicolumn{8}{c}{Semantic} & \multicolumn{8}{c}{Fidelity} & \multicolumn{4}{c}{Aesthetics} \\
& & \multicolumn{4}{c}{CLIPScore L/14 $\uparrow$} & \multicolumn{4}{c}{HPSV2 $\uparrow$} & \multicolumn{4}{c}{PSNR $\uparrow$}& \multicolumn{4}{c}{SSIM $\uparrow$} & \multicolumn{4}{c}{Aesthetic Predictor $\uparrow$} \\ 
& & \multicolumn{4}{c}{\cite{radford2021learning}} & \multicolumn{4}{c}{\cite{wu2023hpsv2}} & \multicolumn{4}{c}{\cite{PSNR2010}} & \multicolumn{4}{c}{\cite{SSIM}} & \multicolumn{4}{c}{\cite{schuhmann2023improved}} \\ 
\cmidrule(lr){3-6} \cmidrule(lr){7-10} \cmidrule(lr){11-14} \cmidrule(lr){15-18} \cmidrule(lr){19-22}
\multicolumn{2}{l}{Method} & 32 & 48 & 64 & Avg & 32 & 48 & 64 & Avg & 32 & 48 & 64 & Avg & 32 & 48 & 64 & Avg & 32 & 48 & 64 & Avg\\ 
\midrule
\multicolumn{2}{l}{PIA} 
& 21.1 & 23.5 & 25.4 & 23.3 
& 0.240 & 0.248 & 0.254 & 0.247 
& 14.1 & \textbf{15.0} & 15.6 & 14.9 
& 0.455 & 0.485 & 0.487 & 0.476 
& 4.47 & 4.59 & 4.85 & 4.64 \\
\addlinespace[2pt]
\multirow{2}{*}{MYOS} & \emph{palette} 
& 18.6 & --- & 21.7 & 21.2 
& 0.231 & --- & 0.239 &  0.235
& 12.4 & --- & 13.7 & 13.1
& 0.407 & --- & 0.418 & 0.412
& 4.38 & --- & 5.04 & 4.71 \\
 & \emph{K-means} 
 & 20.7 & --- & 26.1 & 23.4 
 & 0.245 & --- & 0.261 & 0.253 
 & \textbf{15.2} & --- & \textbf{16.9} & \textbf{16.0} 
 & \textbf{0.467} & --- & \textbf{0.511} & \textbf{0.489} 
 & 4.52 & --- & 4.99 & 4.76 \\
\addlinespace[2pt]
\multicolumn{2}{l}{VectorFusion} 
& \textbf{24.5} & \textbf{25.2} & 27.0 & 25.6 
& \textbf{0.250} & 0.256 & 0.261 & \textbf{0.256} 
& \phantom{0}8.0 & \phantom{0}8.2 & \phantom{0}8.3 & \phantom{0}8.2 
& 0.293 & 0.292 & 0.258 & 0.281 
& 4.16 & 4.59 & 4.86 & 4.54 \\
\addlinespace[2pt]
\multirow{3}{*}{\ourmethod{}} & \emph{palette} 
& 23.1 & 22.6 & 24.5 & 23.4 
& 0.237 & 0.247 & 0.244 & 0.243 
& 10.3 & 11.7 & 11.3 & 11.1 
& 0.388 & 0.444 & 0.398 & 0.410 
& \textbf{4.83} & 4.46 & 5.13 & 4.81 \\
 & \emph{K-means} 
 & 22.7 & 24.0 & 25.8 & 24.2 
 & 0.235 & 0.251 & 0.250 & 0.245 
 & 12.9 & 14.0 & 13.9 & 13.6 
 & 0.442 & \textbf{0.486} & 0.451 & 0.460 
 & 4.78 & \textbf{4.66} & \textbf{5.31} & \textbf{4.92} \\
 & \emph{adaptive} 
 & 23.9 & 25.0 & \textbf{28.4} & \textbf{25.8} 
 & 0.238 & \textbf{0.261} & \textbf{0.263} & 0.254 
 & \phantom{0}8.3 & \phantom{0}8.7 & \phantom{0}9.5 & \phantom{0}8.8 
 & 0.333 & 0.373 & 0.374 & 0.360 
 & 3.84 & 4.21 & 4.79 & 4.28\\
\bottomrule
\end{tabular}
\label{tab:metrics}
\end{table*}
\section{Evaluation}

In this section, we show several comparisons and evaluations of \ourmethod{} with different pixelization methods. First, we propose a visual comparison over several images with diverse styles. Second, we present a quantitative evaluation to assess the pixelization quality through three indirect measures, namely how the result image aligns with the input prompt (semantic), how the result image is alike the input image (fidelity), how aesthetically pleasant the result image looks (aesthetics). Finally, we present the results of a perceptual study over 56 participants that further corroborates the results of the quantitative evaluation.

\subsection{Visual comparison}
\label{sec:visualComp}

Our visual comparison includes pictures of common food, of a person, as well as a painting. The resolution of each image is reduced by a factor of eight during the pixelization process. Results are shown in \figref{fig:visual_comparison}. We compare with various pixelization techniques, namely Pixelated Image Abstraction (PIA) \cite{Gerstner:2012:PIA}, Deep unsupervised pixelization (DUP) \cite{2018ChuDeepUnsupervizedPixelization}, Make Your Own Sprites (MYOS) \cite{Wu2022MakeYourOwnSprite}, and VectorFusion \cite{jain2022vectorfusion}.

Nearest-neighbor interpolation assigns the color value of the nearest original pixel to each pixel in the downscaled image, often resulting in a blocky or pixelated appearance. This method does not produce a color-quantized image, and its lack of semantic and geometric awareness can impair readability. In contrast, PIA \cite{Gerstner:2012:PIA} improves results through spatial and color space clustering, but its lack of semantic understanding can sometimes result in images that are difficult to interpret, such as the character's face and tie.

We also explore state-of-the-art neural techniques. DUP \cite{2018ChuDeepUnsupervizedPixelization} is retrained on its own dataset, and we use pretrained weights for MYOS \cite{Wu2022MakeYourOwnSprite}. For MYOS, we noticed that downscaling the input to divide the size of the image by two prior to using the neural network with cell size of 4 yields better results than directly using a cell size of 8. We therefore decided to present this output in our visual comparison. The primary content of the dataset used for training these neural methods is clip art images, which introduces a domain generalization issue. For example, DUP's outputs on photographic images tend to be overly saturated. MYOS, on the other hand, does not precisely downscale but rather employs ``cell-aware'' techniques based on a cell size of 4, as described in \cite{Wu2022MakeYourOwnSprite}, which can lead to uneven pixelization in practice. This also constrains the pixelization to a finite range (2 to 8 in the case of MYOS). As these methods do not enforce a uniform color within each cell, they ultimately rely on nearest-neighbor downsampling to achieve a distinct pixelized effect.

VectorFusion \cite{jain2022vectorfusion} segments the image into an even square grid and employs a differentiable vector graphics renderer \cite{Li:2020:DVG}. The technique uses score distillation to optimize the renderer parameters, specifically the colors of each cell. Contrary to our method, the color space of VectorFusion output is never constrained. We observe that the output tends to be saturated, suggesting that employing a color palette mitigates the saturation effect typically associated with score distillation methods. Starting with the input image as a base, results bear a general resemblance to the original, yet exhibit notable divergence in certain areas. This is visible in all examples, particularly in the positioning of the bell peppers and the global composition of the painting (\figref{fig:visual_comparison}). This legitimates the use of ControlNet \cite{zhang2023addingControlNet} to encourage more similarity with the input image, which is further validated by the outcomes achieved with \ourmethod{}.

We introduce two variations of our method: the first employs K-means to create a 8-color palette through color space clustering. The second utilizes challenging color palettes, and to demonstrate the distinctiveness of our approach, we also present its initialization. This illustrates the progression from a basic palette association to the final output. Moreover, it shows that a simple downscaling followed by a palette transfer does not provide convincing results, justifying our approach based on an optimization process using an image generator constrained to the color palette. \ourmethod{} consistently delivers outputs that more closely mirror the input image than VectorFusion, while effectively capturing the semantic core of the prompt. This is particularly visible in the case of the painting (second row of \figref{fig:visual_comparison}), where the overall structure of the original image is kept in its pixelized version, but with some divergence in details that conveys meaning more aligned with the input prompt. On this specific example, MYOS achieves a result that is closer to the input image, and VF a result that aligns more to the input prompt, while our method reaches a middle ground between both objectives. Remarkably, this equilibrium is maintained despite the constraints of a limited color palette and substantial color variations in the input image.

\subsection{Quantitative evaluation}
\label{sec:quantitative}

We present a quantitative evaluation of pixelization methods, which assesses fidelity to the input image, semantic similarity and aesthetics qualities of different pixelizations methods. To that effect, we first sample 150 prompts from PartiPrompts \cite{yu2022scalingpartiprompts}. We divide them equally into 3 categories for different target sizes: $32\times32$, $48\times48$, and $64\times 64$. Each set of 50 prompts is sampled according to this allotment: 10 from the 'Animal' category, 10 from the 'People' category, 10 from the 'Outdoor Scenes' category, 5 from the 'Indoor Scenes' category, 5 from the 'Food \& Beverages' category, 5 from the 'Vehicles' category, and 5 from the 'Produce \& Plants' category. We order the prompts by length, associating the shorter prompts to the smallest size, and the longer, more detailed ones to the $64\times 64$ category. To avoid recurrence of the same concept (e.g., the same animal), we resort to a large language model (LLM) \cite{brown2020language} to further filter out repetitive prompts. In some instances, we remove color mentions, to avoid further bias of the results, favoring methods that do not offer a quantization of the color space.
Subsequently, we produce input images corresponding to these prompts using SDXL \cite{podell2023sdxl}, employing a CLIPScore-based \cite{radford2021learning} rejection sampling technique. This involves generating six images and selecting the one with the highest CLIPScore. In instances where SDXL fails to produce an image of satisfactory quality, we turn to DALL·E 3 \cite{BetkerImprovingIG} for additional image generation. The prompts and generated images can be found in the supplementary material.

We quantitatively compare PIA \cite{Gerstner:2012:PIA}, quantized MYOS \cite{Wu2022MakeYourOwnSprite}, and VectorFusion \cite{jain2022vectorfusion} with three variants of our method. The \emph{palettes} variant utilizes a palette randomly sampled from 34 options available on the Lospec website \cite{lospec}. The \emph{K-means} variant employs a color palette determined through K-means clustering of the input color space, with both K-means and PIA using 8 colors for a fair comparison. The \emph{adaptive} variant optimizes the color palette alongside the generator weights without Gumbel sampling, creating an output image that is not quantized and has an unconstrained color space, offering a fairer comparison with VectorFusion. This method also uses 8 colors to ensure the entire color space can be covered by a convex sum of input colors. MYOS only provides pixelization with integer factor. To compare with this method, we first downscale the input image with bicubic interpolation to four times the target output resolution, and then run MYOS with a cell size of $4$, followed by nearest neighbour downscaling to the target output resolution. Because there is no integer pixelization factor for $48 \times 48$ images, we do not present results for this pixelization scale. We then apply two quantizations: a K-means color clustering with 8 colors, and a palette transfer with libimagequant \cite{libimagequant}. The palette used is the same as the palette of the corresponding image produced with \ourmethod{}-palette. Both VectorFusion and \ourmethod{} are run for 6000 steps and initialized with the input image. A selection of the results from \ourmethod{} with palettes and K-means on this dataset is shown in \figref{fig:bigfigure2}.

We use several metrics divided into three categories: semantic, fidelity and aesthetic. We show the results in \tableref{tab:metrics} individually for each size, and the average over the 150 examples.
The semantic evaluation includes CLIPScore L/14 similarity \cite{radford2021learning}, determined by calculating the mean cosine similarity between the CLIP embeddings of the generated images and their respective text captions, and the human preference score \cite{wu2023hpsv2}, a model that can predict human preferences on prompt-images pairs. We can see that VectorFusion and \ourmethod{}-\emph{adaptive} achieves the same score for semantic score, while \ourmethod{}-\emph{K-means} and \emph{palette} are slightly below, because of the constraints imposed by the color quantization. However, their CLIPScore tends to be higher than MYOS and PIA's, which are not driven semantically at all. 

The fidelity to the input image is measured via the peak signal-to-noise ratio (PSNR) and the structural similarity index measure (SSIM) \cite{SSIM}, two classic methods for image similarity measures \cite{PSNR2010}. On this metric, MYOS and PIA generally achieve the best score, as they are entirely driven by their similarity to the input images. Our \ourmethod{}-\emph{K-means} closely follows. In contrast, the lack of spatial conditioning in VectorFusion can make the output greatly diverge from the input, lowering its fidelity score.

Although it is hard to quantitatively evaluate the aesthetics of an image, we use the LAION aesthetic classifier \cite{schuhmann2023improved}, a model specifically trained on human preference to assign aesthetic scores to images. We observe that the \emph{palette} and \emph{K-means} variants of \ourmethod{} outperform PIA, MYOS and VectorFusion on this aesthetics measure. Despite color quantization posing difficulties for semantic or fidelity analysis, the high aesthetic scores for \ourmethod{}-\emph{palette} variant can be attributed to the harmonious color combinations, a feature that is lacking in VectorFusion, which performs poorly on this metric.

While all these metrics have a positive correlation with the quality of good pixelization, none of the metrics assesses directly the pixelization quality itself. To support this argument, we ran bilinear downscaling on our dataset and computed the metrics on the fidelity measure. While this technique is not well-adapted for pixelization purpose, the average PSNR is 16.8, and the average SSIM is 0.512, which are substantially higher than pixelization methods. This is expected, as bilinear interpolation is more faithful to the input image than pixelization method, but it shows that this measure alone is not an indicator of good pixelization. We interpret the results as such: CLIP-based methods like VectorFusion produce results that are more aligned semantically, MYOS and PIA produce the results that are the closest to the input, and \ourmethod{} achieves a satisfying compromise between the two different metrics, while being more aesthetically pleasing overall.

\subsection{Perceptual study}

To mitigate potential bias of the networks used to quantitatively evaluate our method, we conducted a comprehensive perceptual study to compare the effectiveness of PIA, VectorFusion, and \ourmethod{} with \emph{K-means} and \emph{palettes}. A total of 56 participants were recruited for the study. The survey presents participants with a series of images processed using the four different pixelization techniques.

\subsubsection{Design}

To design our perceptual study, we randomly sample 45 images from our quantitative evaluation dataset. Namely, we sample 15 images from each size category ($32\times 32$, $48\times 48$, and $64\times 64$). We further divide the perceptual study into three parts: image fidelity, prompt similarity, and aesthetics appeal. Each part contains 5 images from each size category. The exact questions and layout are shown in \figref{fig:userstudyquestions}, with the first question of each category. For each question, the order of the presented methods is randomized. We also show a visual representation of the answers in \figref{fig:userstudyquestions}, essentially visualising as histograms the results reported in \tableref{tab:userstudyResults}.

\subsubsection{Results}

Participants are asked to rank these 45 images based on specific criteria, such as semantic accuracy with respect to a prompt, fidelity to the input image, and aesthetic appeal. The perceptual study, as detailed in Table \ref{tab:userstudyResults}, offers a comparative evaluation of image pixelization methods across semantic, fidelity, and aesthetics categories. 
Notably, \ourmethod{}-\emph{K-means} excels in semantic integrity, leading with 36.2\% top-rank responses. PIA slightly dominates in fidelity, receiving a 49.3\% first-place rating against 47.1\% for \ourmethod{}-\emph{K-means}. In aesthetics, \ourmethod{}-\emph{palette} emerges as the preferred method with 30.5\% top-rank responses, indicating its superior aesthetic appeal. In contrast, VectorFusion (VF) consistently ranks lower across all categories, suggesting limitations in semantic preservation, fidelity, and aesthetic appeal. The discrepancy between the perceptual study results and the quantitative evaluation for VectorFusion's semantic similarity performance is likely due to a bias of the scoring network towards non-quantized images.

To enhance the interpretability of our results, we showcase the first quartile, the median and the interquartile range in \tableref{tab:userstudyIQR}. The analysis reveals a varied landscape of participant responses. While PIA and \ourmethod{}-\emph{K-means} demonstrate consistent performance across different categories, VF shows a higher variability in participant perceptions, particularly in the aesthetic domain: while it ranks last for more than half of the time,  it also sporadically achieves good aesthetics appreciation. This variability suggests that user opinions on VF's performance are polarized. In contrast, \ourmethod{}-\emph{palette} exhibits a balance between consistency and diversity in the feedback. Overall, the data indicates no single method excels in all aspects; each has unique strengths and weaknesses as perceived by the participants.

%% file: sections/supplementary/4_additional_results.tex
\begin{figure}[t]
	\centering
	\footnotesize
	\setlength{\tabcolsep}{1pt}
	\begin{tabular}{cccc}%
    \includegraphics[width=0.245\linewidth]{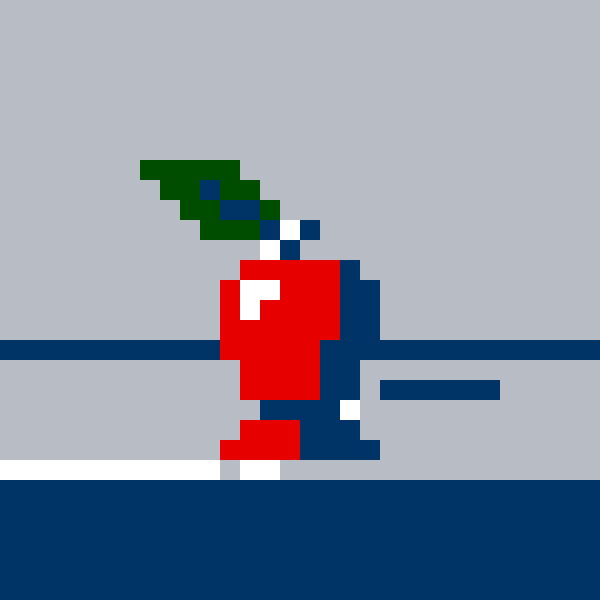}
     &
    \includegraphics[trim=0 0 0 0, clip, width=0.245\linewidth]{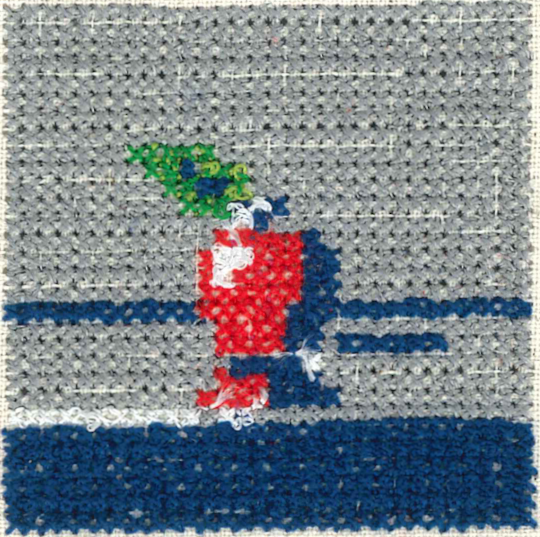} &
    \includegraphics[trim=0 0 0 0, clip, width=0.245\linewidth]{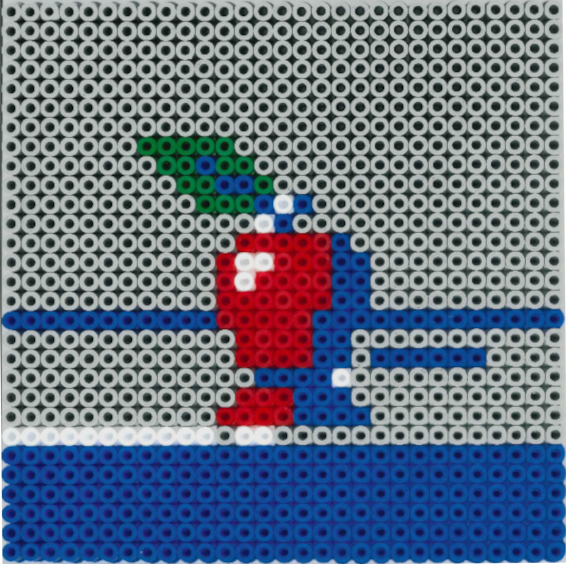} &
    \includegraphics[trim=0 0 0 0, clip, width=0.245\linewidth]{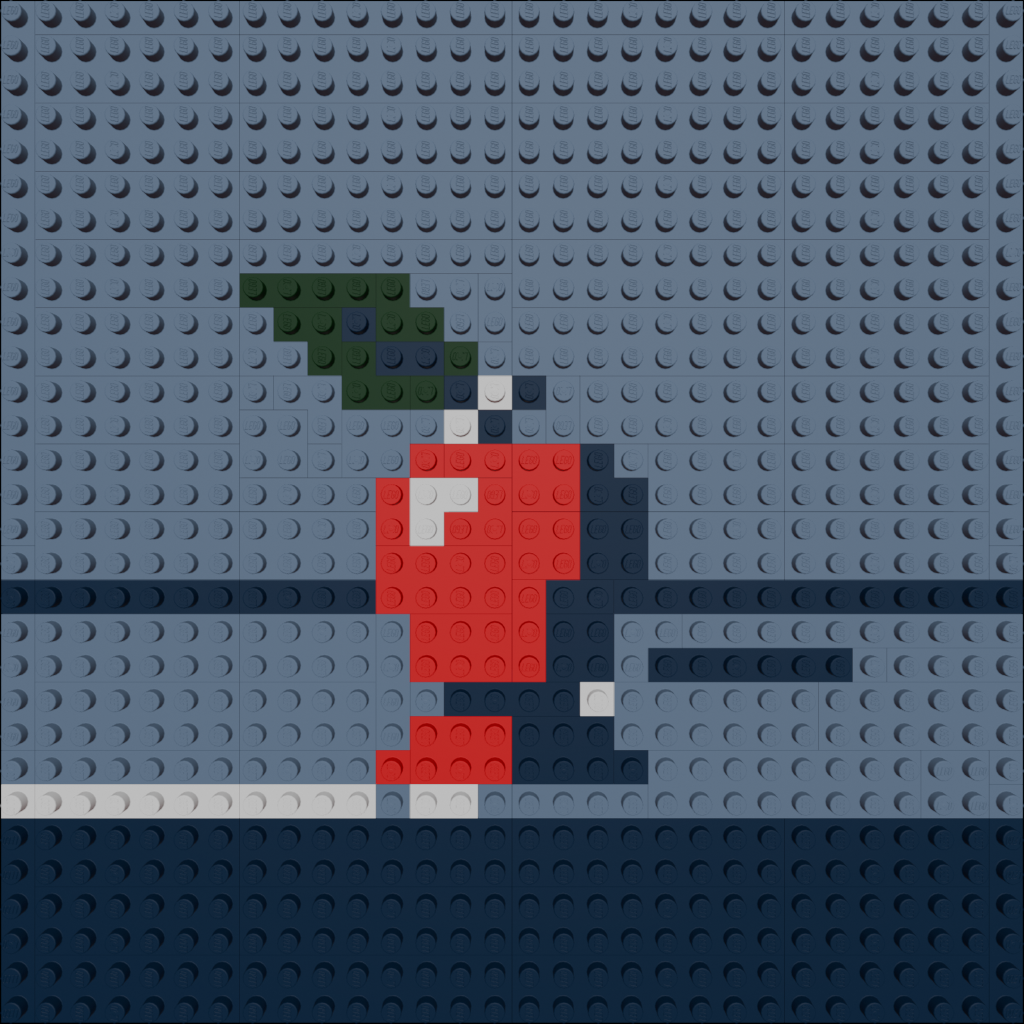} \\
    \ourmethod{} & Embroidery & Fuse beads & Interlocking bricks
	\end{tabular}
	\caption{\ourmethod{} generates low-resolution quantized images that are suitable for many fabrication applications, such as cross-stitch embroidery, fuse beads, or interlocking brick designs. The result image size is $30 \times 30$ pixels, generated without an initialization image, and only conditioned on the prompt ``A red apple with a leaf on a blue table.''}
	\label{fig:fabricationApple}
\end{figure}

\begin{figure}[t]
	\centering
	\small
	\setlength{\tabcolsep}{1pt}
	\begin{tabular}{cccc}%
    \includegraphics[width=0.245\linewidth]{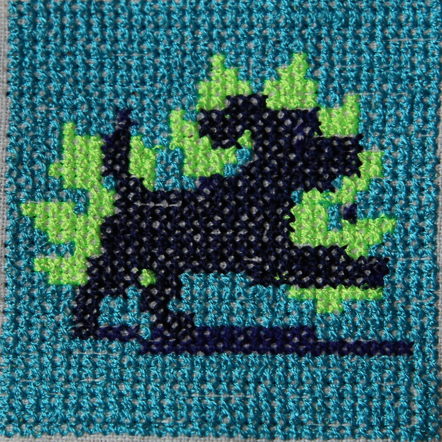} & \includegraphics[width=0.245\linewidth]{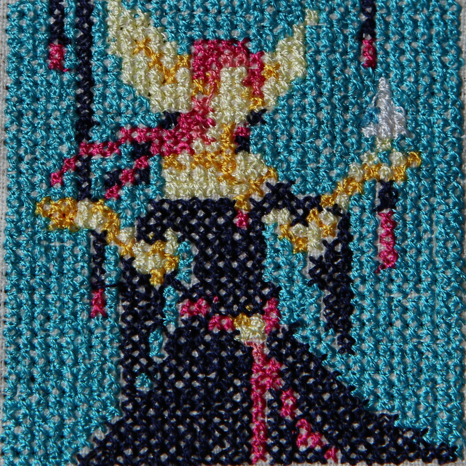} & \includegraphics[width=0.245\linewidth]{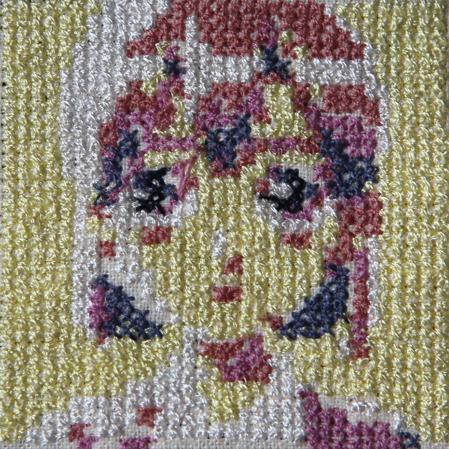} & \includegraphics[width=0.245\linewidth]{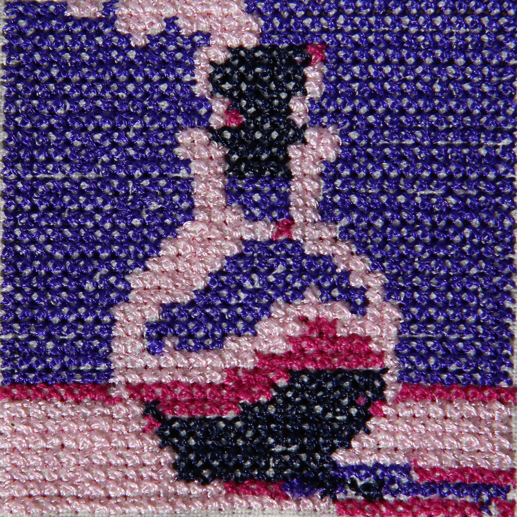} \\
    \includegraphics[width=0.245\linewidth]{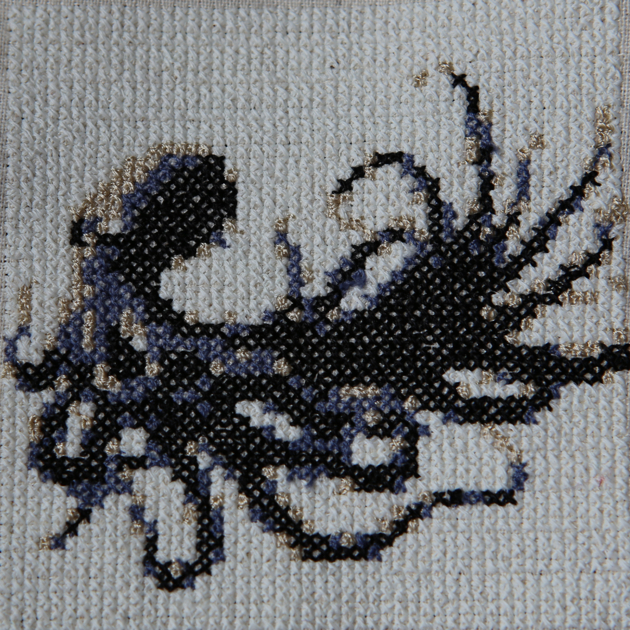} & \includegraphics[width=0.245\linewidth]{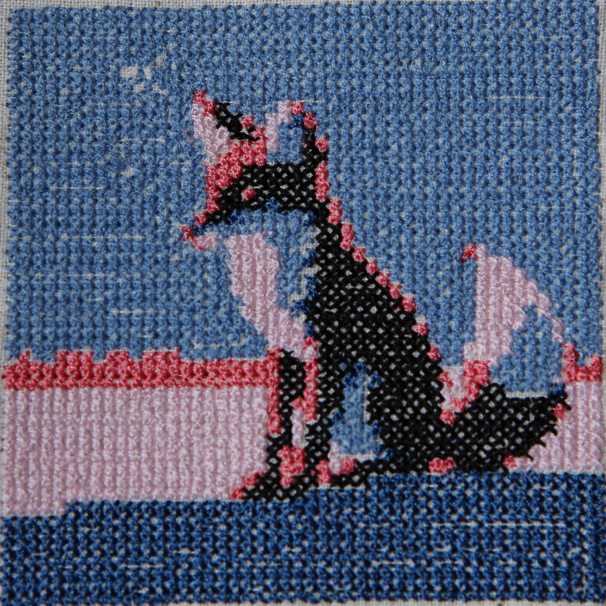} & \includegraphics[width=0.245\linewidth]{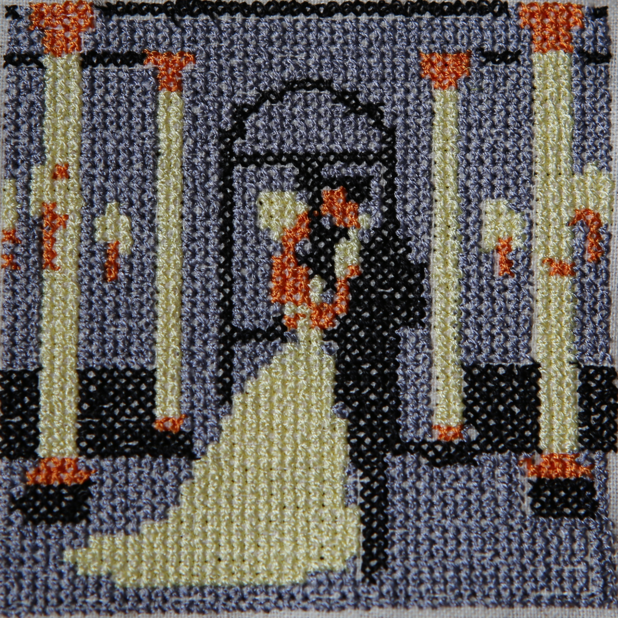} & \includegraphics[width=0.245\linewidth]{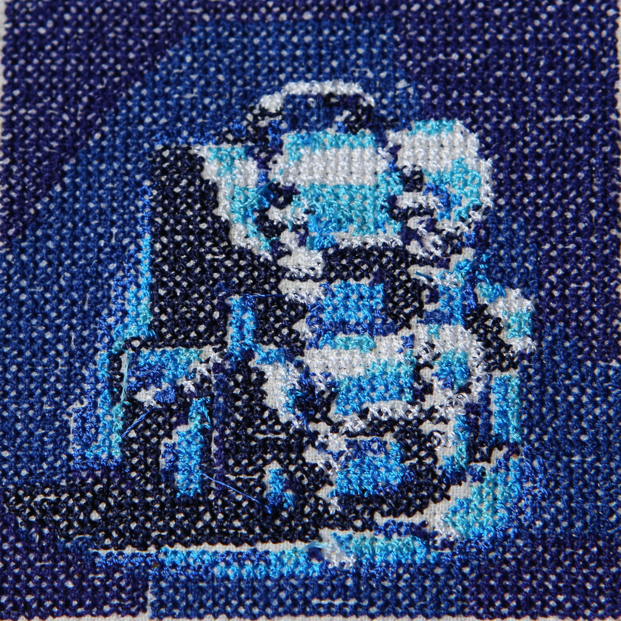} \\
    \includegraphics[width=0.245\linewidth]{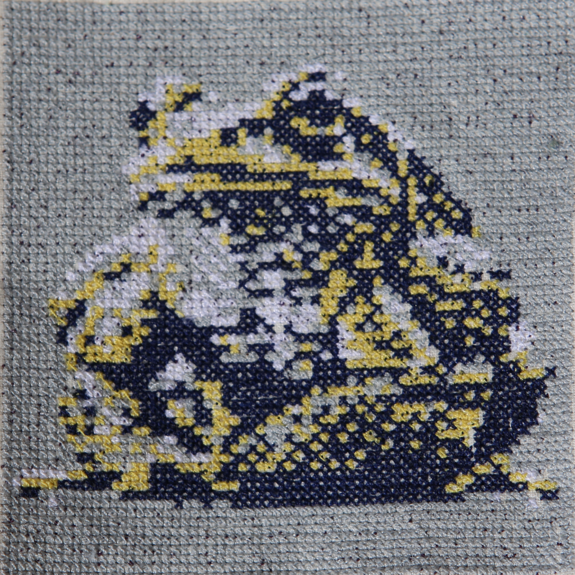} & \includegraphics[width=0.245\linewidth]{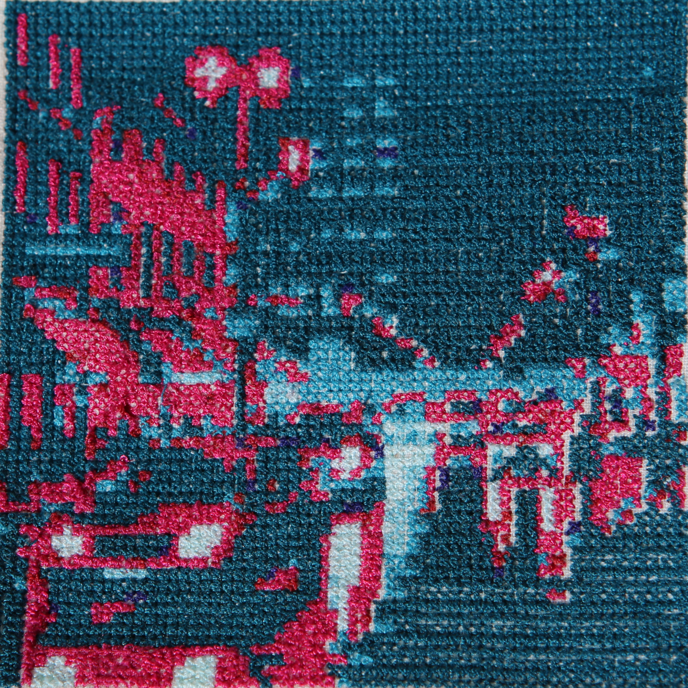} & \includegraphics[width=0.245\linewidth]{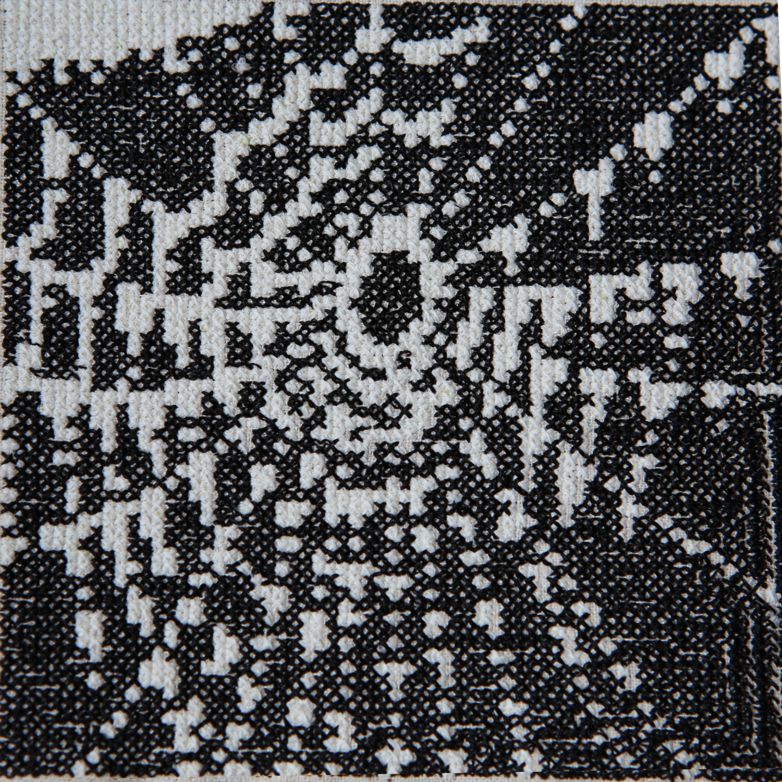} & \includegraphics[width=0.245\linewidth]{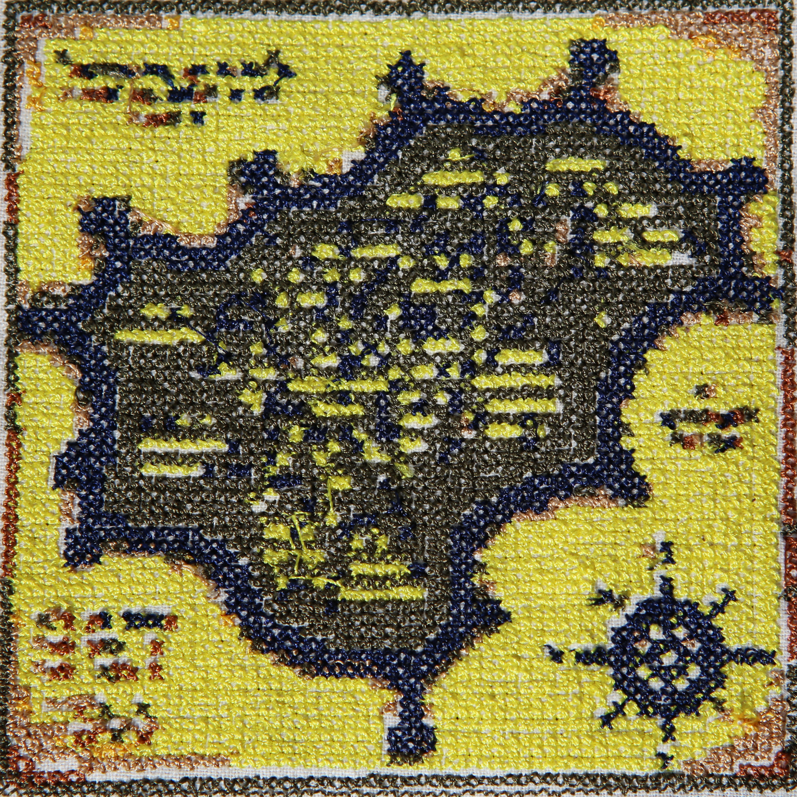} \\
	\end{tabular}
	\caption{Additional examples of cross-stich embroideries showcased on . The sizes of the embroideries are as follow: first row is $32\times32$, second row is $48\times48$, and last row is $64\times64$.}
	\label{fig:fabricationEmbroidery}
\vspace{-0.3cm}
\end{figure}

\section{Additional results}

This section presents more crafted examples and experiments. We show how it is possible to use elements such as small images instead of pixel colors to produce mosaic-like effects.

\subsection{Additional fabrication}

We present additional fabrication examples. \figref{fig:fabricationApple} showcases an example of embroidery, fuse beads and interlocking bricks fabrications, similar to the one presented in the main paper but with lower resolution.
We also present additional cross-stitch embroideries in \figref{fig:fabricationEmbroidery} with various designs, such as different sizes, color palettes and styles. This demonstrates how our method can be used to fabricate diverse yet compelling embroideries.

\subsection{Mosaic}

\begin{figure}[t]
    \center
    \footnotesize
	\setlength{\tabcolsep}{3pt}
    \begin{tabular}{p{0.31\linewidth}p{0.31\linewidth}p{0.31\linewidth}}
    \includegraphics[trim= 0 0 0 0, clip, width=1\linewidth]{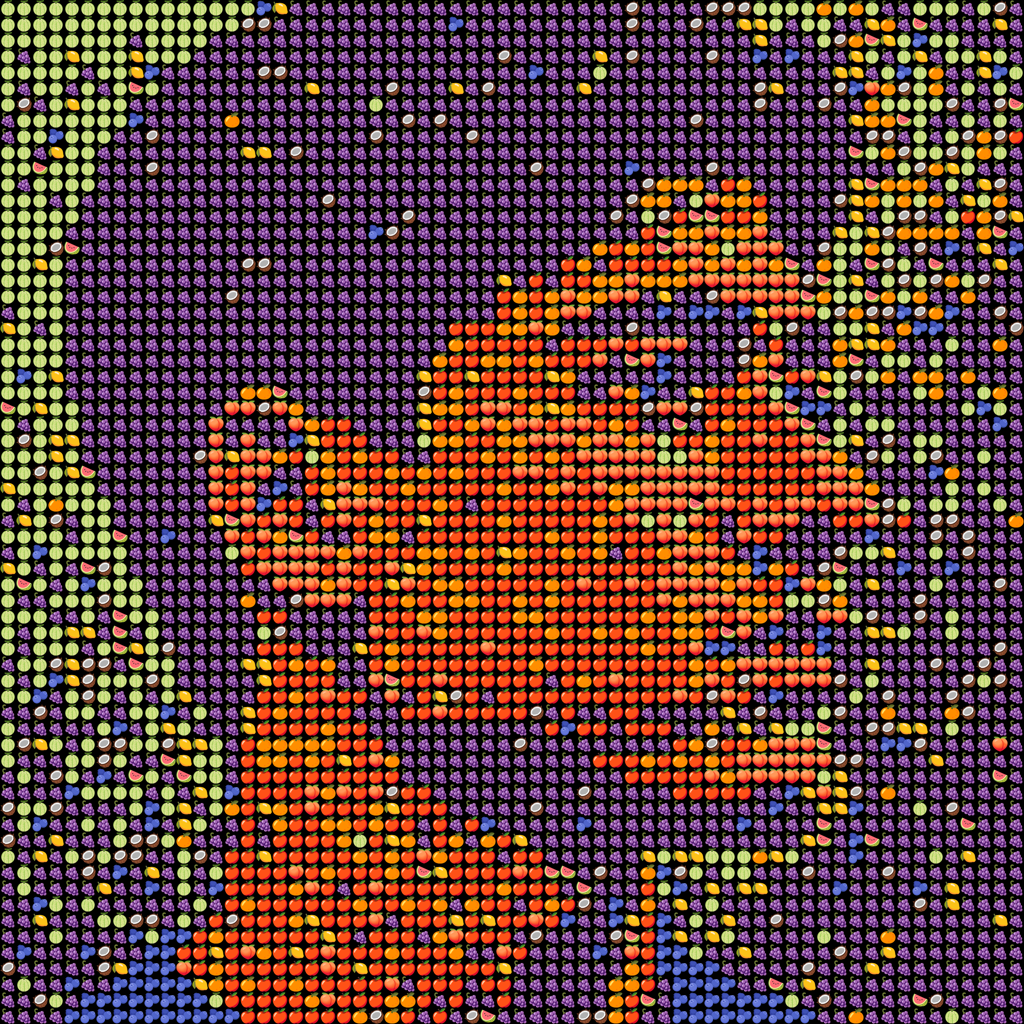} & \includegraphics[trim= 0 0 0 0, clip, width=1\linewidth]{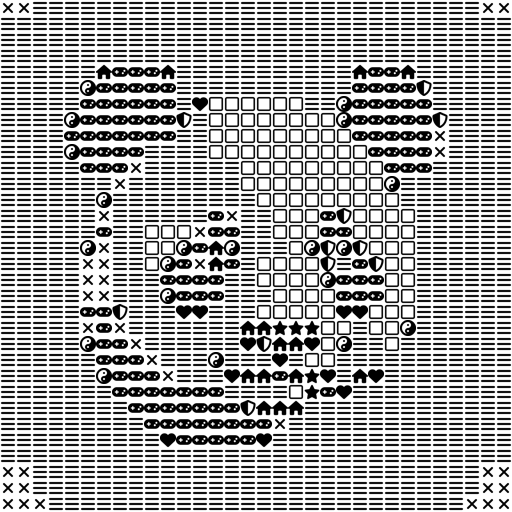} &  \includegraphics[trim= 0 0 0 0, clip, width=1\linewidth]{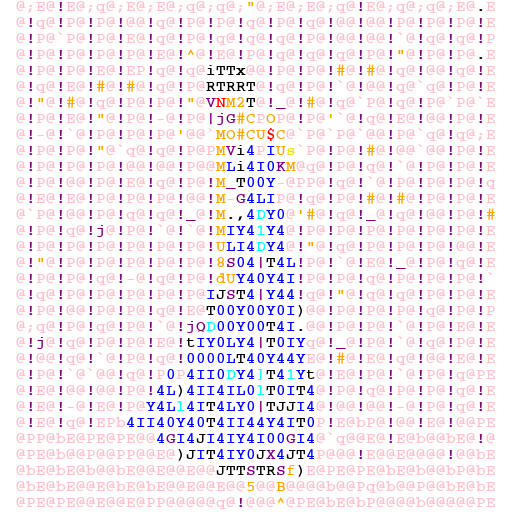} \\
    ``A young's man portrait, looking to the side.'' & ``Face of a panda. Big head and eyes.'' & ``A woman with a long dress.'' \\
    \includegraphics[trim = 0 0 0 0, clip, width=1\linewidth]{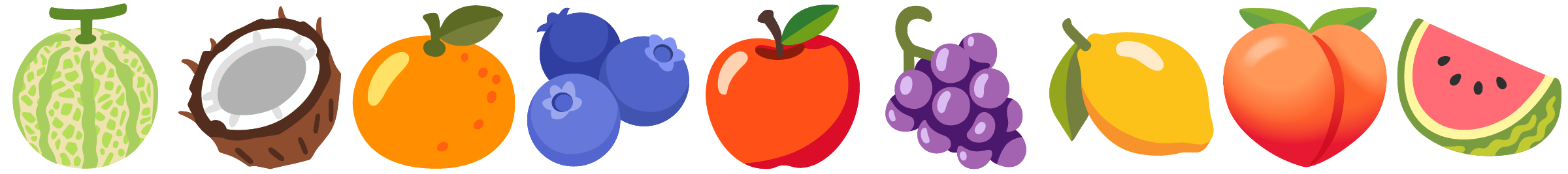} & \includegraphics[trim = 0 0 0 0, clip, width=1\linewidth]{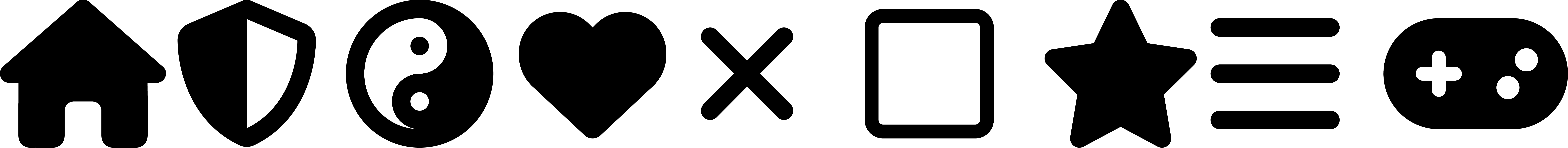} & \includegraphics[trim = 0 0 0 0, clip, width=1\linewidth]{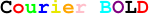}
    \end{tabular} 
    \vspace{-0.3cm}
    \caption{\ourmethod{} demonstrates the ability to create mosaics using diverse elements, such as emojis or icons. Letters can be used to create compelling ASCII art. The results can be better seen on a screen with large zoom.}
    \label{fig:mosaic}
    \vspace{-0.4cm}
\end{figure}

We showcase the generalization capabilities of \ourmethod{} to other unit elements than pixel colors. As outlined in Section 4.1 of the main paper, instead of colors, our generator is capable of using any elements that can be rendered into images of equal dimensions $h \times w$, enabling the creation of images with a mosaic-like effect. Examples of this capability are demonstrated in \figref{fig:mosaic}, where we employ a variety of elements including colorful fruit emojis \cite{emojiterra}, black and white open icons from Font Awesome \cite{fontawesome} and letters in random colors from the monospace font Courier Bold \cite{fontsgeek}.

%% file: sections/supplementary/controlnetWeightsFigure.tex
\begin{figure*}[ht]
\centering
\footnotesize
\setlength{\tabcolsep}{1pt}
\begin{tabular}{ccccccccccccc}
\toprule
\multicolumn{2}{r}{\multirow{2}{*}{\scriptsize Depth}} & \multirow{2}{*}{0.0} & \multirow{2}{*}{0.1} & \multirow{2}{*}{0.2} & \multirow{2}{*}{0.3} & \multirow{2}{*}{0.4} & \multirow{2}{*}{0.5} & \multirow{2}{*}{0.6} & \multirow{2}{*}{0.7} & \multirow{2}{*}{0.8} & \multirow{2}{*}{0.9} & \multirow{2}{*}{1.0} \\
\addlinespace[-0.1cm]
\raisebox{-7pt}[0pt][0pt]{{\scriptsize Canny}} & & & & & & & & & & \\
\raisebox{19pt}[0pt][0pt]{0.0} & & \includegraphics[width=0.084\linewidth]{figures/supplementary/ablation/controlnet/c_0.0_d_0.0.png} & \includegraphics[width=0.084\linewidth]{figures/supplementary/ablation/controlnet/c_0.0_d_0.1.png} & \includegraphics[width=0.084\linewidth]{figures/supplementary/ablation/controlnet/c_0.0_d_0.2.png} & \includegraphics[width=0.084\linewidth]{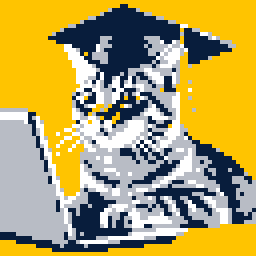} & \includegraphics[width=0.084\linewidth]{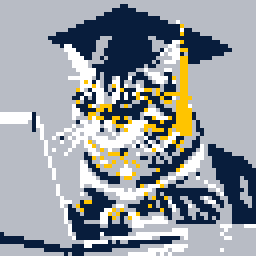} & \includegraphics[width=0.084\linewidth]{figures/supplementary/ablation/controlnet/c_0.0_d_0.5.png} & \includegraphics[width=0.084\linewidth]{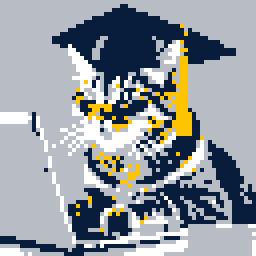} & \includegraphics[width=0.084\linewidth]{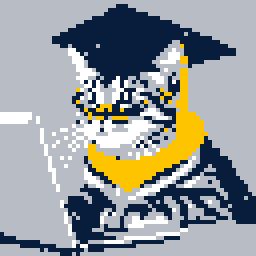} & \includegraphics[width=0.084\linewidth]{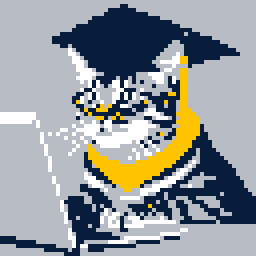} & \includegraphics[width=0.084\linewidth]{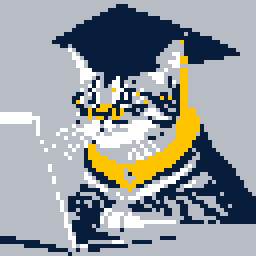} & \includegraphics[width=0.084\linewidth]{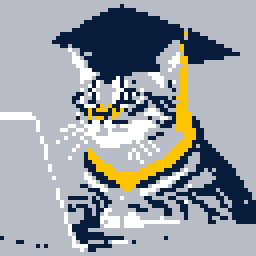} \\

\raisebox{19pt}[0pt][0pt]{0.1} && \includegraphics[width=0.084\linewidth]{figures/supplementary/ablation/controlnet/c_0.1_d_0.0.png} & \includegraphics[width=0.084\linewidth]{figures/supplementary/ablation/controlnet/c_0.1_d_0.1.png} & \includegraphics[width=0.084\linewidth]{figures/supplementary/ablation/controlnet/c_0.1_d_0.2.png} & \includegraphics[width=0.084\linewidth]{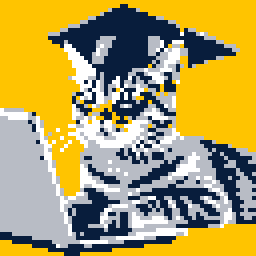} & \includegraphics[width=0.084\linewidth]{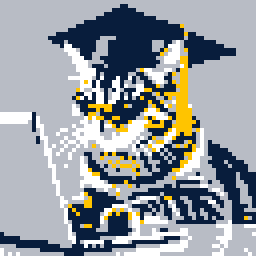} & \includegraphics[width=0.084\linewidth]{figures/supplementary/ablation/controlnet/c_0.1_d_0.5.png} & \includegraphics[width=0.084\linewidth]{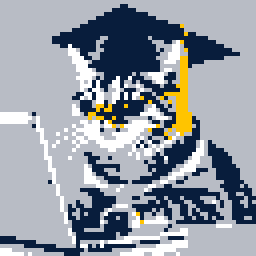} & \includegraphics[width=0.084\linewidth]{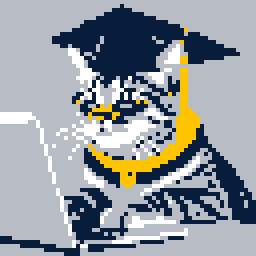} & \includegraphics[width=0.084\linewidth]{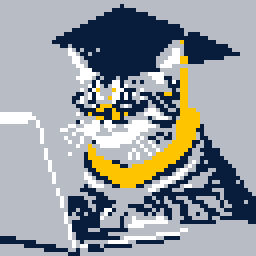} & \includegraphics[width=0.084\linewidth]{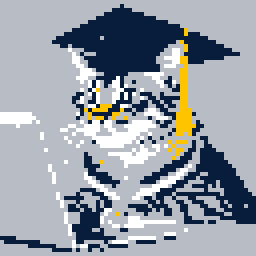} & \includegraphics[width=0.084\linewidth]{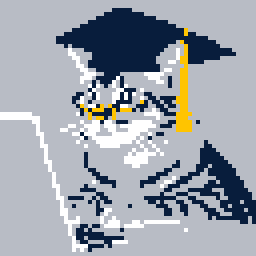} \\

\raisebox{19pt}[0pt][0pt]{0.2} &&  \includegraphics[width=0.084\linewidth]{figures/supplementary/ablation/controlnet/c_0.2_d_0.0.png} & \includegraphics[width=0.084\linewidth]{figures/supplementary/ablation/controlnet/c_0.2_d_0.1.png} & \includegraphics[width=0.084\linewidth]{figures/supplementary/ablation/controlnet/c_0.2_d_0.2.png} & \includegraphics[width=0.084\linewidth]{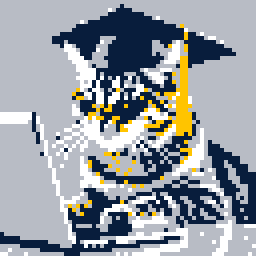} & \includegraphics[width=0.084\linewidth]{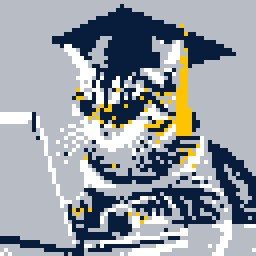} & \includegraphics[width=0.084\linewidth]{figures/supplementary/ablation/controlnet/c_0.2_d_0.5.png} & \includegraphics[width=0.084\linewidth]{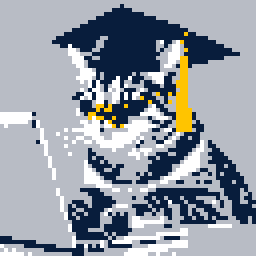} & \includegraphics[width=0.084\linewidth]{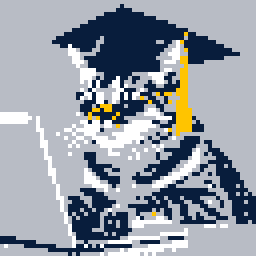} & \includegraphics[width=0.084\linewidth]{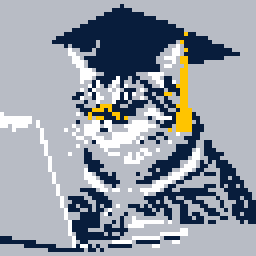} & \includegraphics[width=0.084\linewidth]{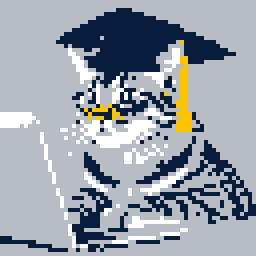} & \includegraphics[width=0.084\linewidth]{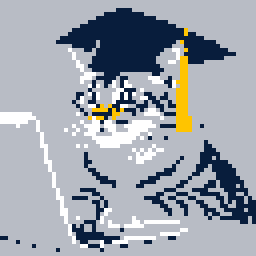} \\

\raisebox{19pt}[0pt][0pt]{0.3} & & \includegraphics[width=0.084\linewidth]{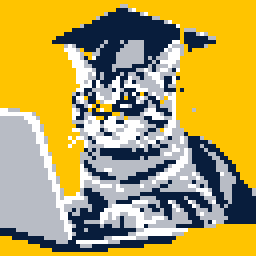} & \includegraphics[width=0.084\linewidth]{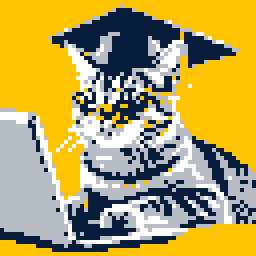} & \includegraphics[width=0.084\linewidth]{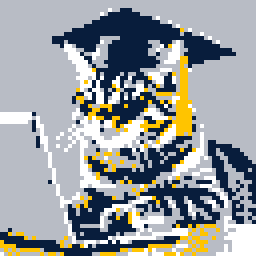} & \includegraphics[width=0.084\linewidth]{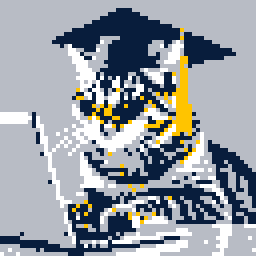} & \includegraphics[width=0.084\linewidth]{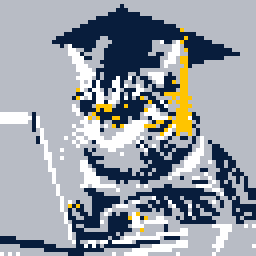} & \includegraphics[width=0.084\linewidth]{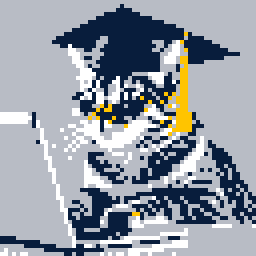} & \includegraphics[width=0.084\linewidth]{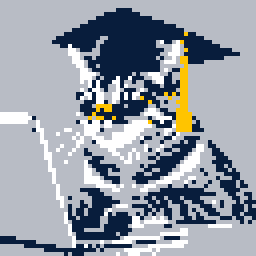} & \includegraphics[width=0.084\linewidth]{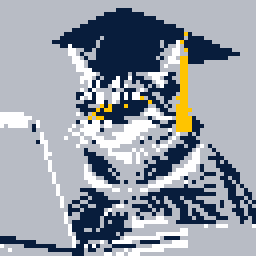} & \includegraphics[width=0.084\linewidth]{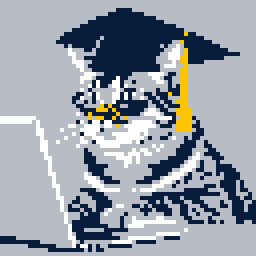} & \includegraphics[width=0.084\linewidth]{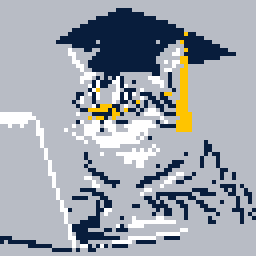} & \includegraphics[width=0.084\linewidth]{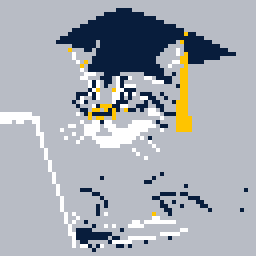} \\

\raisebox{19pt}[0pt][0pt]{0.4} & & \includegraphics[width=0.084\linewidth]{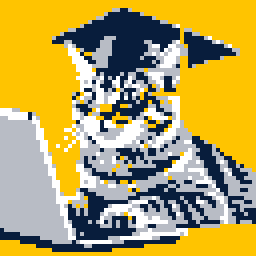} & \includegraphics[width=0.084\linewidth]{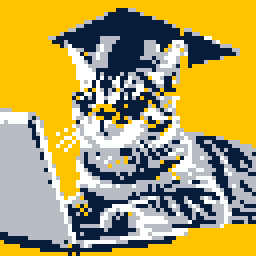} & \includegraphics[width=0.084\linewidth]{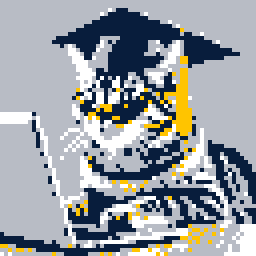} & \includegraphics[width=0.084\linewidth]{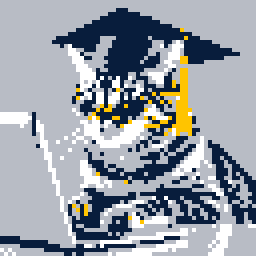} & \includegraphics[width=0.084\linewidth]{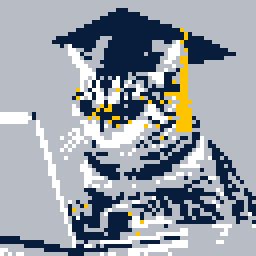} & \includegraphics[width=0.084\linewidth]{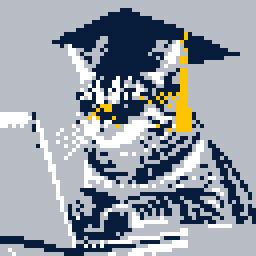} & \includegraphics[width=0.084\linewidth]{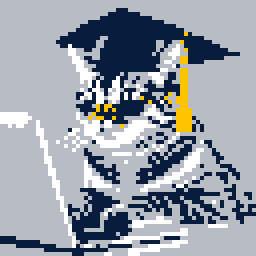} & \includegraphics[width=0.084\linewidth]{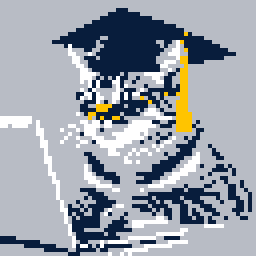} & \includegraphics[width=0.084\linewidth]{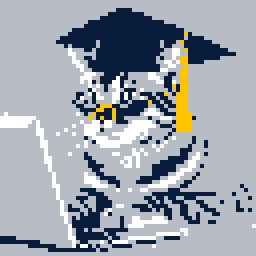} & \includegraphics[width=0.084\linewidth]{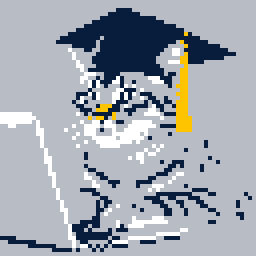} & \includegraphics[width=0.084\linewidth]{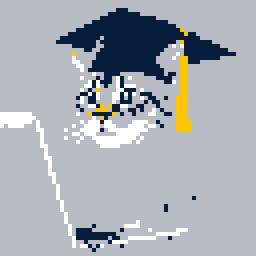} \\

\raisebox{19pt}[0pt][0pt]{0.5} & & \includegraphics[width=0.084\linewidth]{figures/supplementary/ablation/controlnet/c_0.5_d_0.0.png} & \includegraphics[width=0.084\linewidth]{figures/supplementary/ablation/controlnet/c_0.5_d_0.1.png} & \includegraphics[width=0.084\linewidth]{figures/supplementary/ablation/controlnet/c_0.5_d_0.2.png} & \includegraphics[width=0.084\linewidth]{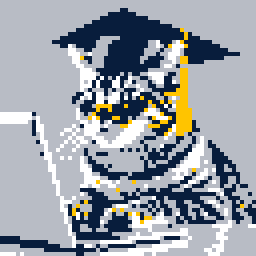} & \includegraphics[width=0.084\linewidth]{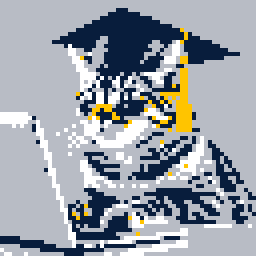} & \includegraphics[width=0.084\linewidth]{figures/supplementary/ablation/controlnet/c_0.5_d_0.5.png} & \includegraphics[width=0.084\linewidth]{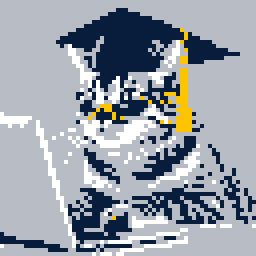} & \includegraphics[width=0.084\linewidth]{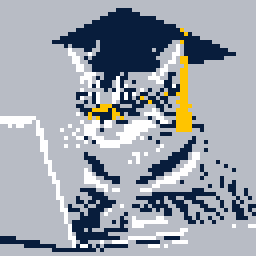} & \includegraphics[width=0.084\linewidth]{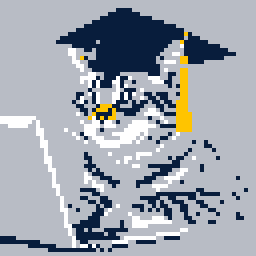} & \includegraphics[width=0.084\linewidth]{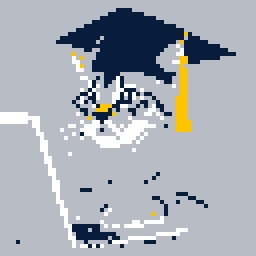} & \includegraphics[width=0.084\linewidth]{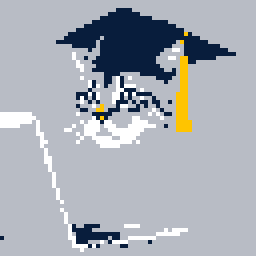} \\

\raisebox{19pt}[0pt][0pt]{0.6} & & \includegraphics[width=0.084\linewidth]{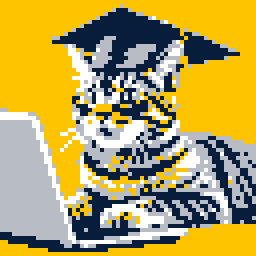} & \includegraphics[width=0.084\linewidth]{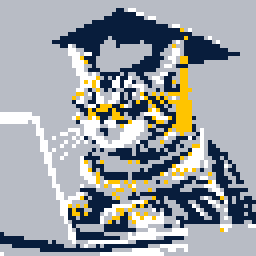} & \includegraphics[width=0.084\linewidth]{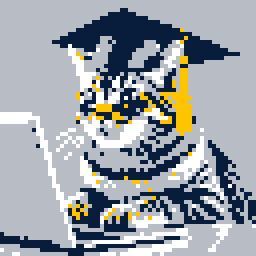} & \includegraphics[width=0.084\linewidth]{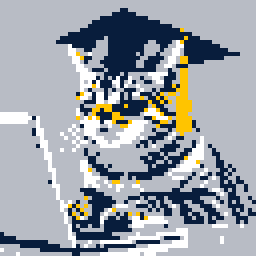} & \includegraphics[width=0.084\linewidth]{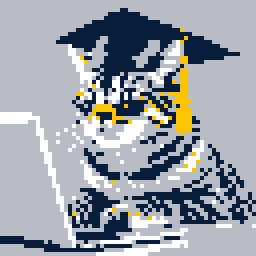} & \includegraphics[width=0.084\linewidth]{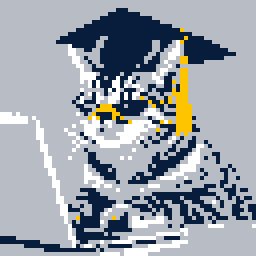} & \includegraphics[width=0.084\linewidth]{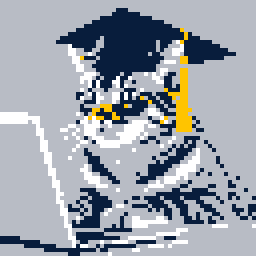} & \includegraphics[width=0.084\linewidth]{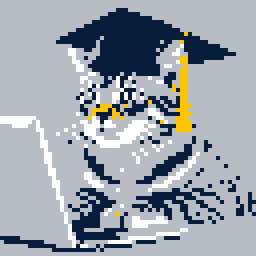} & \includegraphics[width=0.084\linewidth]{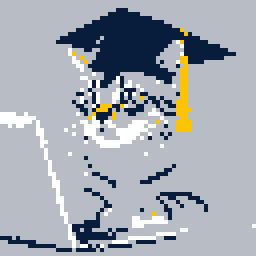} & \includegraphics[width=0.084\linewidth]{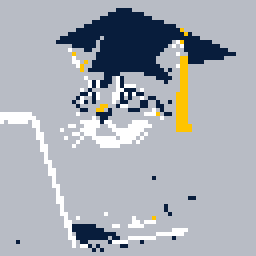} & \includegraphics[width=0.084\linewidth]{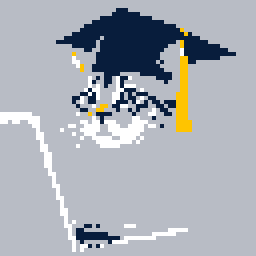} \\

\raisebox{19pt}[0pt][0pt]{0.7} & & \includegraphics[width=0.084\linewidth]{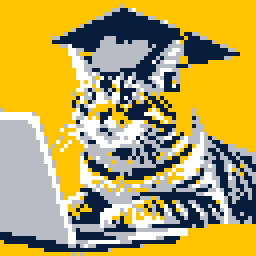} & \includegraphics[width=0.084\linewidth]{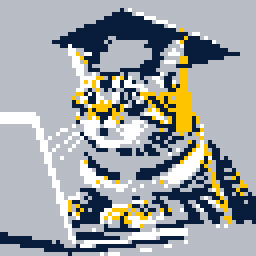} & \includegraphics[width=0.084\linewidth]{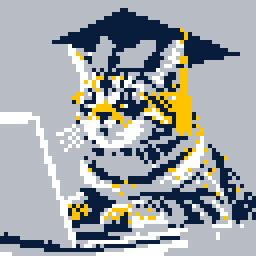} & \includegraphics[width=0.084\linewidth]{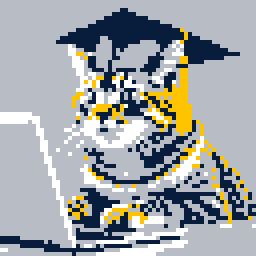} & \includegraphics[width=0.084\linewidth]{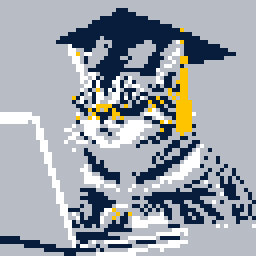} & \includegraphics[width=0.084\linewidth]{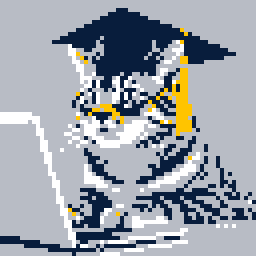} & \includegraphics[width=0.084\linewidth]{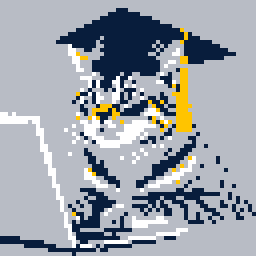} & \includegraphics[width=0.084\linewidth]{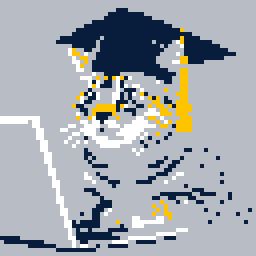} & \includegraphics[width=0.084\linewidth]{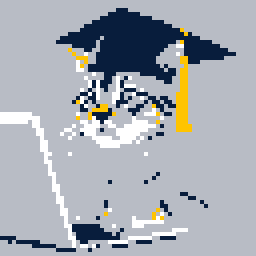} & \includegraphics[width=0.084\linewidth]{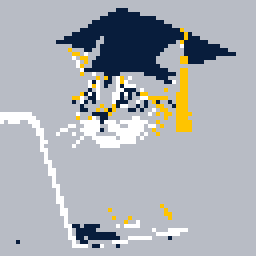} & \includegraphics[width=0.084\linewidth]{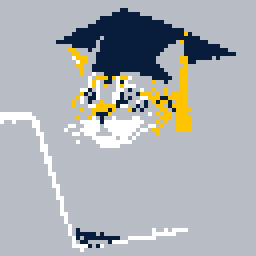} \\

\raisebox{19pt}[0pt][0pt]{0.8} & & \includegraphics[width=0.084\linewidth]{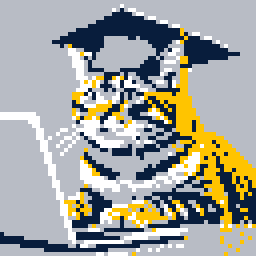} & \includegraphics[width=0.084\linewidth]{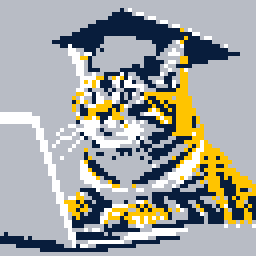} & \includegraphics[width=0.084\linewidth]{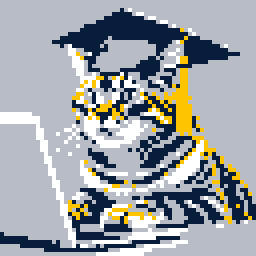} & \includegraphics[width=0.084\linewidth]{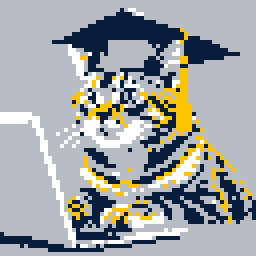} & \includegraphics[width=0.084\linewidth]{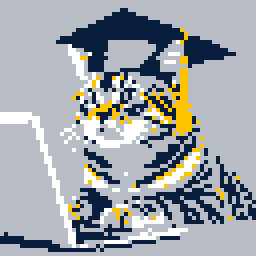} & \includegraphics[width=0.084\linewidth]{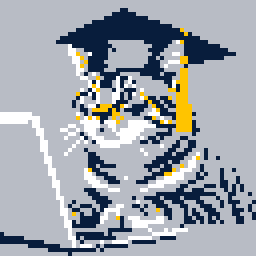} & \includegraphics[width=0.084\linewidth]{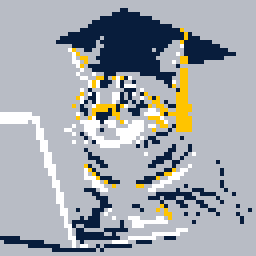} & \includegraphics[width=0.084\linewidth]{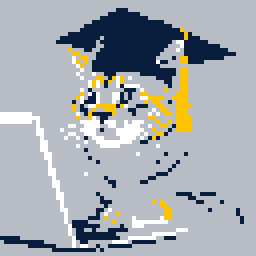} & \includegraphics[width=0.084\linewidth]{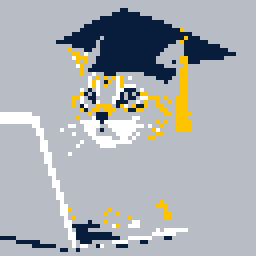} & \includegraphics[width=0.084\linewidth]{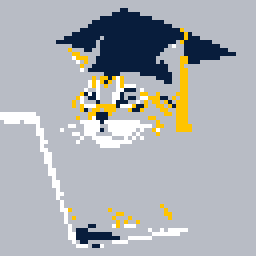} & \includegraphics[width=0.084\linewidth]{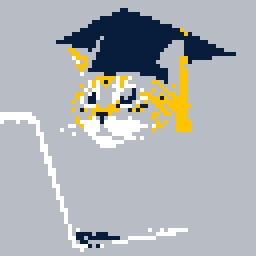} \\

\raisebox{19pt}[0pt][0pt]{0.9} & & \includegraphics[width=0.084\linewidth]{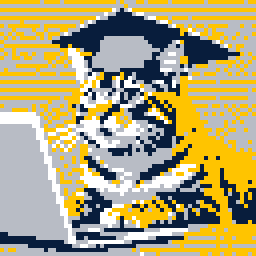} & \includegraphics[width=0.084\linewidth]{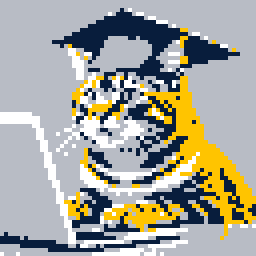} & \includegraphics[width=0.084\linewidth]{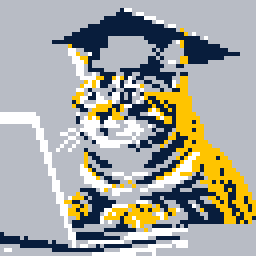} & \includegraphics[width=0.084\linewidth]{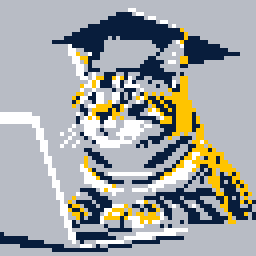} & \includegraphics[width=0.084\linewidth]{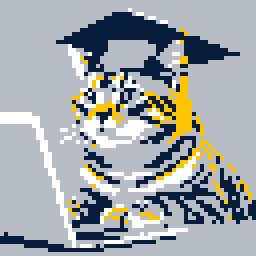} & \includegraphics[width=0.084\linewidth]{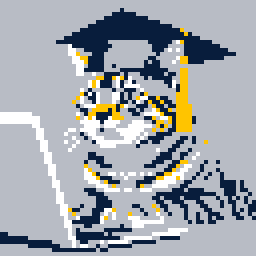} & \includegraphics[width=0.084\linewidth]{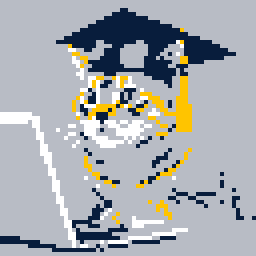} & \includegraphics[width=0.084\linewidth]{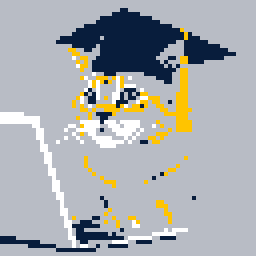} & \includegraphics[width=0.084\linewidth]{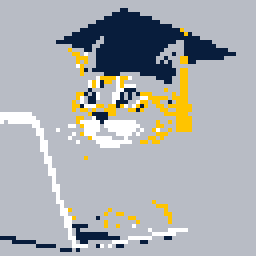} & \includegraphics[width=0.084\linewidth]{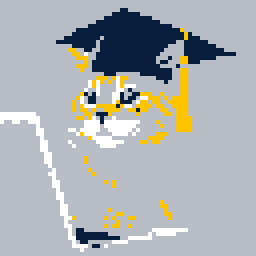} & \includegraphics[width=0.084\linewidth]{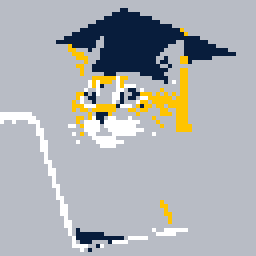} \\

\raisebox{19pt}[0pt][0pt]{1.0} & & \includegraphics[width=0.084\linewidth]{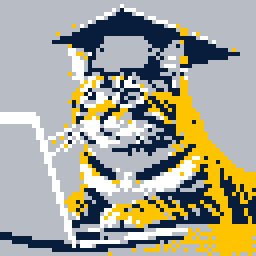} & \includegraphics[width=0.084\linewidth]{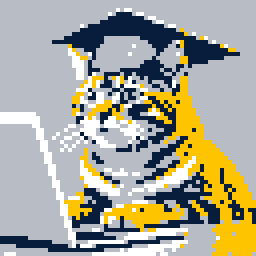} & \includegraphics[width=0.084\linewidth]{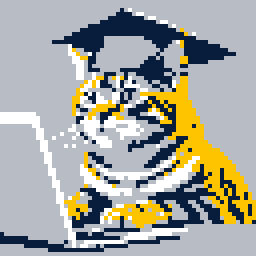} & \includegraphics[width=0.084\linewidth]{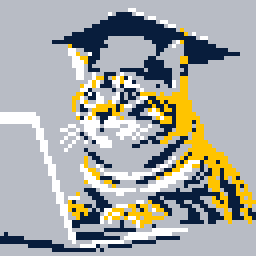} & \includegraphics[width=0.084\linewidth]{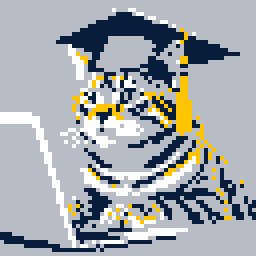} & \includegraphics[width=0.084\linewidth]{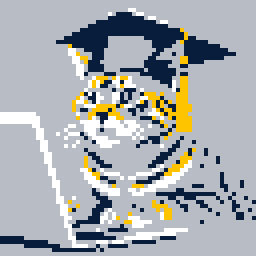} & \includegraphics[width=0.084\linewidth]{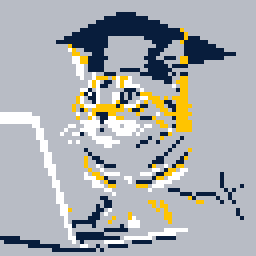} & \includegraphics[width=0.084\linewidth]{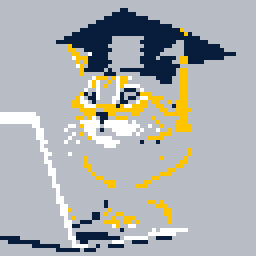} & \includegraphics[width=0.084\linewidth]{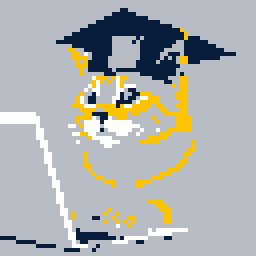} & \includegraphics[width=0.084\linewidth]{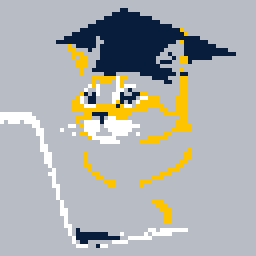} & \includegraphics[width=0.084\linewidth]{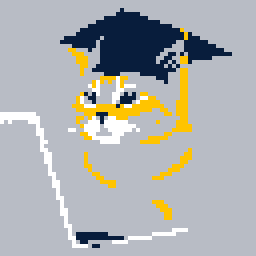} \\
\bottomrule

\end{tabular}
\caption{We present the combined effects of ControlNet \cite{zhang2023addingControlNet} weights on both Canny edge and depth-conditioning networks \cite{vonPlaten2022diffusers}, examined concurrently. We initialize the weights of the generator randomly to disambiguate the contribution of ControlNet to the spatial fidelity of the generation from the influence of the initialization. The cumulative nature of these weights accounts for the distortion observed in the output when their sum exceeds $1$.}
\label{fig:ControlNetWeights}
\end{figure*}

%% file: sections/supplementary/bigfigure_quantitative.tex
\begin{figure*}[t]

\renewcommand{\arraystretch}{1.55}
	\centering
	\small
	\setlength{\tabcolsep}{0pt}
	\begin{tabular}{*{32}c}
 \toprule
 \multicolumn{32}{c}{Palette} \\
 \multicolumn{2}{c}{\multirow{2}{*}{\includegraphics[width=0.06\linewidth]{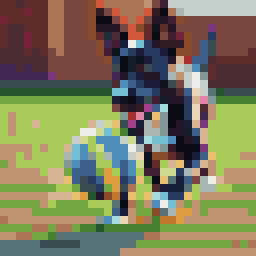}}} & 
\multicolumn{2}{c}{\multirow{2}{*}{\includegraphics[width=0.06\linewidth]{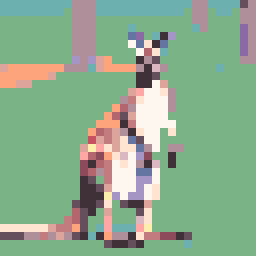} }} & 
\multicolumn{2}{c}{\multirow{2}{*}{\includegraphics[width=0.06\linewidth]{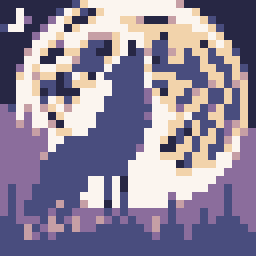} }} & 
\multicolumn{2}{c}{\multirow{2}{*}{\includegraphics[width=0.06\linewidth]{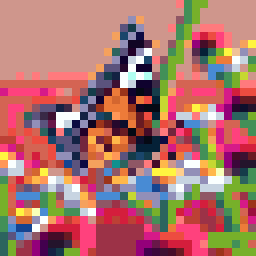} }} &

\multicolumn{3}{c}{\multirow{3}{*}{\includegraphics[width=0.09\linewidth]{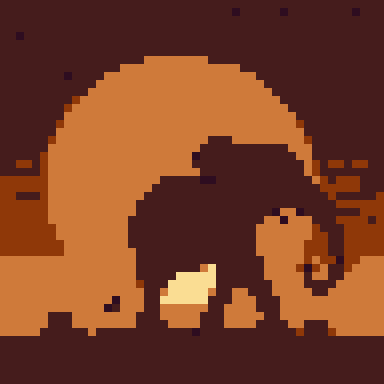}}} & 
\multicolumn{3}{c}{\multirow{3}{*}{\includegraphics[width=0.09\linewidth]{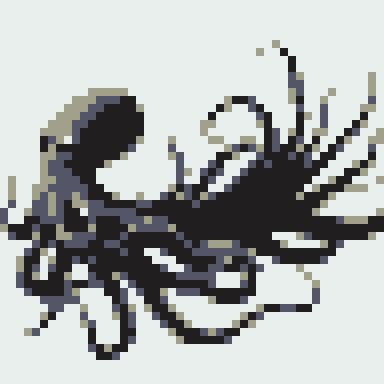} }} & 
\multicolumn{3}{c}{\multirow{3}{*}{\includegraphics[width=0.09\linewidth]{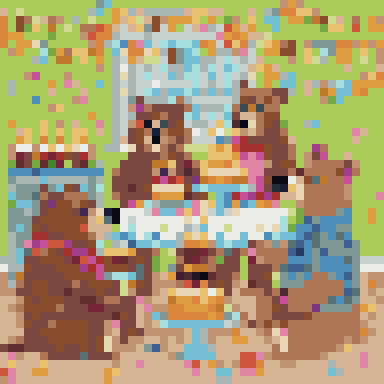} }} & 
\multicolumn{3}{c}{\multirow{3}{*}{\includegraphics[width=0.09\linewidth]{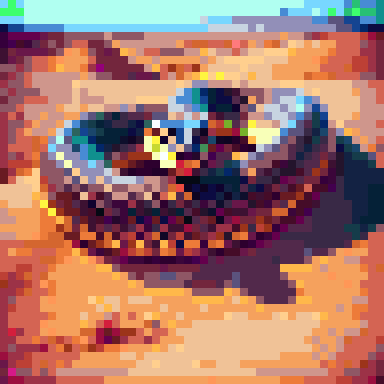} }} & 
\multicolumn{3}{c}{\multirow{3}{*}{\includegraphics[width=0.09\linewidth]{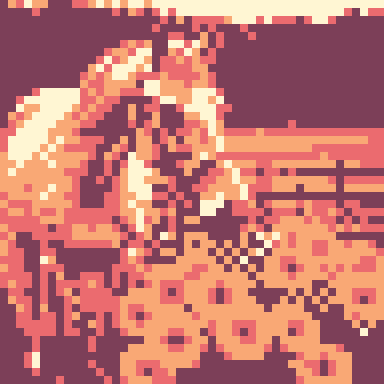} }} & 
\multicolumn{3}{c}{\multirow{3}{*}{\includegraphics[width=0.09\linewidth]{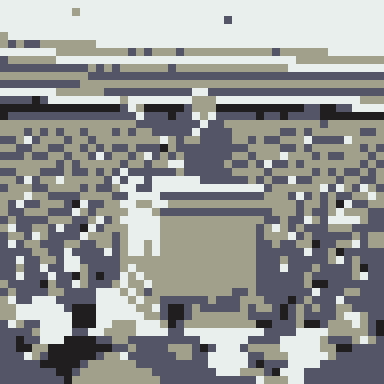} }} & 
\multicolumn{3}{c}{\multirow{3}{*}{\includegraphics[width=0.09\linewidth]{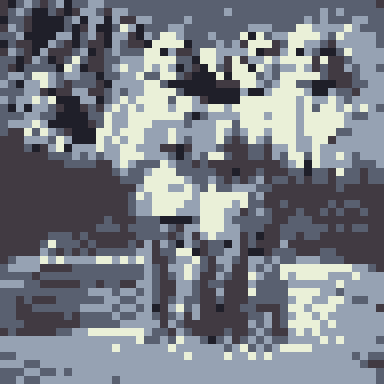} }} & 
\multicolumn{3}{c}{\multirow{3}{*}{\includegraphics[width=0.09\linewidth]{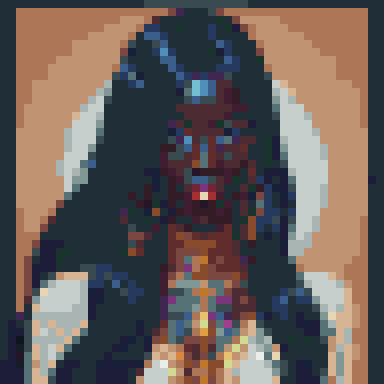} }} \\

\\


\multicolumn{2}{c}{\multirow{2}{*}{\includegraphics[width=0.06\linewidth]{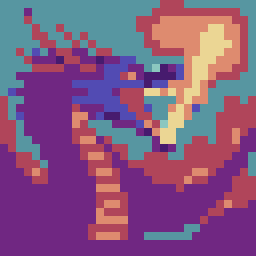} }} & 
\multicolumn{2}{c}{\multirow{2}{*}{\includegraphics[width=0.06\linewidth]{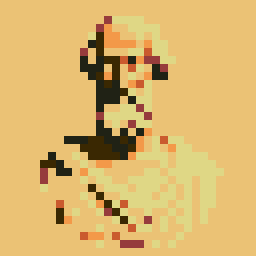} }} & 
\multicolumn{2}{c}{\multirow{2}{*}{\includegraphics[width=0.06\linewidth]{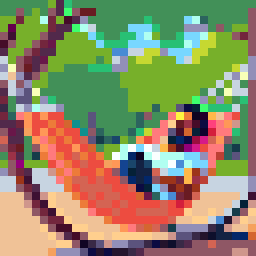} }} & 
\multicolumn{2}{c}{\multirow{2}{*}{\includegraphics[width=0.06\linewidth]{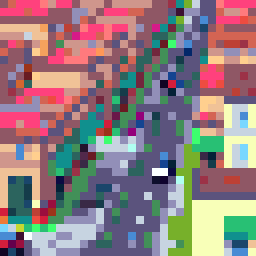} }} & 

 \multicolumn{24}{c}{}

 \\
 

 \multicolumn{8}{c}{} & 
\multicolumn{3}{c}{\multirow{3}{*}{\includegraphics[width=0.09\linewidth]{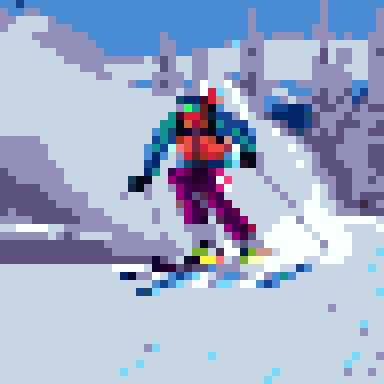} }} & 
\multicolumn{3}{c}{\multirow{3}{*}{\includegraphics[width=0.09\linewidth]{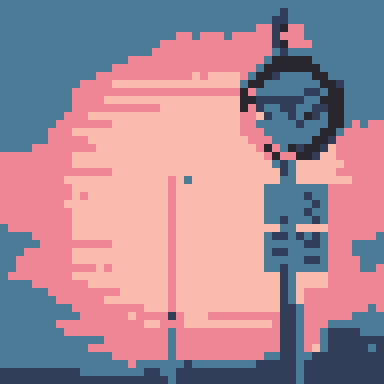} }} & 
\multicolumn{3}{c}{\multirow{3}{*}{\includegraphics[width=0.09\linewidth]{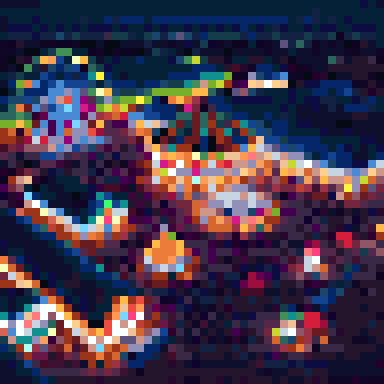} }} & 
\multicolumn{3}{c}{\multirow{3}{*}{\includegraphics[width=0.09\linewidth]{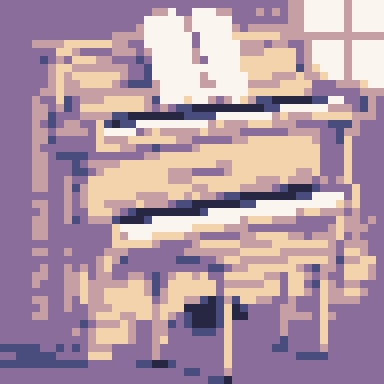} }} & 
\multicolumn{3}{c}{\multirow{3}{*}{\includegraphics[width=0.09\linewidth]{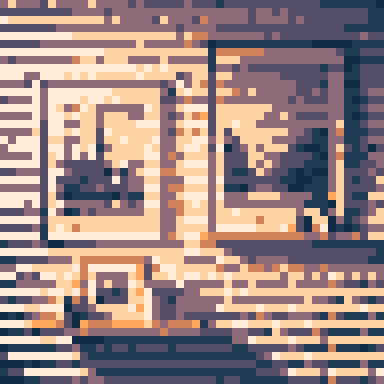} }} & 
\multicolumn{3}{c}{\multirow{3}{*}{\includegraphics[width=0.09\linewidth]{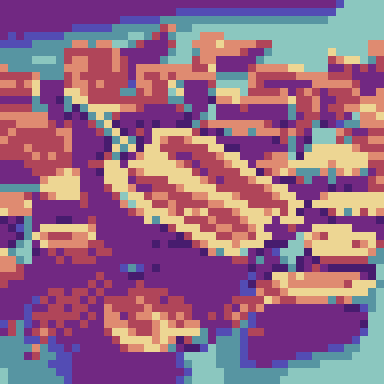} }} & 
\multicolumn{3}{c}{\multirow{3}{*}{\includegraphics[width=0.09\linewidth]{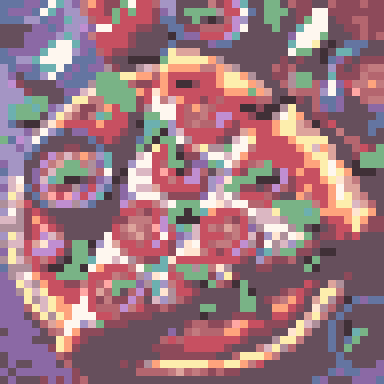} }} & 
\multicolumn{3}{c}{\multirow{3}{*}{\includegraphics[width=0.09\linewidth]{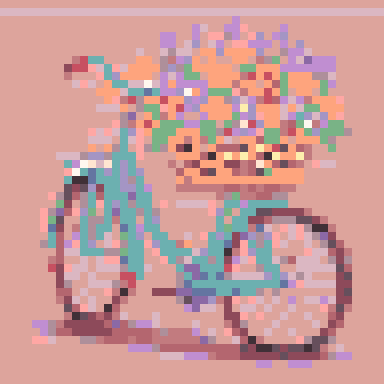} }}
\\
\multicolumn{2}{c}{\multirow{2}{*}{\includegraphics[width=0.06\linewidth]{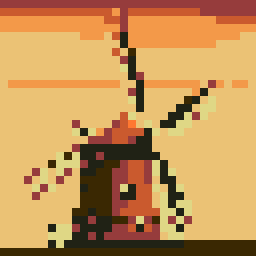} }} & 
\multicolumn{2}{c}{\multirow{2}{*}{\includegraphics[width=0.06\linewidth]{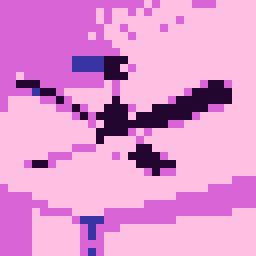} }} & 
\multicolumn{2}{c}{\multirow{2}{*}{\includegraphics[width=0.06\linewidth]{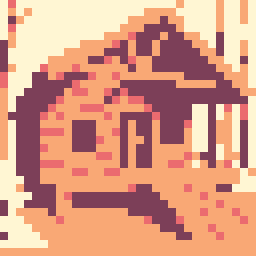} }} & 
\multicolumn{2}{c}{\multirow{2}{*}{\includegraphics[width=0.06\linewidth]{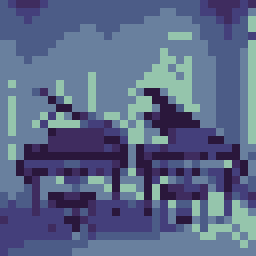} }} & 
 \multicolumn{24}{c}{}
 \\

 
 \\
 
\multicolumn{2}{c}{\multirow{2}{*}{\includegraphics[width=0.06\linewidth]{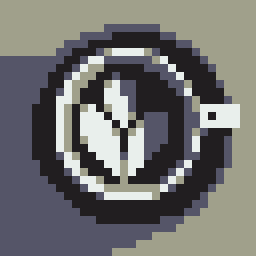} }} & 
\multicolumn{2}{c}{\multirow{2}{*}{\includegraphics[width=0.06\linewidth]{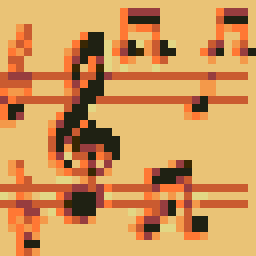} }} & 
\multicolumn{2}{c}{\multirow{2}{*}{\includegraphics[width=0.06\linewidth]{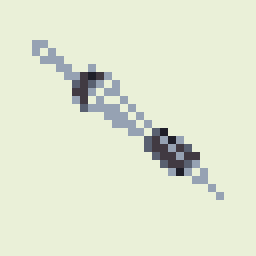} }} & 
\multicolumn{2}{c}{\multirow{2}{*}{\includegraphics[width=0.06\linewidth]{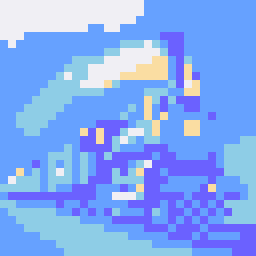} }} &

\multicolumn{4}{c}{\multirow{4}{*}{\includegraphics[width=0.12\linewidth]{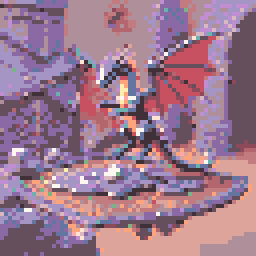}}} &
\multicolumn{4}{c}{\multirow{4}{*}{\includegraphics[width=0.12\linewidth]{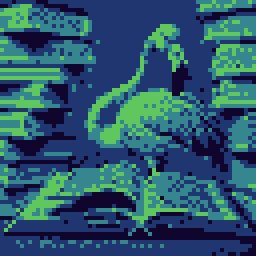} }} &
\multicolumn{4}{c}{\multirow{4}{*}{\includegraphics[width=0.12\linewidth]{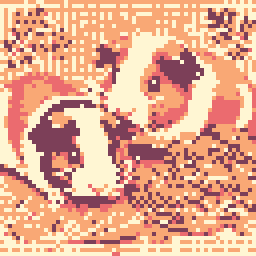} }} &
\multicolumn{4}{c}{\multirow{4}{*}{\includegraphics[width=0.12\linewidth]{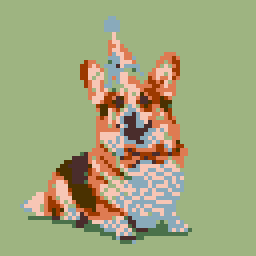} }} &
\multicolumn{4}{c}{\multirow{4}{*}{\includegraphics[width=0.12\linewidth]{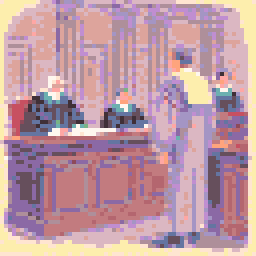} }} &
\multicolumn{4}{c}{\multirow{4}{*}{\includegraphics[width=0.12\linewidth]{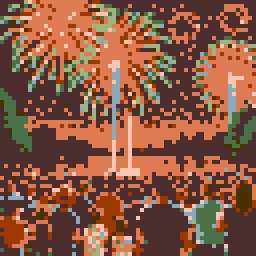} }}  \\
 \\
\multicolumn{2}{c}{\multirow{2}{*}{\includegraphics[width=0.06\linewidth]{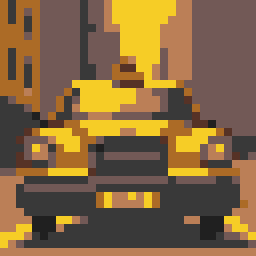} }} & 
\multicolumn{2}{c}{\multirow{2}{*}{\includegraphics[width=0.06\linewidth]{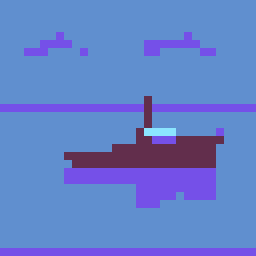} }} & 
\multicolumn{2}{c}{\multirow{2}{*}{\includegraphics[width=0.06\linewidth]{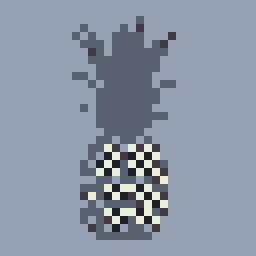} }} & 
\multicolumn{2}{c}{\multirow{2}{*}{\includegraphics[width=0.06\linewidth]{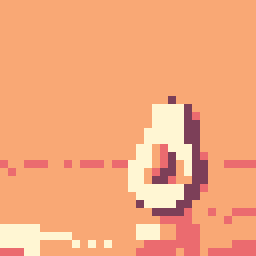} }} & 
\multicolumn{24}{c}{}\\

\\
  \multicolumn{4}{c}{\multirow{4}{*}{\includegraphics[width=0.12\linewidth]{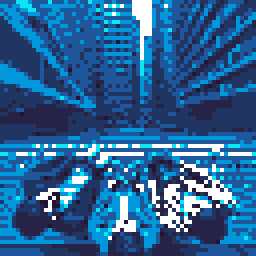} }}
 & \multicolumn{4}{c}{\multirow{4}{*}{\includegraphics[width=0.12\linewidth]{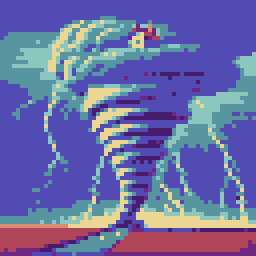} }}
 & \multicolumn{4}{c}{\multirow{4}{*}{\includegraphics[width=0.12\linewidth]{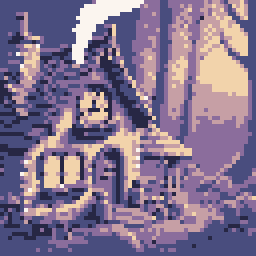} }}
 & \multicolumn{4}{c}{\multirow{4}{*}{\includegraphics[width=0.12\linewidth]{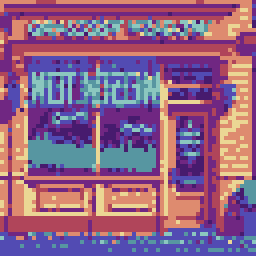} }}
 & \multicolumn{4}{c}{\multirow{4}{*}{\includegraphics[width=0.12\linewidth]{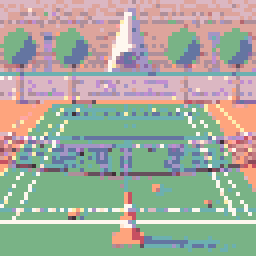} }}
 & \multicolumn{4}{c}{\multirow{4}{*}{\includegraphics[width=0.12\linewidth]{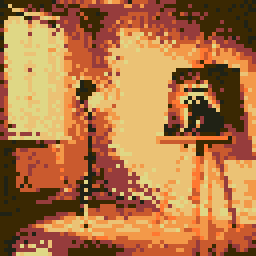} }}
 & \multicolumn{4}{c}{\multirow{4}{*}{\includegraphics[width=0.12\linewidth]{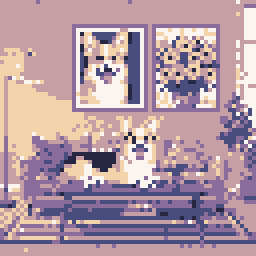} }}
 & \multicolumn{4}{c}{\multirow{4}{*}{\includegraphics[width=0.12\linewidth]{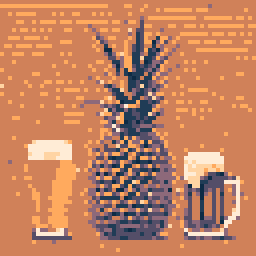} }}
 \\
 \\
 \\
 \\
  \multicolumn{4}{c}{\multirow{4}{*}{\includegraphics[width=0.12\linewidth]{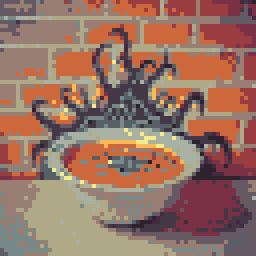} }}
 & \multicolumn{4}{c}{\multirow{4}{*}{\includegraphics[width=0.12\linewidth]{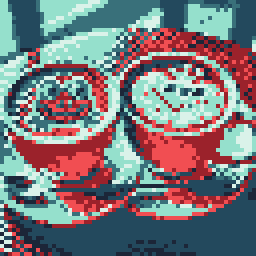} }}
 & \multicolumn{4}{c}{\multirow{4}{*}{\includegraphics[width=0.12\linewidth]{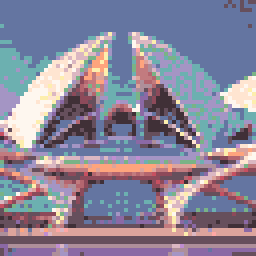} }}
 & \multicolumn{4}{c}{\multirow{4}{*}{\includegraphics[width=0.12\linewidth]{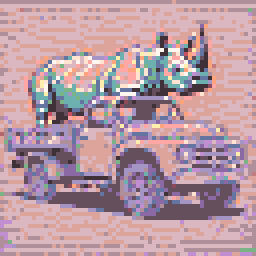} }}
 & \multicolumn{4}{c}{\multirow{4}{*}{\includegraphics[width=0.12\linewidth]{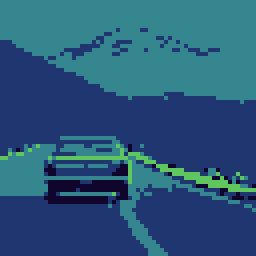} }}
 & \multicolumn{4}{c}{\multirow{4}{*}{\includegraphics[width=0.12\linewidth]{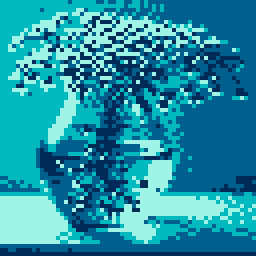} }}
 & \multicolumn{4}{c}{\multirow{4}{*}{\includegraphics[width=0.12\linewidth]{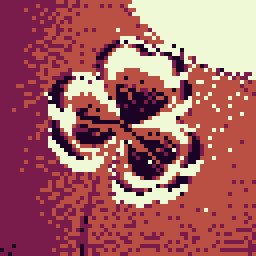} }}
 & \multicolumn{4}{c}{\multirow{4}{*}{\includegraphics[width=0.12\linewidth]{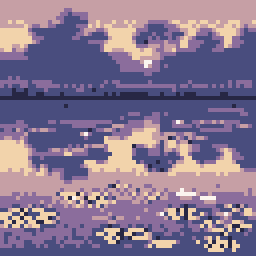} }}
 \\
 \\
 \\
 \\
 \midrule

 \multicolumn{32}{c}{K-means} \\
\multicolumn{2}{c}{\multirow{2}{*}{\includegraphics[width=0.06\linewidth]{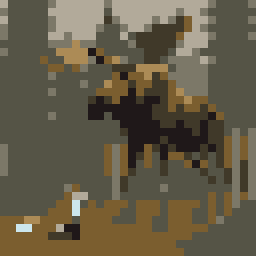} }} & 
\multicolumn{2}{c}{\multirow{2}{*}{\includegraphics[width=0.06\linewidth]{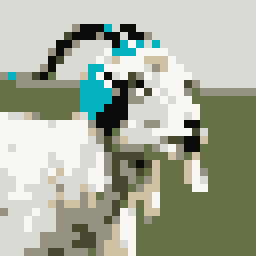} }} & 
\multicolumn{2}{c}{\multirow{2}{*}{\includegraphics[width=0.06\linewidth]{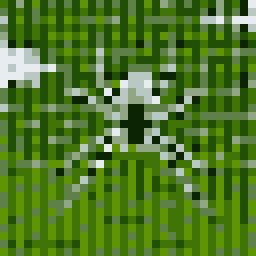} }} & 
\multicolumn{2}{c}{\multirow{2}{*}{\includegraphics[width=0.06\linewidth]{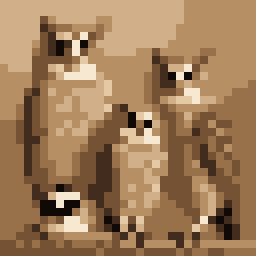} }} &

\multicolumn{3}{c}{\multirow{3}{*}{\includegraphics[width=0.09\linewidth]{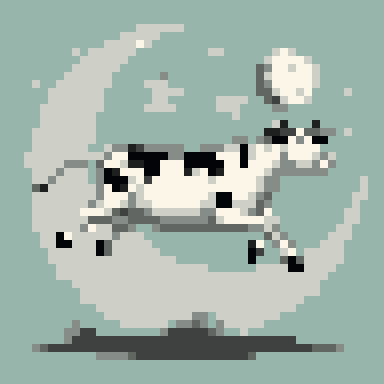} }} & 
\multicolumn{3}{c}{\multirow{3}{*}{\includegraphics[width=0.09\linewidth]{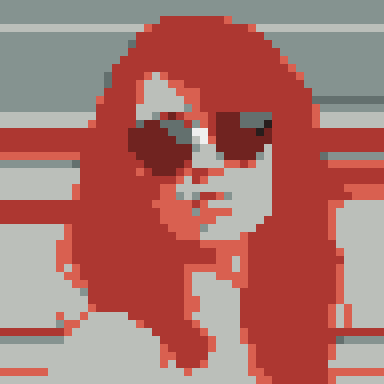} }} & 
\multicolumn{3}{c}{\multirow{3}{*}{\includegraphics[width=0.09\linewidth]{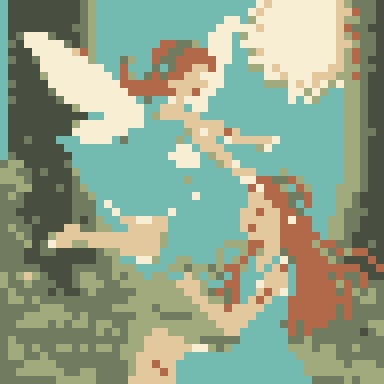} }} & 
\multicolumn{3}{c}{\multirow{3}{*}{\includegraphics[width=0.09\linewidth]{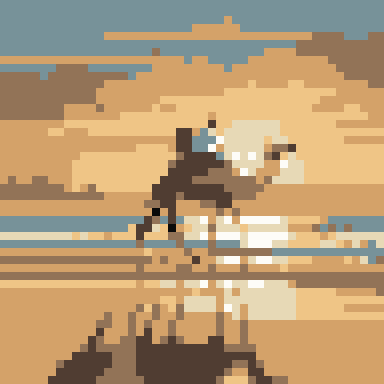} }} & 
\multicolumn{3}{c}{\multirow{3}{*}{\includegraphics[width=0.09\linewidth]{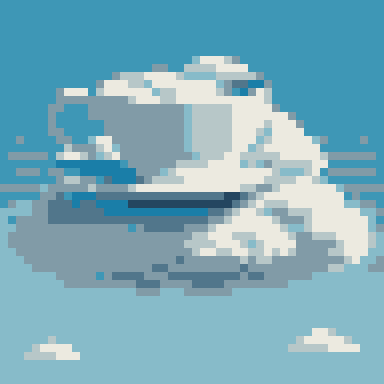} }} & 
\multicolumn{3}{c}{\multirow{3}{*}{\includegraphics[width=0.09\linewidth]{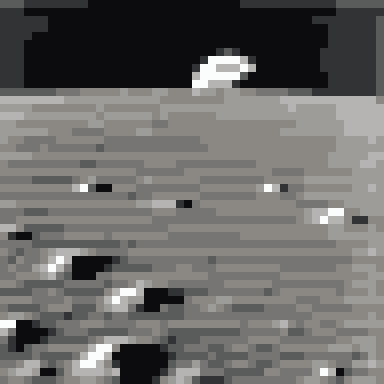} }} & 
\multicolumn{3}{c}{\multirow{3}{*}{\includegraphics[width=0.09\linewidth]{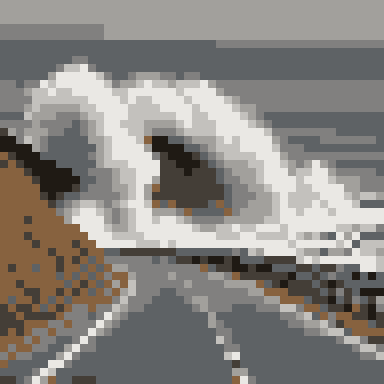} }} & 
\multicolumn{3}{c}{\multirow{3}{*}{\includegraphics[width=0.09\linewidth]{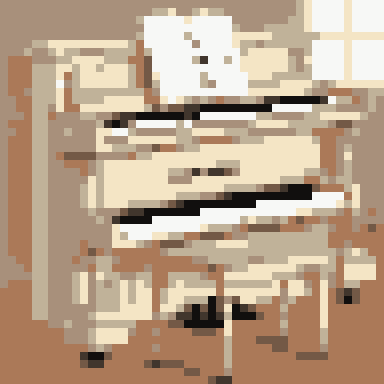} }} \\

\\


\multicolumn{2}{c}{\multirow{2}{*}{\includegraphics[width=0.06\linewidth]{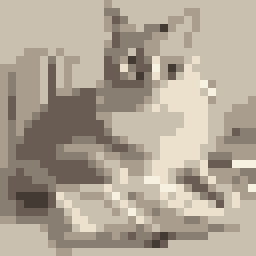} }} & 
\multicolumn{2}{c}{\multirow{2}{*}{\includegraphics[width=0.06\linewidth]{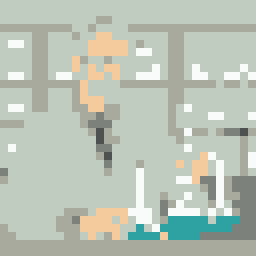} }} & 
\multicolumn{2}{c}{\multirow{2}{*}{\includegraphics[width=0.06\linewidth]{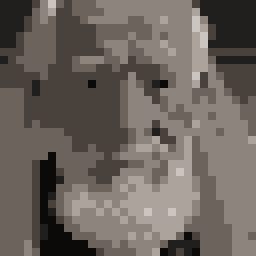} }} & 
\multicolumn{2}{c}{\multirow{2}{*}{\includegraphics[width=0.06\linewidth]{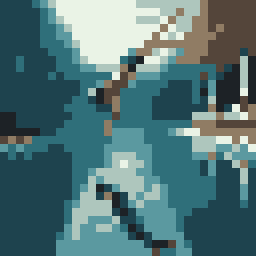} }} & 

 \multicolumn{24}{c}{}

 \\
 

 \multicolumn{8}{c}{} & 
\multicolumn{3}{c}{\multirow{3}{*}{\includegraphics[width=0.09\linewidth]{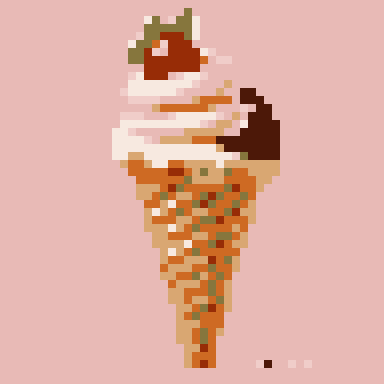} }} & 
\multicolumn{3}{c}{\multirow{3}{*}{\includegraphics[width=0.09\linewidth]{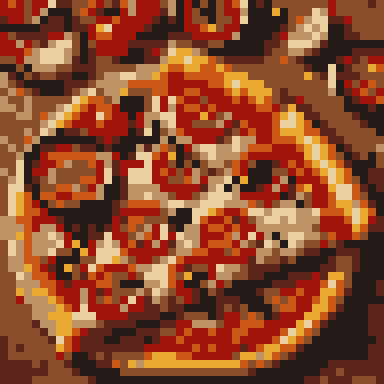} }} & 
\multicolumn{3}{c}{\multirow{3}{*}{\includegraphics[width=0.09\linewidth]{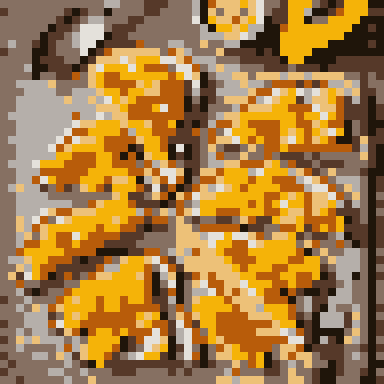} }} & 
\multicolumn{3}{c}{\multirow{3}{*}{\includegraphics[width=0.09\linewidth]{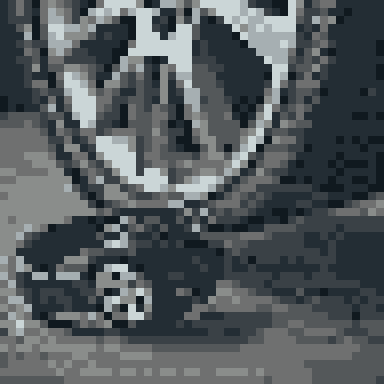} }} & 
\multicolumn{3}{c}{\multirow{3}{*}{\includegraphics[width=0.09\linewidth]{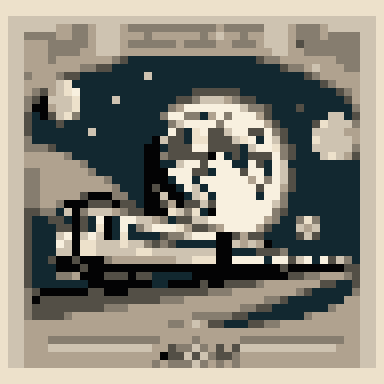} }} & 
\multicolumn{3}{c}{\multirow{3}{*}{\includegraphics[width=0.09\linewidth]{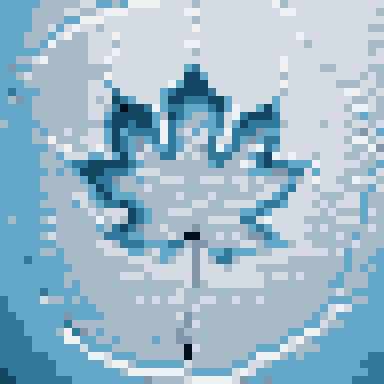} }} & 
\multicolumn{3}{c}{\multirow{3}{*}{\includegraphics[width=0.09\linewidth]{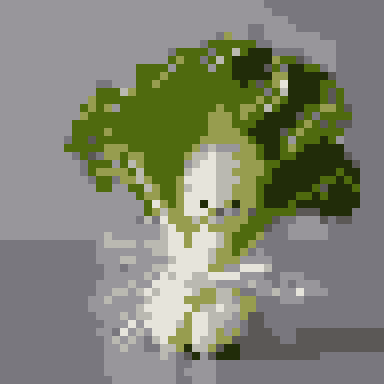} }} & 
\multicolumn{3}{c}{\multirow{3}{*}{\includegraphics[width=0.09\linewidth]{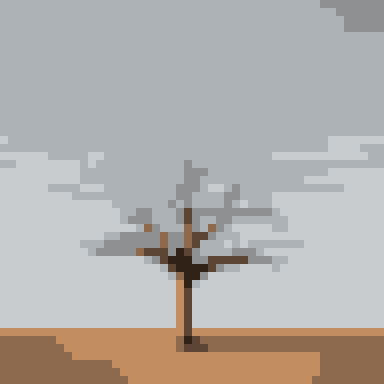} }}
\\
\multicolumn{2}{c}{\multirow{2}{*}{\includegraphics[width=0.06\linewidth]{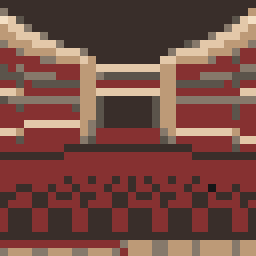} }} & 
\multicolumn{2}{c}{\multirow{2}{*}{\includegraphics[width=0.06\linewidth]{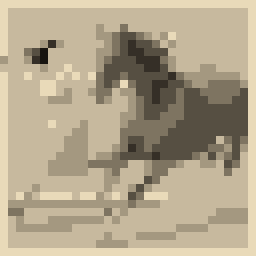} }} & 
\multicolumn{2}{c}{\multirow{2}{*}{\includegraphics[width=0.06\linewidth]{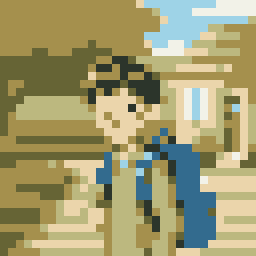} }} & 
\multicolumn{2}{c}{\multirow{2}{*}{\includegraphics[width=0.06\linewidth]{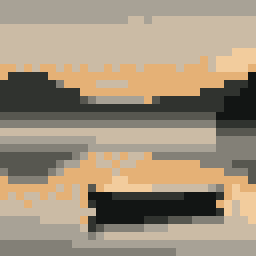} }} & 
 \multicolumn{24}{c}{}
 \\

 
 \\
 
\multicolumn{2}{c}{\multirow{2}{*}{\includegraphics[width=0.06\linewidth]{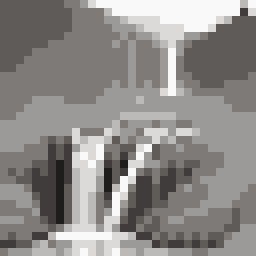} }} & 
\multicolumn{2}{c}{\multirow{2}{*}{\includegraphics[width=0.06\linewidth]{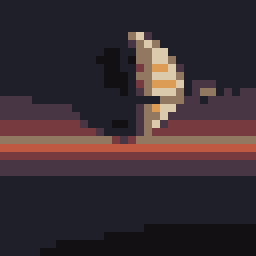} }} & 
\multicolumn{2}{c}{\multirow{2}{*}{\includegraphics[width=0.06\linewidth]{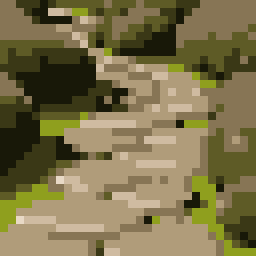} }} & 
\multicolumn{2}{c}{\multirow{2}{*}{\includegraphics[width=0.06\linewidth]{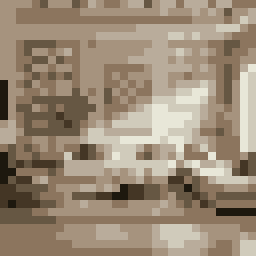} }} &

\multicolumn{4}{c}{\multirow{4}{*}{\includegraphics[width=0.12\linewidth]{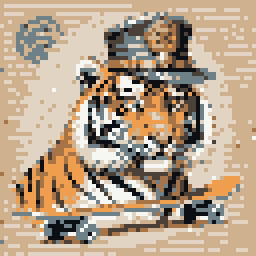} }} &
\multicolumn{4}{c}{\multirow{4}{*}{\includegraphics[width=0.12\linewidth]{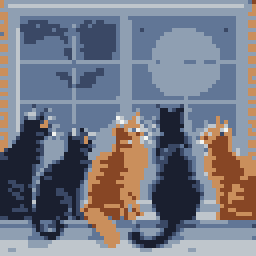} }} &
\multicolumn{4}{c}{\multirow{4}{*}{\includegraphics[width=0.12\linewidth]{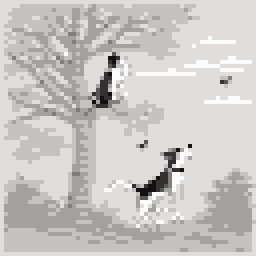} }} &
\multicolumn{4}{c}{\multirow{4}{*}{\includegraphics[width=0.12\linewidth]{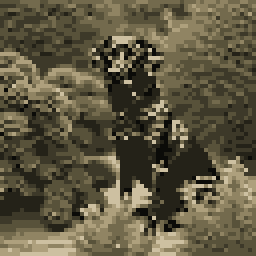} }} &
\multicolumn{4}{c}{\multirow{4}{*}{\includegraphics[width=0.12\linewidth]{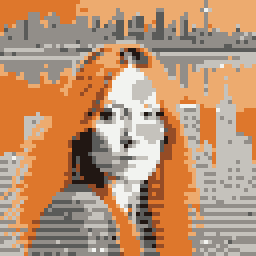} }} &
\multicolumn{4}{c}{\multirow{4}{*}{\includegraphics[width=0.12\linewidth]{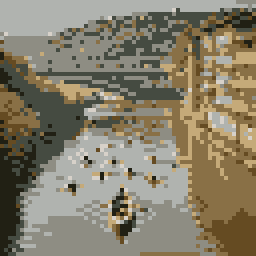} }}  \\
 \\
\multicolumn{2}{c}{\multirow{2}{*}{\includegraphics[width=0.06\linewidth]{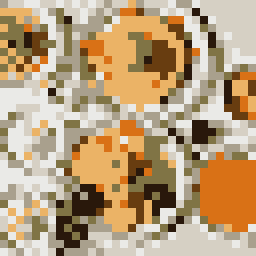} }} & 
\multicolumn{2}{c}{\multirow{2}{*}{\includegraphics[width=0.06\linewidth]{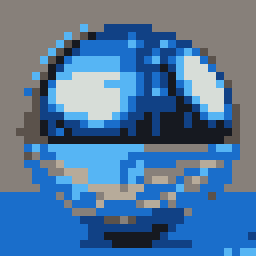} }} & 
\multicolumn{2}{c}{\multirow{2}{*}{\includegraphics[width=0.06\linewidth]{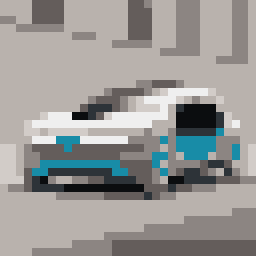} }} & 
\multicolumn{2}{c}{\multirow{2}{*}{\includegraphics[width=0.06\linewidth]{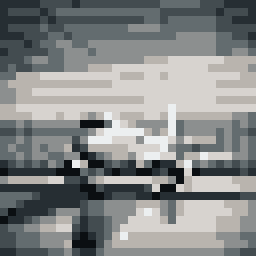} }} & 
\multicolumn{24}{c}{}\\

\\
  \multicolumn{4}{c}{\multirow{4}{*}{\includegraphics[width=0.12\linewidth]{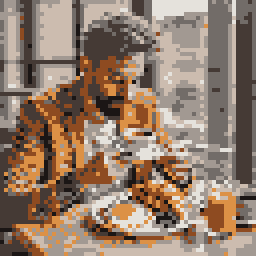} }}
 & \multicolumn{4}{c}{\multirow{4}{*}{\includegraphics[width=0.12\linewidth]{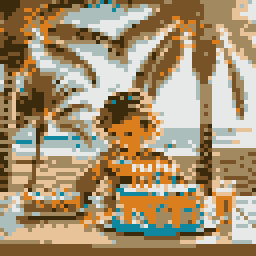} }}
 & \multicolumn{4}{c}{\multirow{4}{*}{\includegraphics[width=0.12\linewidth]{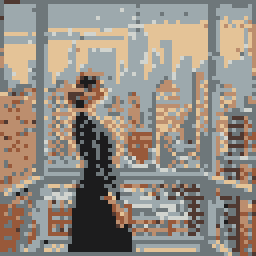} }}
 & \multicolumn{4}{c}{\multirow{4}{*}{\includegraphics[width=0.12\linewidth]{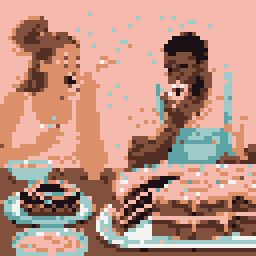} }}
 & \multicolumn{4}{c}{\multirow{4}{*}{\includegraphics[width=0.12\linewidth]{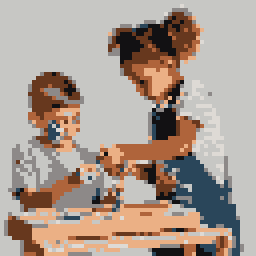} }}
 & \multicolumn{4}{c}{\multirow{4}{*}{\includegraphics[width=0.12\linewidth]{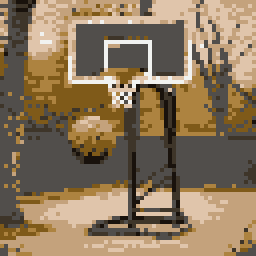} }}
 & \multicolumn{4}{c}{\multirow{4}{*}{\includegraphics[width=0.12\linewidth]{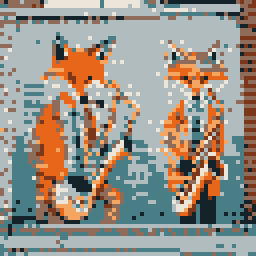} }}
 & \multicolumn{4}{c}{\multirow{4}{*}{\includegraphics[width=0.12\linewidth]{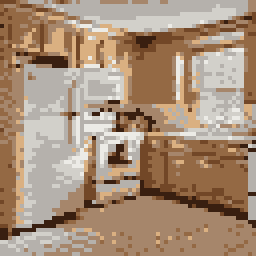} }}
 \\
 \\
 \\
 \\
  \multicolumn{4}{c}{\multirow{4}{*}{\includegraphics[width=0.12\linewidth]{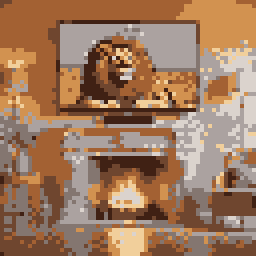} }}
 & \multicolumn{4}{c}{\multirow{4}{*}{\includegraphics[width=0.12\linewidth]{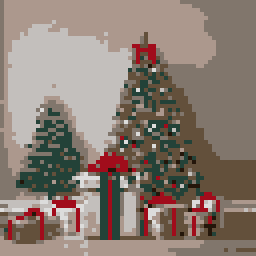} }}
 & \multicolumn{4}{c}{\multirow{4}{*}{\includegraphics[width=0.12\linewidth]{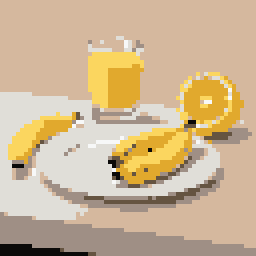} }}
 & \multicolumn{4}{c}{\multirow{4}{*}{\includegraphics[width=0.12\linewidth]{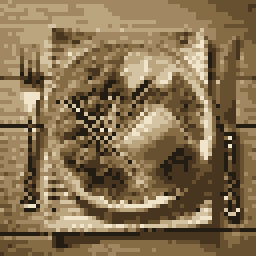} }}
 & \multicolumn{4}{c}{\multirow{4}{*}{\includegraphics[width=0.12\linewidth]{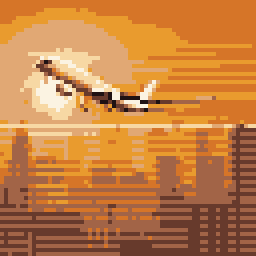} }}
 & \multicolumn{4}{c}{\multirow{4}{*}{\includegraphics[width=0.12\linewidth]{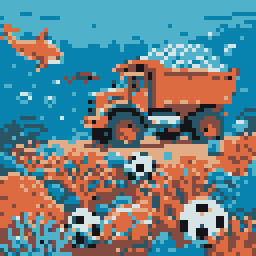} }}
 & \multicolumn{4}{c}{\multirow{4}{*}{\includegraphics[width=0.12\linewidth]{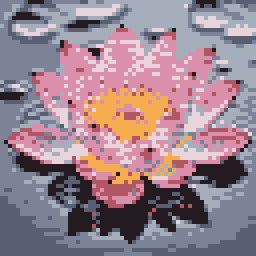} }}
 & \multicolumn{4}{c}{\multirow{4}{*}{\includegraphics[width=0.12\linewidth]{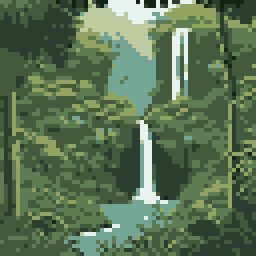} }}
 \\
 \\
 \\
 \\
 \bottomrule
 \addlinespace[-0.2cm]
 	\end{tabular}
	\caption{Results of \ourmethod{} on the dataset for quantitative evaluation are presented for both \emph{palette} (top) and \emph{K-means} (bottom) versions. Results in three different sizes are shown: $32 \times 32$, $48 \times 48$, and $64 \times 64$, each scaled proportionally to its dimensions.}
	\label{fig:bigfigure2}
\end{figure*}

%% file: sections/supplementary/userstudy_image.tex
\begin{figure*}[ht]
    \centering
    \setlength{\tabcolsep}{1pt}
    \begin{tabular}{cc}
    {User study questionnaire} & {Histogram of user responses} \\    \includegraphics[width=0.470\linewidth]{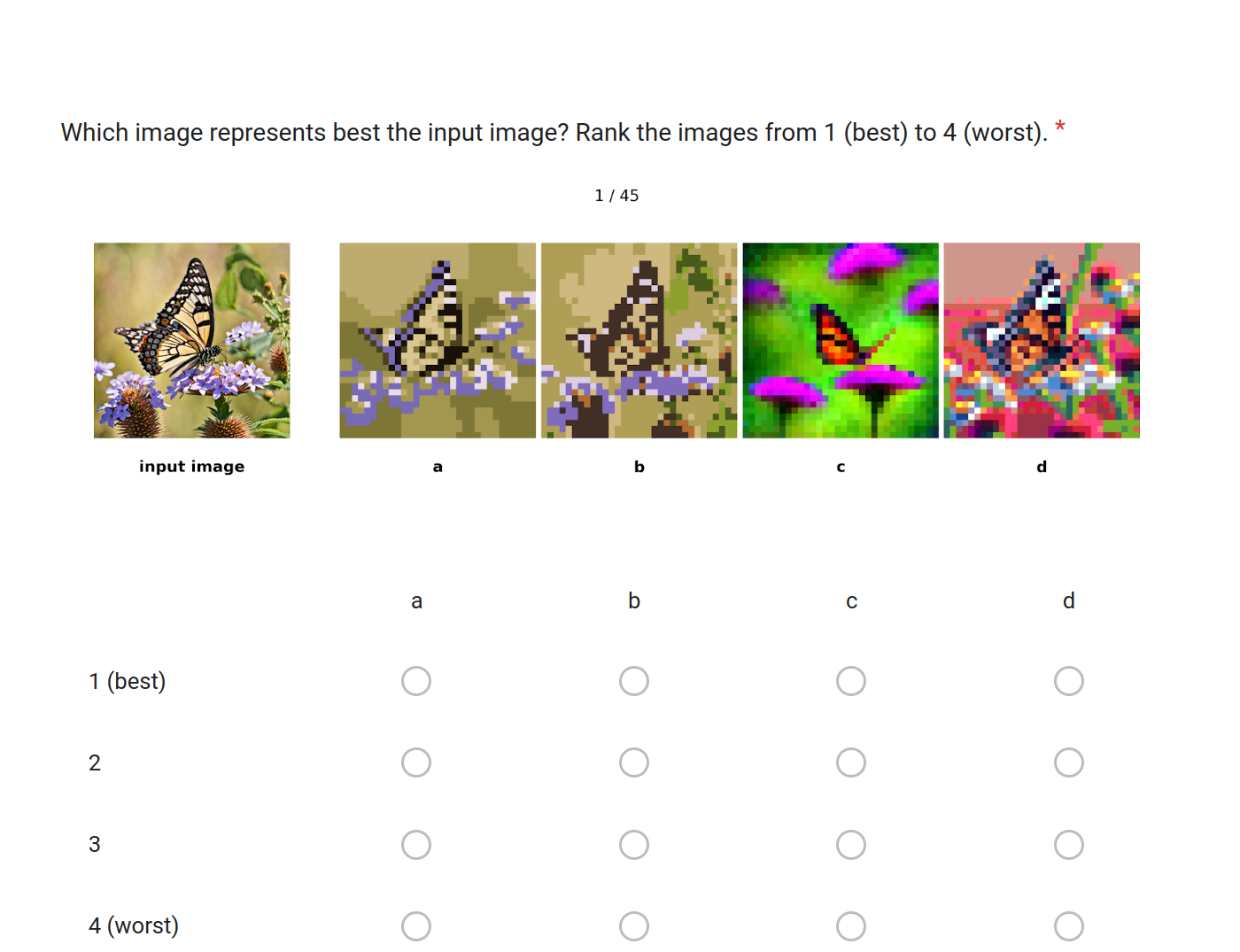} & \includegraphics[width=0.525\linewidth]{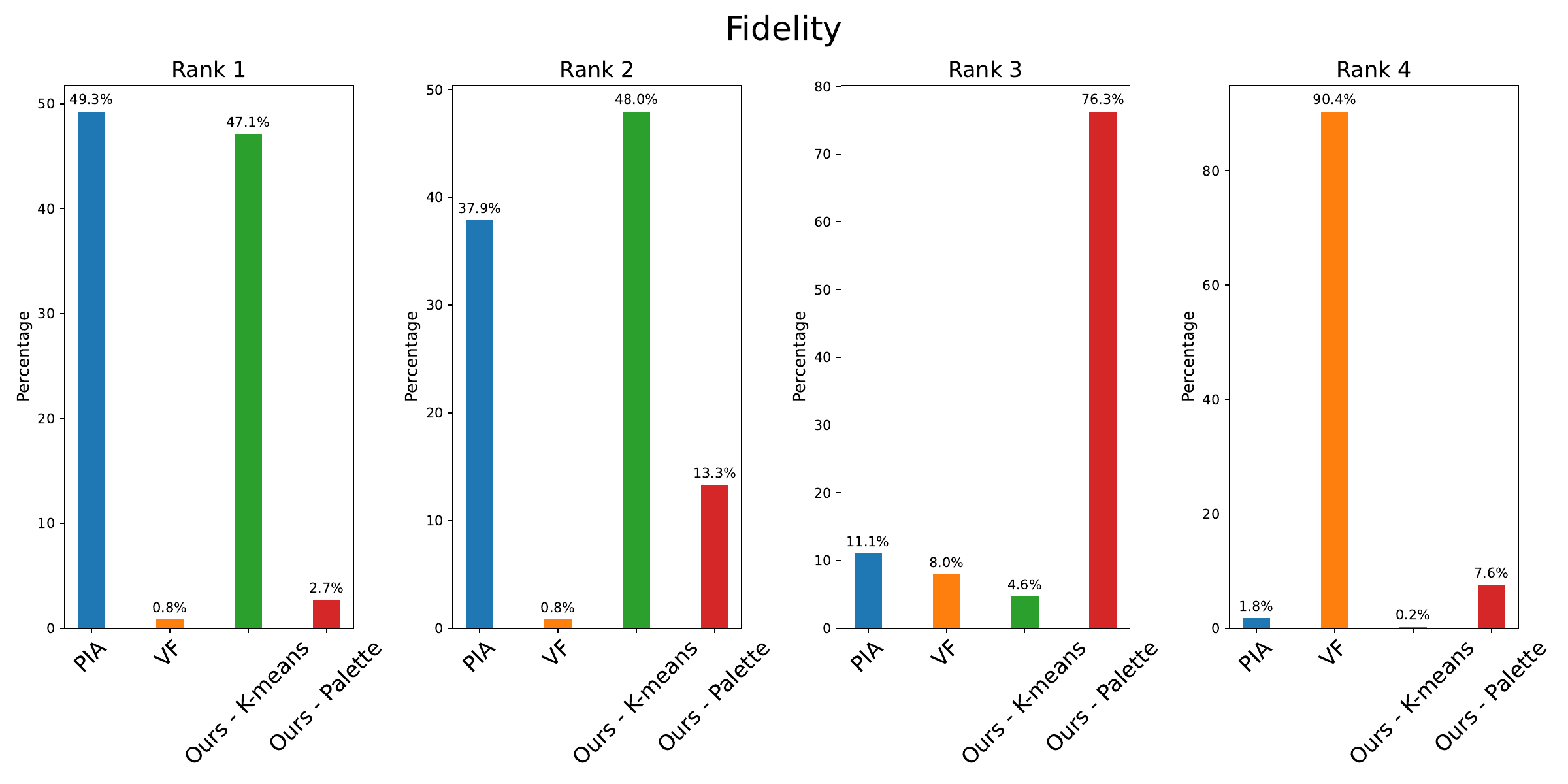} \\
    \includegraphics[width=0.470\linewidth]{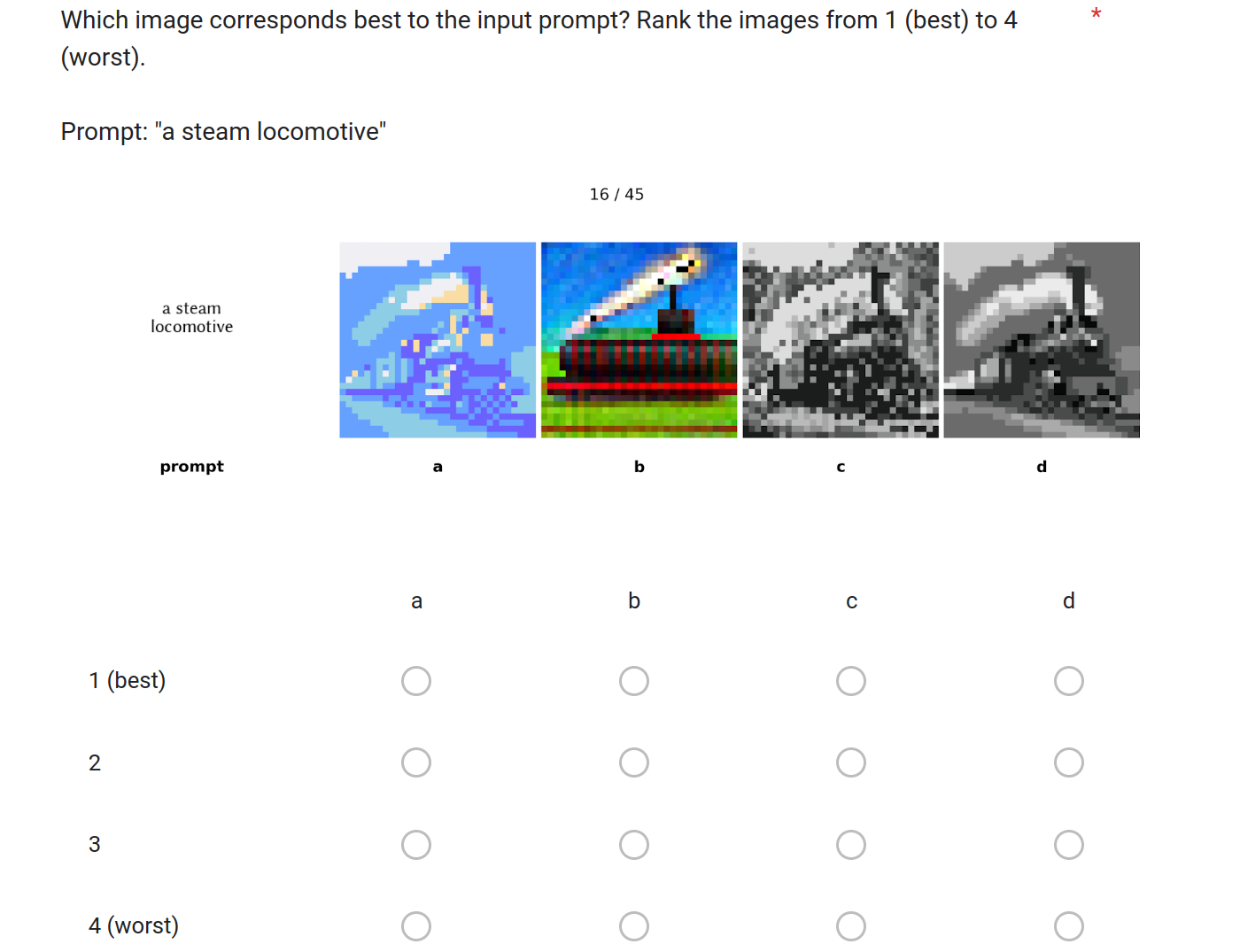} & \includegraphics[width=0.525\linewidth]{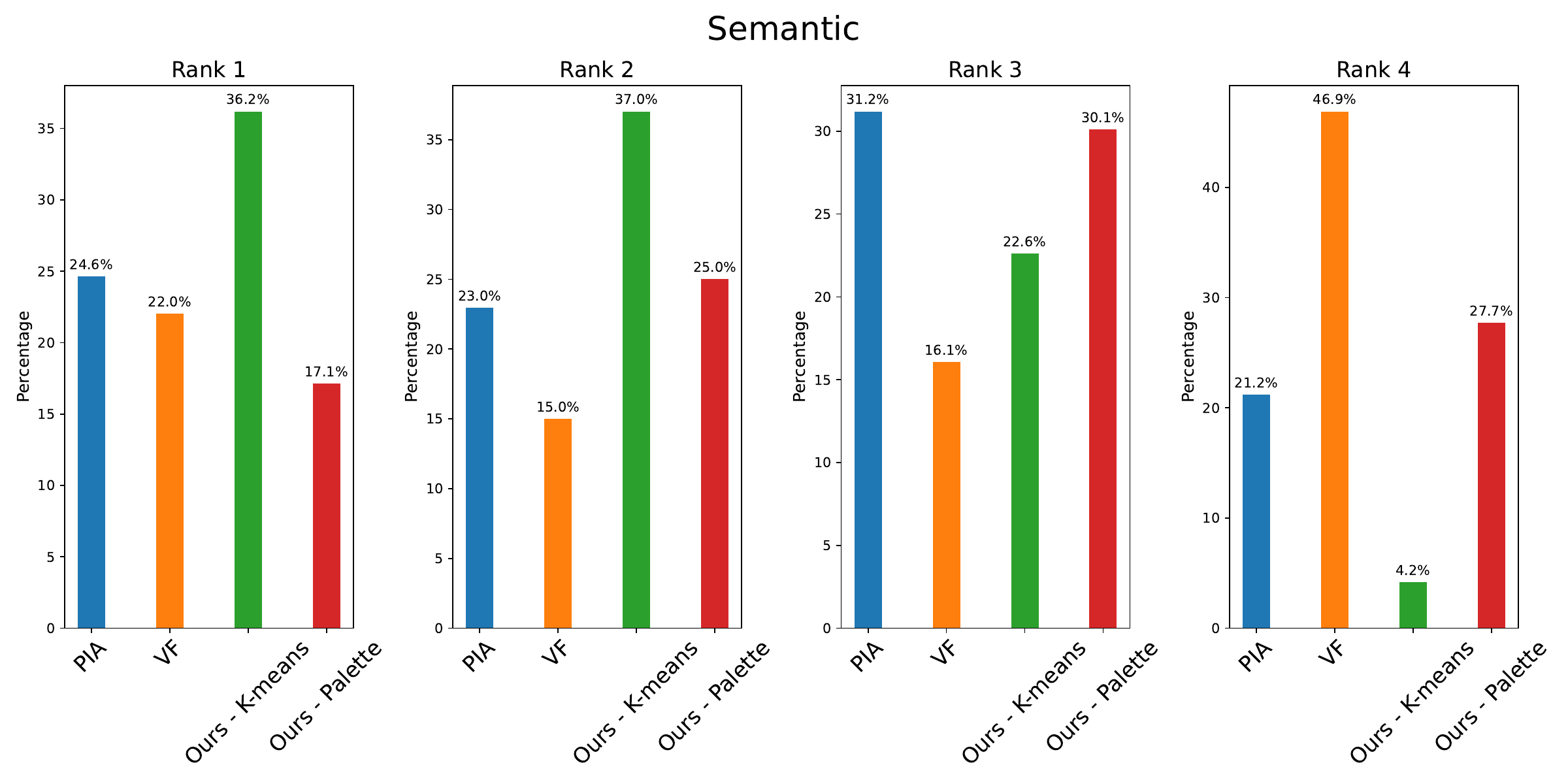}\\
    \includegraphics[width=0.470\linewidth]{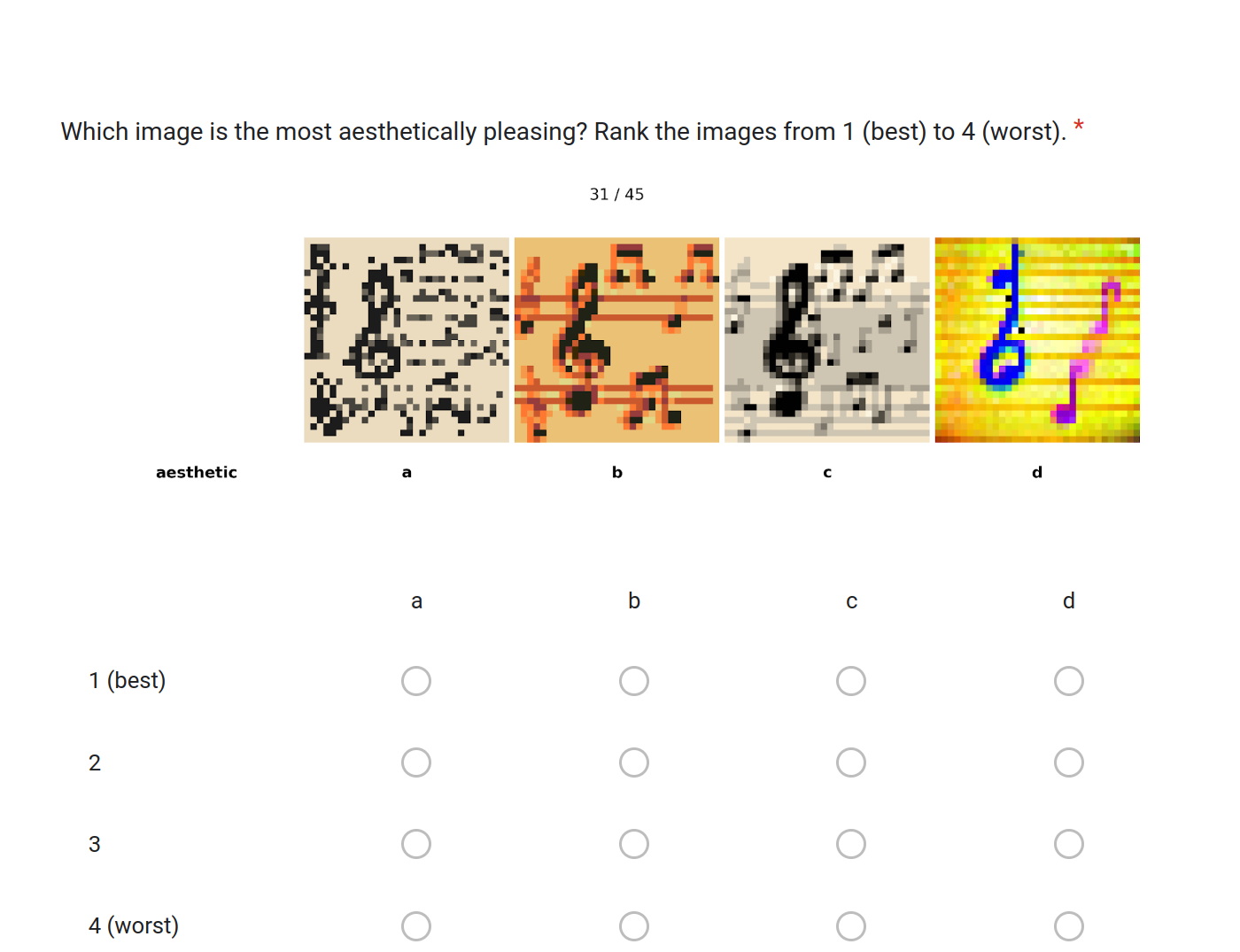} & \includegraphics[width=0.525\linewidth]{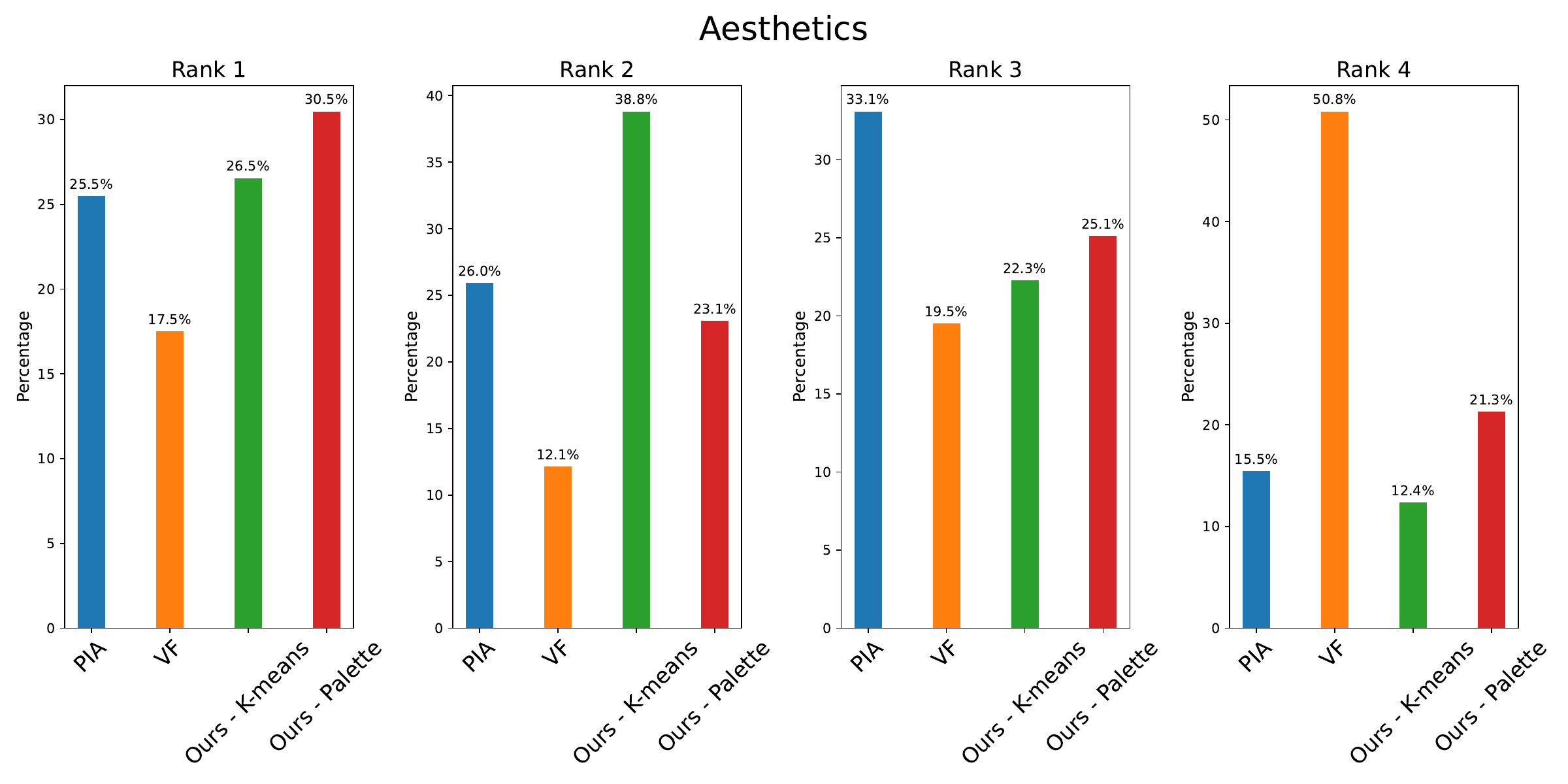}
    \end{tabular}
    \caption{Overview of the user study design and outcomes. The left column displays the questionnaires used in the study, categorized into three sections: image fidelity (first 15 questions), prompt similarity (next 15 questions), and aesthetics preference (last 15 questions). Presentation order of different methods is randomized in each section. The right column visualizes the aggregate user responses across these categories as histograms.}
    \label{fig:userstudyquestions}
\end{figure*}